\documentclass[final,3p,numafflabel]{elsarticle}
\usepackage{amsmath}
\usepackage{amssymb}
\usepackage{amsthm}
\usepackage{newtxtext}
\usepackage{newtxmath}
\usepackage{microtype}
\usepackage{xurl}
\usepackage{graphicx}
\usepackage{subcaption}
\usepackage{tabularx}
\usepackage{booktabs}
\usepackage{array}
\usepackage{multirow}
\usepackage{mathtools}
\usepackage{bm}
\usepackage{xspace}

\usepackage{algorithm}
\usepackage{algorithmic}

\usepackage{wrapfig}

\usepackage[dvipsnames,svgnames,x11names]{xcolor}
\definecolor{lightblue}{HTML}{2970CC}

\usepackage[
    colorlinks=true,
    linkcolor=lightblue,
    urlcolor=lightblue,
    citecolor=lightblue,
    breaklinks=true
]{hyperref}

\theoremstyle{plain}
\newtheorem{theorem}{Theorem}[section]
\newtheorem{proposition}[theorem]{Proposition}

\newtheorem{corollary}[theorem]{Corollary}

\theoremstyle{definition}

\theoremstyle{remark}
\newtheorem{remark}[theorem]{Remark}

\newcommand{\ourname}{AdjointFlow\xspace}
\newcommand{\ournameRED}{RED-Flow\xspace}
\newcommand{\R}{\ensuremath{\mathbb{R}}}
\newcommand{\calF}{\mathcal{F}}
\newcommand{\calM}{\mathcal{M}}
\newcommand{\calY}{\mathcal{Y}}
\newcommand{\calN}{\mathcal{N}}
\newcommand{\calD}{\mathcal{D}}
\newcommand{\calI}{\mathcal{I}}
\renewcommand{\mid}{\,\vert\,}

\begin{document}

\begin{frontmatter}

\title{AdjointFlow: Accelerating Scientific Inverse Problems with Physics-Based Transport}

\author[aff1]{Ethan Hanold}
\author[aff2]{Nicholas M.\ Boffi}
\author[aff1]{Qin Li}
\author[aff3]{Leonardo Zepeda-N\'u\~nez}

\affiliation[aff1]{
    organization={University of Wisconsin--Madison},
    city={Madison},
    state={WI},
    country={USA}
}

\affiliation[aff2]{
    organization={Carnegie Mellon University},
    city={Pittsburgh},
    state={PA},
    country={USA}
}

\affiliation[aff3]{
    organization={Google Research},
    city={Mountain View},
    state={CA},
    country={USA}
}

\begin{abstract}
Inverse problems are ubiquitous across scientific applications ranging from seismic imaging to astrophysics to nondestructive testing, yet classical solvers often struggle with nonconvexity and repeated evaluations of expensive physical models. Diffusion models can amortize this computation offline, but largely data-driven conditional models underutilize known physics and suffer from ill-conditioned transport maps when the target distribution is concentrated. Plug-and-play (PnP) methods explicitly leverage the physical model to guide a learned prior, improving data consistency but reintroducing costly iterative computations. We instead propose AdjointFlow, a framework built on stochastic interpolants that injects the known physics into the generative transport itself. We design a data-dependent, adjoint-based initialization to reduce transport distance to the target, modify the transport velocity to remain regular as the endpoint distributions concentrate, and parameterize the network to build the initialization directly into the learned velocity. Together, these enable accurate reconstruction at substantially lower cost than generative models starting from Gaussian noise. Experiments on inverse scattering, seismic imaging, and black hole imaging show that AdjointFlow consistently produces more accurate reconstructions than conditional diffusion models while being up to five orders of magnitude faster at inference than PnP methods. Optional inference-time refinement that queries the physical model can further improve data consistency, matching or exceeding the accuracy of PnP methods without retraining.
\end{abstract}


\end{frontmatter}


\section{Introduction}

Inverse problems are ubiquitous across science and engineering, arising in diverse domains ranging from molecular X-ray diffraction~\cite{friedrich1913interferenzerscheinungen} and seismic imaging~\cite{bleistein2001mathematics} to medical tomography~\cite{natterer2001mathematics} and astrophysics~\cite{bouman2016computational}.
In each case, the objective is to recover unknown structural or material properties of a medium from indirect measurements, typically acquired at the boundary.
Unlike standard image-processing tasks, these scientific inverse problems are governed by physical forward models that can be highly nonlinear, often leading to a nonconvex optimization landscape containing numerous nonphysical, or ``spurious,'' local minima~\cite{Leeuwen_Herrmann:2013}.
Consequently, computing or learning the inverse map from measurements to the underlying medium can be especially difficult.

\begin{figure*}[t]
	\centering
	\includegraphics[width=0.9\linewidth]{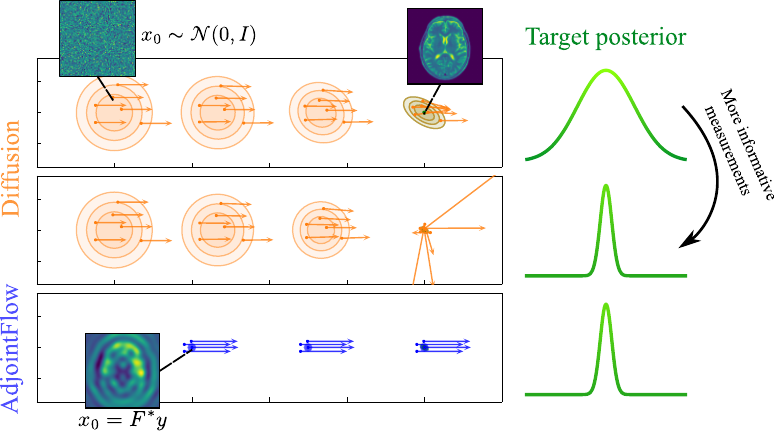}
	\caption{\textbf{Overview.}
    Traditional diffusion-based approaches start from uninformative Gaussian noise and can suffer from ill-conditioned transport fields when the target posterior becomes concentrated, such as when more informative measurements are acquired.
    Conversely, \textbf{\ourname} starts from a data-dependent initialization -- for example, one constructed from the adjoint $F^*y$ as shown above -- reducing transport distance relative to Gaussian noise and producing favorable learning conditions for the solver.
    \ourname further leverages a noise-free velocity which remains regular near concentrated endpoints.}
	\label{fig:overview}
\end{figure*}

Wave-based inverse problems are perhaps the most representative of these challenges~\cite{Hahner_Hohage2001_inverse_problem_estimates}.
Historically\footnote{For a more extensive review of the literature please see Section~\ref{sec:related}.}, hand-crafted asymptotic approximations~\cite{ColtonKress2013:integrals} have been used to obtain computationally efficient, qualitative reconstructions~\cite{cakoni2005qualitative,colton1998inverse}.
However, the advent of modern computers coupled with advances in numerical simulation has made PDE-constrained optimization~\cite{tarantola2005inverse,Pratt:1999,virieux2014introduction} the prevalent approach for numerical inversion.
These optimization methods are based on well-established computational techniques and can yield high-quality reconstructions~\cite{Virieux_Operto}. Nevertheless, their nonconvex optimization landscapes are often plagued by spurious local minima, making the reconstruction sensitive to initialization and requiring carefully designed algorithmic strategies~\cite{li2016full,Symes_Carazzone:1991,chen1997inverse}.
Moreover, every iteration necessitates several calls to the forward map (and its adjoint), which in turn entails solving the underlying physical model -- typically a system of partial differential equations (PDEs). Repeated PDE solves render the overall computation prohibitive as the desired reconstruction resolution increases.

Over the past decade, machine learning (ML) has emerged as a promising paradigm for inverse problems, offering the ability to \emph{amortize} the cost of inversion.
Within this paradigm, a neural network is trained offline on a collection of examples and subsequently deployed to perform rapid inference, typically bypassing the iterative optimization and repeated PDE solves.
ML-based approaches broadly fall into two categories.
\emph{Deterministic} methods directly learn a point estimate of the reconstruction, typically by training highly-tailored neural networks to approximate the inverse map~\cite{Butterfly-Net2,KhooYingSwitchNet:2019,zhang2024_equivariant}.
\emph{Generative} methods, by contrast, seek to learn the underlying distribution rather than a single point estimate. Modern generative models based on diffusion~\cite{song2019generative,ho2020ddpm} or flow matching~\cite{lipman2023flow,albergo2025stochastic} generate samples by learning a transport map from a simple base distribution to a target distribution. 
In the context of inverse problems, this target distribution is the conditional posterior distribution of the unknown parameters given the measurements, providing a natural framework for approximating and sampling from the Bayesian posterior~\cite{stuart2010bayesian}.
Such generative approaches can be classified based on how they incorporate the relationship between measurements and unknown parameters:
Diffusion-based plug-and-play methods first learn an unconditional prior and then incorporate measurement data during inference~\cite{jalal2021robust,chung2023diffusion, zhang2024daps} using highly tailored guidance terms. 
Conditional generative models instead learn the measurement-reconstruction relationship during training, allowing them to estimate the posterior directly~\cite{Dasgupta:2026,Zhang2025_BPD}.

Generative models have recently demonstrated strong performance on inverse problems~\cite{jalal2021robust,song2022solving,chung2023diffusion,wang2023priorfwi,feng2023score,zhang2024daps,Zhang2025_BPD,huang2026selfflow,chung2023cddb,Dasgupta:2026,dawnfm2026}, providing highly accurate reconstructions.
However, most of these methods adopt a fixed Gaussian reference distribution; here we argue that \textbf{\emph{Gaussian-to-posterior transport is poorly suited to scientific inverse problems}}.
We instead adapt the reference distribution and transport velocity to the structure of these problems and obtain an iterative, deterministic method that retains the noise-augmented training of generative models while being able to match or surpass their accuracy at a substantially lower inference cost.

Two features of scientific inverse problems motivate this departure from a fixed Gaussian reference.
First, an independent Gaussian reference discards information already available
through the physical forward model. 
As such, generation must begin from uninformative noise even when cheap physics-based approximations are available, leaving the full relationship between measurements and unknown parameters to be learned from data or injected via guidance.
Second, scientific inverse problems can exhibit strong posterior concentration
as the quantity or quality of measurements increases, {as} described by the Bernstein--von Mises theorem~\cite{vandervaart1998asymptotic}.
Transporting order-one Gaussian noise to an increasingly
concentrated conditional target requires increasingly contractive dynamics and can lead to velocity fields with poor endpoint regularity. 
Such rapidly varying fields can be difficult to approximate accurately and integrate numerically, creating challenges for both learning and inference and ultimately degrading the overall performance of the generative model~\cite{gao2024convergence,kunkel2026distribution}.
Together, these observations motivate a physics-aware reconstruction method with transport dynamics which are easier to learn and which permit accurate reconstruction using fewer integration steps.

Our proposed method, \ourname, addresses these two limitations through two complementary design choices.
First, we exploit the flexibility of the stochastic interpolant (SI) framework~\cite{albergo2025stochastic,albergo2024couplings} to replace the
independent Gaussian reference with a data-dependent, \textit{physics-aware
initialization}.
Specifically, we use an adjoint-based approximation to map measurements into the unknown parameter space, converting paired measurement-reconstruction data into paired transport endpoints.
We find that a reference distribution centered at a physical approximation -- and, in the concentrated posterior regime, one that is similarly concentrated -- reduces transport distance to the desired reconstruction; see Figure~\ref{fig:overview} for an illustration.
Second, we introduce a modified SI velocity field tailored to concentrated endpoint distributions, which remains well-conditioned as these endpoints become singular.
We show that together, these design choices admit an especially natural \textit{data-dependent target prediction} parameterization of the learned velocity, which analytically incorporates the known physical approximation and reduces the task of learning a velocity field to the task of predicting the target reconstruction itself.
In summary, \ourname replaces a generic Gaussian-to-data transport problem with a  physics-informed refinement problem whose velocity is better behaved near concentrated targets and whose learning objective is directly aligned with the desired reconstruction.

Beyond these core components, the flexibility of the SI framework also allows \ourname to interface with inference-time refinement strategies without retraining the underlying model.
Such refinement permits additional computation to be introduced at deployment -- for example, via calls to the forward model -- when greater reconstruction accuracy or physical consistency is desired.
Finally, we evaluate \ourname across a diverse collection of scientific inverse problems, demonstrating that these modeling choices translate into meaningful improvements in both accuracy and efficiency.

\vspace{1em}
Our \textbf{main contributions} can be summarized as follows:
\begin{enumerate} 
    \item We incorporate governing physics into the transport initialization through a singular, \textit{adjoint-based reference distribution}, and analyze conditions under which this choice reduces transport distance to the target posterior.
    \item We characterize the regularity issues that arise in the SI velocity field with highly concentrated endpoint distributions and develop a modified \textit{noise-free velocity} with improved regularity in this regime.
    \item We combine these design choices into an iterative, deterministic reconstruction method which admits a natural \textit{data-dependent target prediction} parameterization. This parameterization analytically incorporates the physical initialization and reduces velocity learning to predicting the target reconstruction from time-dependent, noise-augmented states.
	\item We demonstrate that \ourname can be combined with optional plug-and-play algorithms that interface with the physical forward model at inference time, enabling training-free improvements to reconstruction accuracy and physical consistency.
	\item We evaluate our framework on scientific inverse problems from two benchmarks~\cite{Zhang2025_BPD, zheng2025inversebench} covering inverse scattering, seismic imaging, and black hole imaging. We observe up to $\sim$70\% lower reconstruction error compared to state-of-the-art inversion algorithms -- even when they use larger or more complex architectures -- and speedups of one to five orders of magnitude.
\end{enumerate}

The remainder of the paper is organized as follows.
In Section~\ref{sec:problem}, we review background details on scientific inverse problems and stochastic interpolants.
In Section~\ref{sec:method}, we introduce the methodological framework of \ourname, developing a generative algorithm for scientific inverse problems in the concentrated posterior regime.
We present numerical results in Section~\ref{sec:experiments} and ablate our primary design choices in Section~\ref{sec:ablations}.
In~Section~\ref{sec:related} we provide a comprehensive discussion of related work.
We conclude with final remarks in Section~\ref{sec:conclusion}.

\section{Background and Preliminaries}
\label{sec:problem}

In this section, we briefly review scientific inverse problems, which provide the motivating problem setting, and stochastic interpolants, which provide our methodological framework.
We emphasize the features leveraged by our method and the limitations it is designed to address.

\subsection{Inverse Problems}\label{sec:inverse_problems}

The goal in inverse problems {is} to recover an unknown signal from indirect, possibly noisy measurements.
The measurement procedure is represented by a forward problem $y = \calF(m)$, where the forward function $\calF: \calM \rightarrow \calY$ maps parameters\footnote{Namely, system or PDE parameters such as wavespeed or refractive index. Often, these parameters are functions.} 
$m \in \calM$ to observational data $y \in \calY$, and the inverse problem subsequently asks to recover $m$ from $y$.
In scientific settings, $\calF$ typically involves a physical model characterized by a PDE with measurements obtained from the resulting physical state (such as at the domain boundary).

\begin{figure}
	\centering
	\includegraphics[width=0.72\linewidth]{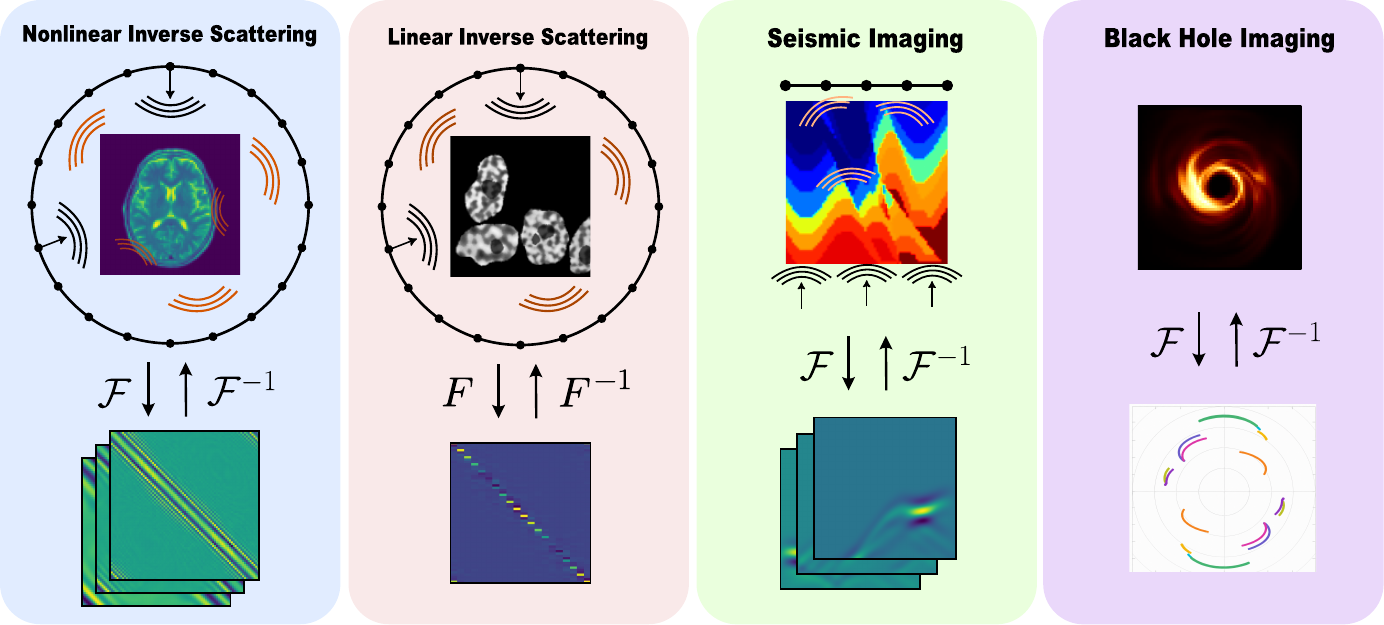}
	\caption{Diagram of the different settings and modalities of inverse problems we consider. Top: a diagram of the data acquisition setting, including a representative example of the parameters $m$ represented by the wavespeed (in the case of scattering problems) or brightness (in the case of black holes). Bottom: an example of the measurement data used for the reconstruction. Note that the full aperture scattering is measured in the frequency domain, while the geophysical scattering is measured in the time domain.}
	\label{fig:imaging_regimes}
\end{figure}

While we primarily focus on inverse scattering, a canonical wave-based scientific inverse problem illustrated in the three leftmost settings in Figure~\ref{fig:imaging_regimes}, our framework is general and applicable to arbitrary scientific inverse problems, particularly those which admit an efficient linearized solution.

A classical approach to numerically solve the inverse problem is to formulate it as a PDE-constrained optimization problem. 
In this setting, one searches for the parameters $m$ whose predicted measurements $\calF(m)$ best match the true observed data $y$. 
Namely, one seeks the minimizer to the following least-squares problem:
\begin{equation}\label{eq:inv_problem_optimization_objective}
    m^\star=\underset{m\in\calM}{\operatorname{argmin}}\,\frac{1}{2}\|\calF(m)-y\|_2^2+\frac\epsilon2\|m\|^2.
\end{equation}

For a linear forward operator $F$, an explicit solution is given by the Tikhonov-regularized~\cite{tikhonov1977solutions} reconstruction
\begin{equation}\label{eq:FBP}
m^\star=(F^*F+\epsilon I)^{-1}F^*y,
\end{equation}
sometimes referred to as the filtered back-projection formula (FBP), where $\epsilon>0$ is a parameter introduced to mitigate the numerical rank deficiency of $F^*F$.

$F^*$ is the adjoint of the forward operator, and is often called the \textit{back-projection operator}. 
Intuitively, $F^*$ takes the data measured at the boundary and \emph{propagates it back} into the unknown domain $\calM$, 
providing a rough estimate of where the perturbation is located. 
The operator $(F^*F+\epsilon I)^{-1}$ then acts as a \emph{filtering operator}, which compensates and sharpens the estimate produced by the back-projection. 
Solving the entire linear system is often prohibitive in practice; however, using only the back-projection operator can already yield a strong low-pass approximation -- see, for example,~Figure~\ref{fig:four_panels} -- at a fraction of the computational cost.

In essence, an adjoint can act as an approximate inverse providing an inexpensive initial reconstruction. However, as we show in what follows, in some cases even an approximation of the adjoint can still convey useful information to the reconstruction.

If the problem is nonlinear, a natural extension is to linearize the forward operator around a simple background medium $\overline{m}\in\calM$:
\begin{equation}\label{eq:linearization}
    \calF(\overline{m} + \delta m) \approx \calF(\overline{m}) + F\delta m,
\end{equation}
where now $F=\frac{\partial\calF}{\partial m}(\overline{m})$ is the Jacobian matrix of the forward operator\footnote{For wave-based inverse problems this linearization is often called the Born approximation~\cite{Tarantola:Inversion_of_seismic_reflection_data_in_the_acoustic_approximation}.}.
This approximation simplifies the inversion back into a linearized least-squares minimization with data misfit term $\,\frac{1}{2}\| y - (\mathcal{F}(\overline{m}) + F \delta m) \|^2$, which can be solved with the FBP formula.
Note that the linearized problem is posed in the perturbation $\delta m$ rather than in $m$ itself, so the back-projection is applied to the residual $\widetilde y=y-\calF(\overline m)$ and returns an estimate of $\delta m$. 

When high contrast or strong nonlinearity undermines linearized, FBP-like approximations~\cite{Symes_2009}, one can instead address the fully nonlinear problem through iterative gradient-based optimization.
Adjoints remain central in this setting: the adjoint-state method~\cite{Plessix_2006:ajoint_state} is a key algorithm for evaluating the data misfit gradient with one forward and one adjoint PDE solve, regardless of the dimension of $m$.

However, difficulties still remain.
For one, the map $\calF$ typically involves a compact operator which attenuates high-frequency details in $m$ and renders the problem ill-posed~\cite{hadamard1902problemes} in the sense that small changes in $y$ can lead to large changes in $m$.
If measurement noise is present, or if the physical model contains epistemic uncertainties, the challenge of obtaining a high-quality reconstruction is even more formidable.
Thus, while PDE-constrained optimization provides a general recipe for solving scientific inverse problems, it can still struggle to mitigate instability and characterize uncertainty.

\subsection{Bayesian Perspective}\label{sec:bayesian}
One solution to these challenges is to approach the inverse problem from a Bayesian perspective which explicitly characterizes the uncertainty.
Following Bayes' theorem, we assume the unknown parameters follow a conditional posterior distribution
\begin{equation}\label{eq:bayes_formula}
    p(m \mid y) \propto p(y \mid m) p(m),
\end{equation}
where $p(y \mid m)$ is the likelihood function and $p(m)$ is a prior distribution on $m$. 
A reconstruction $m$ can then be obtained, for example, by computing the maximum a posteriori (MAP) estimator
\begin{equation}\label{eq:map_estimator}
    m^\mathrm{MAP} = \underset{m \in \calM}{\operatorname{argmax}}\, p(m \mid y).
\end{equation}

A regime of particular relevance for scientific inverse problems is that of the small-noise or increasing-information limit. 
In such a regime, identifiable inverse problems -- those in which unknown parameters are uniquely determined by noiseless observations -- can exhibit posterior concentration around the ground truth $m^\dagger$, provided the inversion is also stable and the prior satisfies suitable support conditions~\cite{lu2017bernsteinvonmisestheoremhigh,Agapiou_2013}. 
Many scientific inverse problems, such as wave-based inverse problems, admit uniqueness under sufficiently informative measurements~\cite{colton1998inverse}, and can exhibit posterior concentration with enough data.
Conversely, in common image-based inverse problems such as inpainting or super-resolution, the forward operator typically has a nontrivial null space and the posterior remains spread out even as the noise vanishes.

For instance, in a finite-dimensional setting under suitable regularity conditions, the Bernstein--von Mises theorem characterizes this concentration by an asymptotically Gaussian posterior with vanishing covariance centered around an estimator $\widetilde{m}(y)\to m^\dagger$:
\begin{equation}\label{eq:small_noise_limit}
    p(m\mid y) \approx \calN\bigg(\widetilde{m}(y), \sigma_1^2\calI(m^\dagger)^{-1}\bigg)
    \rightsquigarrow
    \delta_{m^\dagger}
    \quad\textrm{as}\quad
    \sigma_1\,\to\,0\,,
\end{equation}
where $\rightsquigarrow$ denotes weak convergence, $\calI(m^\dagger)$ is the Fisher information matrix and $\sigma_1^2$ is an effective posterior variance scale with $\sigma_1^2\sim \frac{1}{n}$ or $\sigma_1^2\sim\sigma_y^2$ for an increasing number of measurements $n$ or decreasing measurement noise $\sigma_y$, respectively.
This observation motivates us to examine solvers that work effectively in the regime of concentrated posterior distributions.

Although direct evaluation of the posterior is typically intractable, many algorithms aim to generate samples from it~\cite{stuart2010bayesian}.
A common class of methods is based on Markov chains constructed so that their invariant measure is the posterior distribution, as in Langevin Monte Carlo.
Unfortunately, sampling from the posterior can be computationally prohibitive, as each likelihood evaluation requires applying the forward map $\calF$.
Moreover, theoretical guarantees of efficient sampling for these algorithms often rely on assumptions that may not hold in practice.
Recently, generative models have been used to shift much of this computational cost to an offline training stage, enabling more efficient online reconstruction.

\subsection{Stochastic Interpolants}\label{sec:stochastic_interpolants}

The generative problem is to produce new samples from a distribution $\rho_{X}$ given only a dataset of samples $\{(x^i)\}_{i=1}^n$ drawn from $\rho_{X}$.
The stochastic interpolant (SI) framework~\cite{albergo2023si,albergo2025stochastic} solves this problem by defining an interpolation process $I_t$
which connects two endpoint distributions $\rho_0$ and $\rho_1$ and learning a map which transports samples from one endpoint to the other. 
Then, by choosing an easy-to-sample ``base distribution'' for $\rho_0$ and the target distribution $\rho_{X}$ for $\rho_1$, one can transform samples from $\rho_0$ into samples from $\rho_X$ by simply solving an ordinary differential equation (ODE) or stochastic differential equation (SDE) that realizes $I_t$.

Specifically, we define an interpolant
\begin{equation} \label{eq:si}
	I_t = \alpha_t x_0 + \beta_t x_1 + \gamma_t z, \qquad t\in[0,1],
\end{equation}
where $(x_0,x_1)\sim \rho(x_0, x_1)$ for a distribution $\rho(x_0, x_1)$ which has marginals $\rho_0$ and $\rho_1$, and $z\sim \calN(0, \sigma^2I)$ is independent of $(x_0,x_1)$. The coefficients must satisfy $\alpha_0=\beta_1=1$ and $\alpha_1=\beta_0=\gamma_0=\gamma_1=0$, be differentiable for $t\in(0,1)$, and be continuous for $t\in[0,1]$, but otherwise leave room for flexibility in how one designs the transport map.

In the SI paper~\cite{albergo2025stochastic} the authors use $\alpha_t=1-t$, $\beta_t=t$ and $\gamma_t=\sqrt{2t(1-t)}$, though different choices can produce other common generative models.
For example, a variance-preserving diffusion~\cite{ho2020ddpm,song2021scorebased} can be represented as $\alpha_t=\sqrt{1 - t^2}$ and $\beta_t=t$, and flow matching~\cite{lipman2023flow} as $\alpha_t=1-t$ and $\beta_t=t$.
Additionally, standard flow and diffusion models often choose $\gamma_t=0$ given that the base distribution $\rho_0$ is already chosen to be a Gaussian independent of $\rho_1$.

We then let $\rho_t=\operatorname{Law}(I_t)$ for $t\in(0,1)$.
Under suitable integrability and regularity assumptions,
these laws satisfy the continuity equation
\begin{equation}\label{eq:b_velocity}
    \partial_t\rho_t + \nabla\cdot(b_t\rho_t) = 0\,,\quad\text{with}\quad b_t(x) =\mathbb{E}[\dot{I}_t \mid I_t = x],
\end{equation}
where the expectation in $b$ is taken over both $(x_0,x_1)\sim\rho$ and $z\sim\calN(0,\sigma^2I)$. 
Equivalently, for $\epsilon_t\geq0$ and $\sigma^2\gamma_t>0$, the density satisfies
\begin{equation}\label{eq:bd_velocity}
    \partial_t\rho_t + \nabla\cdot((b_t-\epsilon_t(\sigma^2\gamma_t)^{-1}\calD_t)\rho_t) = \epsilon_t\Delta\rho_t\,,\quad\text{with}\quad \calD_t(x)=\mathbb{E}\big[z \mid I_t = x\big]\,.
\end{equation}
Thus, any sample drawn from $\rho_0$ and transported according to either~\eqref{eq:b_velocity} or~\eqref{eq:bd_velocity} generates a sample from $\rho_1$ at $t=1$.

To realize this in practice, one must learn\footnote{In what follows we use a `hat' to denote learned functions.} a velocity field $\hat{b}_t(x)\approx b_t(x)$ by solving a squared-error regression problem with the flow matching objective
\begin{equation}\label{eq:loss_cfm}
	\hat{b} := \operatorname*{argmin}_{b} \, \mathbb{E}_{t}\mathbb{E}_{x_0,x_1,z}\left[\left\|b_t(I_t) - \dot{I}_t\right\|_2^2\right],
\end{equation}
where the inner expectation is over training data $(x_0,x_1)$ and the independent noise $z$, and $\hat{b}$ is parametrized by a neural network. If SDE sampling is desired, a similar objective can be used to learn $\hat{\calD}_t(x)\approx \calD_t(x)$.

Once $\hat{b}$ and $\hat{\calD}_t$ are learned, we may generate samples from the target distribution by initializing $x_{t=0} \sim \rho_0$ and numerically integrating the probability flow ODE
\begin{equation}\label{eq:flow_ode}
    \dot{x}_t = \hat{b}_t(x_t)\,,
\end{equation}
or SDE
\begin{equation}\label{eq:flow_sde}
    dx_t = \left[\hat{b}_t(x_t)-\epsilon_t(\gamma_t\sigma^2)^{-1}\hat{\calD}_t(x_t)\right]dt + \sqrt{2\epsilon_t}\,dW_t\,,
\end{equation}
from $t=0$ to $t=1$, where $W_t$ is Brownian noise.

\begin{remark}\label{rem:conditional_si}
Extending the SI framework to conditional sampling is straightforward. Given a dataset of paired samples $\{(x^i,y^i)\}_{i=1}^n$ drawn from a joint distribution $\rho_{XY}(x, y)$, we seek to learn a generative model that, given new data $y$, produces samples from the conditional distribution $\rho_{X|Y}(x \mid y)$.
The SI formulation can be directly extended to this conditional setting by making all functions parameterized by an external variable $y$; for example, one would extend $b_t(x)$ to $b_t(x;y)=\mathbb{E}[\dot{I}_t \mid I_t = x; y]$. 
The training objective \eqref{eq:loss_cfm} is also modified so that the inner expectation is over paired training data $(x,y)$ in addition to $z$.
\end{remark}

\section{Methodology}
\label{sec:method}

In this section we introduce our proposed framework for \ourname, a generative model designed for scientific inverse problems in the concentrated posterior regime.
The key insight of \ourname is to combine an informative physics-based initialization with a velocity formulation suited to concentrated posterior distributions.
The adjoint of the forward map provides a physically meaningful initial estimate, while posterior concentration motivates concentrating the initial distribution around this estimate.
Since standard SI velocity fields can become ill-conditioned near concentrated endpoints, this construction also motivates a velocity formulation that remains regular in this regime.

We develop these components as follows.
In~Section~\ref{sec:adjoint_base}, we explore how to define an informative, physically relevant initial distribution by leveraging the adjoint of the forward map, including how posterior concentration motivates concentrating this initialization as well.
In~Section~\ref{sec:noise_free} we devise a velocity field which remains regular even with highly concentrated endpoints.
Section~\ref{sec:target_prediction} combines these two ideas into a cohesive generative method by leveraging a data-dependent target prediction parameterization that directly aligns the learning objective with the desired reconstruction.
Finally, Section~\ref{sec:red_flow} demonstrates how \ourname can incorporate inference-time scaling for improved physical consistency without retraining.

\subsection{Physics-Based Couplings with Stochastic Interpolants}\label{sec:adjoint_base}

Generative models for scientific inverse problems look to sample from the conditional posterior distribution $\rho_1=p(m\mid y)$ by learning a conditional velocity field $\hat{b}_t(x; y)$ parameterized by measurements $y$, as detailed in~Remark~\ref{rem:conditional_si}.
While the standard convention is to define an independent Gaussian reference distribution $\rho_0=\calN(0, I)$, the SI framework~\eqref{eq:si} is not restricted to this choice: the endpoint distributions $\rho_0$ and $\rho_1$ may be arbitrary, and even coupled~\cite{albergo2024couplings}.
Therefore, taking motivation from work showing that optimal-transport-based couplings can simplify learned flows and improve training and inference efficiency~\cite{tong2024improving,pooladian2023multisampleflowmatchingstraightening}, we investigate how the design of $\rho_0$ can affect the transport distance between endpoints, and what choices can minimize it.

The paired-data setting inherent to inverse problems provides a natural correspondence from which to construct endpoint distributions: each training sample provides measurements $y$ together with the corresponding unknown parameters $m$. 
However, the domains are mismatched:
$y\in\calY$ and $m\in\calM$ generally occupy different spaces and cannot be directly interpolated. 
A physics-based reconstruction map resolves this mismatch by converting each measurement $y$ into an approximation of $m$ in parameter space $\calM$.

A natural choice -- if the problem is linear or admits an efficient linearization -- is to use the adjoint $F^*y$\footnote{Or the regularized pseudoinverse $F^\dagger_\epsilon=(F^* F+\epsilon I)^{-1}F^*$, if one can be computed efficiently.} to approximate $\calF^{-1}$ and map measurements $y$ into $\calM$. 
Specifically, one may use~\eqref{eq:linearization} to define a physics-based initialization
\begin{equation}\label{eq:reference_initializer}
    a(y)=\overline{m}+F^*\big(y-\calF(\overline m)\big),
\end{equation}
where $\overline{m}$ is a pre-specified background medium.
In the case of a homogeneous background $\overline{m}=0$ with $\calF(0)=0$, the initialization reduces to $a(y) = F^*y$.
Henceforth, we use $a(y)$ to denote an inexpensive physics-based approximation used to initialize the interpolant, even if it may not directly correspond to the adjoint operator (see, for example,~Section~\ref{sec:black_hole_imaging}).

Though $a(y)$ is not the exact inversion, it contains physically meaningful information. 
We may therefore hypothesize that $a(y)$ can serve as an informative center to the reference distribution, for example, by defining $\rho_0=\calN(a(y),\,\sigma_0^2I)$.
We are further left with the question of how to concentrate $\rho_0$ around $a(y)$, and may naturally hypothesize that the optimal concentration of $\rho_0$ should match the concentration of $\rho_1$.

To aid our analysis, we first introduce the Wasserstein-2 distance, a cornerstone of optimal-transport theory which measures the minimum root-mean-square Euclidean displacement required to transport one distribution into the other.
The distance is defined as
\begin{equation}\label{eq:wasserstein2}
    \mathcal{W}_{2}(\rho_0,\rho_1)=\left(\inf _{\gamma \in \Pi (\rho_0 ,\,\rho_1 )}\mathbb{E}_{(x_0,x_1)\sim \gamma }\left\|x_0-x_1\right\|_2^2\right)^{1/2},
\end{equation}
where $\Pi(\rho_0,\rho_1)$ is the set of all couplings of the distributions $\rho_0$ and $\rho_1$.

To decompose the analysis of the transport distance between endpoints into differences of centers and spreads, we leverage the following property of the $\mathcal{W}_2$ distance: for any distributions $\rho_0$ and $\rho_1$ with finite second moments and means $\mu_0$ and $\mu_1$, 
\begin{equation}\label{eq:wasserstein_source_translation}
    \mathcal W_2^2(\rho_0,\rho_1)=\|\mu_0-\mu_1\|_2^2+ \mathcal W_2^2(\overline\rho_0,\overline\rho_1),
\end{equation}
where $\overline\rho_0$ and $\overline\rho_1$ are the shifted versions of $\rho_0$ and $\rho_1$ which have zero mean.
This result comes immediately from examining~\eqref{eq:wasserstein2}, since the act of translation induces a bijection between couplings and the cross term in $\mathbb{E}\|x_0-x_1\|_2^2$ vanishes once the distributions have zero mean.

Using this fact, we now demonstrate that centering the reference distribution around a physics-based approximation can reduce the transport distance to the target posterior relative to a mean-zero reference distribution (such as a standard Gaussian initialization).
We leave the proofs to~\ref{app:proofs}.

\begin{proposition}\label{prop:reduced_transport_case_linear}
Let $y = Fm + \varepsilon$ for a linear forward operator $F$ with $\varepsilon \sim \mathcal{N}(0,\sigma_y^2 I)$ and a flat prior on $m \in \mathbb{R}^d$, and suppose $r = \|F^*F - I\|_{\mathrm{op}} < 1$. 
Furthermore, with $\mu = (F^*F)^{-1}F^*y$ the mean of the posterior $\rho_1 = p(m \mid y)$, take any zero-mean distribution $\rho_0$ with finite second moment, and $\rho_0^{\mathrm{adj}}$ its translation by $F^*y$. Then
\begin{equation}
    \mathcal W_2^2(\rho_0,\rho_1) - \mathcal W_2^2(\rho_0^{\mathrm{adj}},\rho_1)
    \geq (1-r^2)\|\mu\|_2^2,
\end{equation}
so $\mathcal W_2(\rho_0^{\mathrm{adj}},\rho_1) < \mathcal W_2(\rho_0,\rho_1)$ whenever $F^*y \neq 0$.
\end{proposition}

We can extend this result to nonlinear operators by leveraging the linearization in~\eqref{eq:linearization}.
Namely, we suppose $\mathcal F$ is differentiable at a fixed background $\overline m$ with Jacobian $F = \frac{\partial\mathcal F}{\partial m}(\overline m)$, and reformulate the problem in the perturbation coordinates
\begin{equation}
    \delta m = m - \overline m, \qquad
    \widetilde y = y - \mathcal F(\overline m),
\end{equation}
where $\rho_1$ is now the posterior over $\delta m$ (assumed to have finite second moment and mean $\mu$).

\begin{proposition}[Nonlinear extension]\label{prop:reduced_transport_case_nonlinear}
Suppose $\mathcal F$ has $L$-Lipschitz Jacobians and $F$ satisfies $r = \|F^*F - I\|_{\mathrm{op}} < 1$. 
Suppose further that there exists an unconstrained least-squares solution $m^\star=\operatorname{argmin}\|y - \mathcal F(m)\|_2^2$ with residual $e = y - \mathcal F(m^\star)$. Defining $\delta^\star = m^\star - \overline m$ and $\tau = \|\mu - \delta^\star\|_2$, if
\begin{equation}\label{eq:nonlinear_assumption}
    \tfrac L2\|F\|_{\mathrm{op}}\|\delta^\star\|_2^2 + L\|e\|_2\|\delta^\star\|_2 + 2\tau < (1-r)\|\delta^\star\|_2,
\end{equation}
then for any zero-mean $\rho_0$ with finite second moment and $\rho_0^{\mathrm{adj}}$ its translation by $F^*\widetilde y$, we have $\mathcal W_2(\rho_0^{\mathrm{adj}},\rho_1) < \mathcal W_2(\rho_0,\rho_1)$.
\end{proposition}

Each term in the inequality~\eqref{eq:nonlinear_assumption} has a clear meaning: 
$2\tau$ accounts for the offset between the posterior mean and the least-squares solution,
$\tfrac L2\|F\|_{\mathrm{op}}\|\delta^\star\|_2^2$ is the Taylor remainder of the linearization\footnote{In wave-based problems this residual represents multiple scattering effects, and~\eqref{eq:born_validity_condition} can be interpreted as a Born-type validity condition (see~Section~\ref{sec:nonlin_frequency_inverse_scattering}).},
$L\|e\|_2\|\delta^\star\|_2$ couples the least-squares residual to the nonlinearity, 
and $(1-r)\|\delta^\star\|_2$ is the margin bought by the adjoint's accuracy as an inverse.

In particular, in the identifiable, stable small-noise regime
described in~Section~\ref{sec:bayesian}, with uniform integrability of $\|m\|$ under the posterior distributions, we have $m^\star \to m^\dagger$, $e \to 0$, and $\tau \to 0$; the sufficient condition then becomes
\begin{equation}\label{eq:born_validity_condition}
    \|m^\dagger-\overline m\|_2 < \frac{2(1-r)}{L\|F\|_{\mathrm{op}}}.
\end{equation}

\begin{remark}\label{rem:transport_scope}
The bound $r<1$ is a scale-dependent condition which depends on the units of $F^*$. In practice we input a normalized $a(y)$ to the neural network, which fixes the choice of units, and only the condition number $\kappa$ of $F^*F$ remains intrinsic. For ill-conditioned operators where $\kappa\gg1$, the guaranteed margin $1-r =\mathcal{O}(\kappa^{-1})$ is small. However, $r<1$ is a sufficient but not necessary condition: the proofs only use bounds on $F^*F - I$ along relevant directions, which can hold even when $r \geq 1$ if these directions are well-conditioned.
\end{remark}

Having addressed where to center $\rho_0$, we now ask how concentrated it should be.
We show that to minimize the transport distance in the case of Gaussian endpoints, $\rho_0$ should aim to match the spread of $\rho_1$.

\begin{proposition}
\label{prop:reduced_transport_case_concentrated}
Assume a $d$-dimensional isotropic Gaussian target
$\rho_1=\calN(\mu_1,\sigma_1^2I)$ and reference distribution
$\rho_0=\calN(\mu_0,\sigma_0^2I)$. Then
\begin{equation}
    \mathcal W_2^2(\rho_0,\rho_1)
    =
    \|\mu_0-\mu_1\|_2^2
    +
    d(\sigma_0-\sigma_1)^2,
\end{equation}
which, for fixed $\mu_0$, is minimized at $\sigma_0=\sigma_1$.
Consequently, in the singular-endpoint limit $\sigma_1\to0$ motivated by~\eqref{eq:small_noise_limit}, the optimal base variance also vanishes. In particular, when $\sigma_1=0$, the minimum is attained by $\rho_0=\delta_{\mu_0}$.
\end{proposition}
\begin{proof}
This result can be quickly inferred from the Wasserstein-2 formula for Gaussians:
\begin{equation}\label{eq:wasserstein2_gaussian}
\mathcal W_2^2(\calN(a,\Sigma_A),\calN(b,\Sigma_B))
=\|a-b\|_2^2+\operatorname{tr}\!\left(\Sigma_A+\Sigma_B-2(\Sigma_A^{1/2}\Sigma_B\Sigma_A^{1/2})^{1/2}\right).
\end{equation}
With isotropic Gaussian endpoints, this becomes $\|\mu_0-\mu_1\|_2^2 + \operatorname{tr}\big((\sigma_0-\sigma_1)^2I\big)=\|\mu_0-\mu_1\|_2^2+d(\sigma_0-\sigma_1)^2$ when $I$ is the $d\times d$ identity.
\end{proof}

These conclusions suggest that, if we suspect a concentrated posterior distribution, a singular reference distribution $\rho_0=\delta_{a(y)}$ -- e.g. using an adjoint-based approximation $a(y)=F^*y$ -- can provide an initialization that is geometrically better aligned to the target than a standard Gaussian distribution.

We verify this numerically by computing Wasserstein-2 distances for all datasets considered in this paper. 
As reported in~Table~\ref{table:wasserstein}, the concentrated, physics-based reference $\delta_{a(y)}$ has a smaller transport distance than a standard Gaussian on every dataset considered (see~\ref{app:transport_distances} for more details).
We additionally examine empirically how transport distance relates to reconstruction accuracy in our ablations (see~Table~\ref{table:init_ablation}).

\subsection{A Noise-Free Velocity Field for Concentrated Endpoints}\label{sec:noise_free}

The second component of \ourname is a modified SI velocity field designed to remain regular even with singular endpoints.
Our construction is motivated by the concentrated-posterior regime~\eqref{eq:small_noise_limit}, which also favors concentrating the physics-based initialization, as shown in~Proposition~\ref{prop:reduced_transport_case_concentrated}.
However, concentration at either endpoint can introduce difficulties for the standard SI velocity field.
The following proposition makes this precise: when either endpoint distribution collapses to a point mass, the spatial Lipschitz constant cannot remain uniformly bounded near that endpoint.

\begin{proposition}\label{prop:analytic_b}
    Suppose 
    \begin{equation}
        x_0\sim\rho_0=\calN(\mu_0, \sigma_0^2I),\qquad x_1\sim \rho_1=\calN(\mu_1, \sigma_1^2\Sigma_1), \qquad z\sim\calN(0,\sigma^2I),
    \end{equation}
    are independent Gaussians and the interpolant covariance $C_t=\alpha_t^2\sigma_0^2I+\beta_t^2\sigma_1^2\Sigma_1+\gamma_t^2\sigma^2I$ satisfies $C_t\succ0$ for $t\in(0,1)$. Then, for any $t\in(0,1)$ the SI velocity field \eqref{eq:b_velocity} admits an analytic form
    \begin{equation}\label{eq:analytic_b}
        b_t(x) = \dot{\mu}(t) + K_b(t)\big(x-\mu(t)\big)
        \qquad\textrm{with}\qquad
        \mu(t) := \alpha_t\mu_0 + \beta_t\mu_1,\qquad
        K_b(t) := \tfrac{1}{2}\dot{C}_tC^{-1}_t,
    \end{equation}
    where $C_t=\operatorname{Cov}(I_t)$.
    In the singular-endpoint limit~\eqref{eq:small_noise_limit} where $\sigma_1=0$, $K_b(t)$ becomes unbounded as $t\uparrow1$; thus, $b_t$ fails to remain uniformly Lipschitz in $x$ near the target endpoint. Namely,
    \begin{equation}\label{eq:ill_conditioned_b}
        \limsup_{t\uparrow1}\;\|\nabla_xb_t\|=
        \limsup_{t\uparrow1}\;\|K_b(t)\|=\infty.
    \end{equation}
    Similarly, if $\sigma_0=0$, the spatial Lipschitz constant cannot remain uniformly bounded near the base endpoint, as
    \begin{equation}
        \limsup_{t\downarrow0}\;\|\nabla_xb_t\|=
        \limsup_{t\downarrow0}\;\|K_b(t)\|=\infty.
    \end{equation}
\end{proposition}

The unbounded Lipschitz constant identified by~Proposition~\ref{prop:analytic_b} creates spatial sensitivity that one would like to numerically avoid.
Since $b_t(\hat{x}_t)-b_t(x_t)=K_b(t)(\hat{x}_t-x_t)$, small state errors from learned approximation or numerical discretization can produce large velocity errors, potentially complicating learning and requiring smaller integration steps.
The behavior along exact interpolant trajectories also depends on the interpolant: for a linear interpolant, velocities remain bounded as $t\to1$ along exact interpolant trajectories, yet they diverge for a variance-preserving diffusion interpolant (see~\ref{sec:analytic_b}).
Another common strategy known as \textit{early stopping} avoids the endpoint singularity by truncating the integration at a fixed $\underline{t} < 1$, yet this generally introduces bias by terminating transport before fully reaching the target distribution.

While the preceding examples concern transporting a nondegenerate source to a singular target, Proposition~\ref{prop:reduced_transport_case_concentrated} suggests a more natural design in this regime: choosing a concentrated source as well.
However, with deterministic endpoints and no bridge noise, training states collapse onto the trajectory $I_t=\mu(t)$, leaving the learned field unconstrained away from it.
Adding bridge noise $I_t=\mu(t)+\gamma_tz$ gives the training distribution full support at interior times, but the standard SI velocity must generate and remove this noise artificially, leading to the endpoint ill-conditioning identified in~Proposition~\ref{prop:analytic_b}.
To retain noise augmentation during training while separating its contribution to the velocity, we use the standard decomposition
\begin{equation}\label{eq:velocity_decomposition}
    b_t(x) = v_t(x) + \dot{\gamma}_t\calD_t(x),
\end{equation}
where $\calD_t$ is as in~\eqref{eq:bd_velocity} and we term
\begin{equation}\label{eq:noise_free_velocity}
    v_t(x)=\mathbb{E}\big[\dot{\alpha}_t x_0 + \dot{\beta}_t x_1 \mid I_t = x\big]
\end{equation} 
the \textit{noise-free velocity} since the noise bridge term $\dot{\gamma}_tz$ has been removed from the expectation.
The following corollary establishes spatial regularity of the noise-free velocity for singular endpoints and, under suitable interpolant schedules, exact recovery of the target mean even when the target has nonzero variance.

\begin{corollary}\label{cor:analytic_v}
    Under the same assumptions as Proposition~\ref{prop:analytic_b}, the noise-free velocity field $v_t$ admits an analytic form
    \begin{equation}
        v_t(x) = \dot{\mu}(t) + K_v(t)\big(x-\mu(t)\big)
        \qquad\textrm{with}\qquad
        K_v(t) := \bigg(\dot{\alpha}_t \alpha_t \sigma_0^2I + \dot{\beta}_t \beta_t \sigma_1^2\Sigma_1\bigg)\, C^{-1}_t,
    \end{equation}
    where $\mu(t)$ and $C_t$ are as in \eqref{eq:analytic_b}.
    In the singular-endpoint limit~\eqref{eq:small_noise_limit} with $\sigma_1=0$, if $\sigma_0=0$ and $\sigma>0$ then $K_v\equiv0$ and $v_t(x) = \dot{\mu}(t)$, which has uniform spatial regularity since $\nabla_xv_t=0$. 
    Moreover, if $\sigma_0=0$, $\sigma>0$, and $\Sigma_1\succ0$, choosing $\alpha_t=1-t$, $\beta_t=t$, and $\gamma_t=\sqrt{t(1-t)}$ ensures that the noise-free ODE $\dot{x}_t=v_t(x_t)$ initialized at $x_0=\mu_0$ uniquely recovers the target mean $x_1=\mu_1$.
\end{corollary}

Corollary~\ref{cor:analytic_v} thus motivates using the noise-free velocity $v_t$ with a deterministic source ($\sigma_0=0$) and nonzero bridge noise ($\sigma>0$).
The proofs of Proposition~\ref{prop:analytic_b} and Corollary~\ref{cor:analytic_v} are left to \ref{app:proofs}, and leverage the conditional expectation formula for jointly Gaussian random variables as their primary tool.

\begin{remark}
    In practice, the combination of a deterministic reference distribution with the noise-free velocity yields an iterative, but \emph{deterministic}, reconstruction method.
    Namely, for fixed measurements this method produces a single point reconstruction rather than a sample from a non-degenerate posterior.
    Randomness enters only through time-dependent noise augmentation during training.    
    A similar training strategy appears in~\cite{reu2025gradient}, which adds noise-augmented states to a standard flow-matching loss to improve generalization of flow-based generative models.
    Here, we derive this construction from the regularity of interpolants with concentrated endpoints.
\end{remark}

\subsection{Data-Dependent Target-Prediction}\label{sec:target_prediction}
We now combine the main ideas from Section~\ref{sec:adjoint_base} and Section~\ref{sec:noise_free} into a cohesive framework. 
To attain a base distribution geometrically aligned with the target,~Proposition~\ref{prop:reduced_transport_case_nonlinear} and~Proposition~\ref{prop:reduced_transport_case_concentrated} suggest choosing a singular base distribution located at a physically informative initialization.
If the forward model can be linearized efficiently, the adjoint typically provides a strong initial guess and a natural initialization is $\delta_{F^*y}$.
Corollary~\ref{cor:analytic_v} then shows that to generate samples when the endpoint distributions are singular, the noise-free velocity $v_t$ could be numerically advantageous because it recovers regularity where the standard SI velocity $b_t$ lacks it.

Hence, we define \textbf{\ourname} as the reconstruction method which solves the ODE
\begin{equation}\label{eq:flow_ode_adjointflow}
    \dot{x}_t = v_t(x_t;y), \qquad x_{t=0} = a(y),
\end{equation}
where $a(y)$ denotes a physics-based approximation, with the notable example being $a(y)=F^*y$ when an efficient linearization of the forward model is available.
As $x_0$ is now a deterministic function of $y$, the noise-free velocity naturally decomposes as
\begin{equation}
    v_t(x; y) =
    \mathbb{E}\big[\dot{\alpha}_t x_0 + \dot{\beta}_t x_1 \mid I_t = x;y\big]
    = \dot{\alpha}_t \, a(y) + \dot{\beta}_t\mathbb{E}\big[ x_1 \mid I_t = x;y\big].
\end{equation}
This expression motivates a \textit{data-dependent target-prediction} (DD-TP) parameterization of the velocity, where a network is trained to predict the \emph{clean target} -- a ``denoiser-style'' objective -- rather than the tangential velocity directly.
Specifically, we learn the target-prediction network
\begin{equation}\label{eq:target_prediction_network}
    \hat{m}_t(x; y) \;\approx\; \mathbb{E}[x_1 \mid I_t=x\,;y]
\end{equation}
and define the velocity as
\begin{equation}\label{eq:factorized_flow}
    \hat{v}_t(x; y) \;=\; \dot{\alpha}_t \, a(y) \,+\, \dot{\beta}_t\hat{m}_t(x; y).
\end{equation}

Although target and velocity prediction are equivalent on a theoretical level, they induce different finite-capacity hypothesis classes. 
Multiple recent works have argued for TP~\cite{li2025letdenoise, han2026selfconsistent} on the basis that it has a more stable training signal than velocity prediction.
Intuitively, a TP parameterization allows the network to focus on refining the residual not already explained by the adjoint approximation.
Moreover, as $t\to1$, target prediction approaches an identity map, whereas velocity prediction must continue to represent the measurement-dependent displacement from $a(y)$.

For \ourname, target prediction is particularly natural because the initialization $x_0(y)=a(y)$ is deterministic and exact.
In contrast, the standard TP is typically derived for a linear interpolant $I_t = (1-t)x_0 + tx_1$ with $x_0\sim\calN(0,I)$ by eliminating the unknown $x_0$ to get
\begin{equation}\label{eq:uncoupled_tp} 
\hat{b}_t(x;y)=\frac{\hat{m}_t(x;y)-x}{1-t}.
\end{equation} 
While the data-dependent TP directly hard-codes the physical source into velocity, standard TP must reconstruct the velocity indirectly which introduces an undesirable $(1-t)^{-1}$ factor.

\begin{wrapfigure}{r}{0.55\linewidth}
    \centering
    \includegraphics[width=\linewidth]{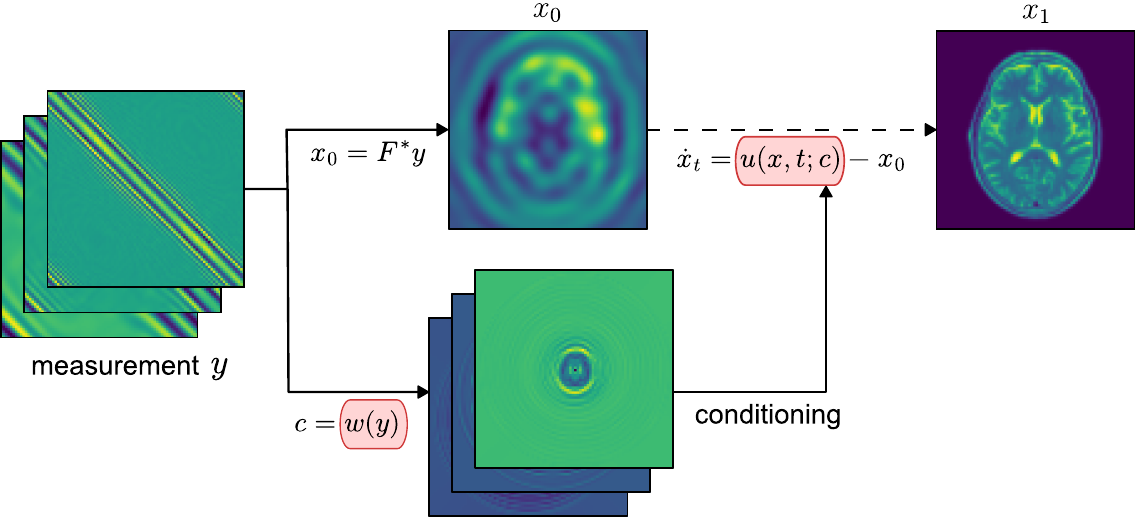}
    \caption{\textbf{Model architecture.}
    A data encoding network $w$ extracts domain-specific features $c=w(y)$ to condition the target-prediction network $u(x,t;c)$, which then forms a velocity $\dot{x}_t = u(x, t; c) - x_0$ to be integrated, with initial state $x_0=a(y)$ at $t=0$, to produce an estimate $x_1\approx m^\dagger$ at $t=1$. Red boxes denote learned components.}
    \label{fig:architecture}
\end{wrapfigure}

We also emphasize that while~\eqref{eq:target_prediction_network} superficially resembles a direct reconstruction network, its role in~\eqref{eq:factorized_flow} leads to a different learning problem: target prediction is supervised across time-dependent, noise-perturbed intermediate states and composed iteratively at inference.
TP thus incorporates mechanisms such as supervised iterative computation and training-time stochastic regularization, which have been associated with improved performance over single-step regression in recent work~\cite{agrawalla2026floq,pan2026adonoisingdispellingmyths}, even when inference produces a deterministic prediction.

The training and inference algorithms for \ourname are provided in Algorithm~\ref{alg:adjointflow_train} and Algorithm~\ref{alg:adjointflow_inference}. 
Even with the modifications to the standard SI framework, the implementation remains straightforward. 
We use the interpolant $I_t = (1-t)x_0 + tx_1 +\sqrt{t(1-t)}z$; the empirical performance of other choices of $\gamma_t$ is investigated in Section~\ref{sec:ablations}.
We parameterize the TP network as $\hat{m}_t(x; y) = u_\theta(x, t; w_\phi(y))$ where $u_\theta$ is the primary target-prediction network and $w_\phi$ is a bespoke data-encoding model designed to process the measurements. Figure~\ref{fig:architecture} provides an illustration of this architecture.

\begin{algorithm}[tb]
   \caption{\ourname Training}
   \label{alg:adjointflow_train}
\begin{algorithmic}
   \STATE {\bfseries Input:} paired dataset $D=\{(y^i, m^i)\}_{i=1}^{|D|}$; physics-based approximate inverse function $a$; untrained target-prediction network $u_\theta$; untrained data encoding network $w_\phi$; batch size $n_b$; bridge noise level $\sigma$.
   \REPEAT
      \FOR{$i = 1, \dots, n_b$}
         \STATE Sample pair $(y^i, m^i)\sim D$, time $t_i \sim U(0,1)$, and noise $z_i \sim \mathcal{N}(0, \sigma^2I)$.
         \STATE Compute the physics-based initialization $x_0^i = a(y^i)$.
         \STATE Define the target $x_1^i = m^i$.
         \STATE Compute the interpolant $x_{t_i}^i = (1-t_i) x_0^i + t_i x_1^i + \sqrt{t_i(1-t_i)} \,z_i$.
      \ENDFOR
      \STATE Compute empirical target-prediction loss
      \[ \hat{L}(\theta, \phi) = \frac{1}{n_b} \sum_{i=1}^{n_b} \bigg\| u_\theta(x_{t_i}^i, t_i; w_\phi(y^i)) - x_1^i\bigg\|^2.\]
      \STATE Take a gradient step on $\hat{L}(\theta, \phi)$ to update $\theta$ and $\phi$.
   \UNTIL{converged}
   \STATE {\bfseries Return:} Trained target-prediction model $u_\theta$ and data encoder $w_\phi$.
\end{algorithmic}
\end{algorithm}

\begin{algorithm}[tb]
   \caption{\ourname Inference (Forward Euler)}
   \label{alg:adjointflow_inference}
\begin{algorithmic}
   \STATE {\bfseries Input:} measurement data $y$; trained models $u_\theta$ and $w_\phi$; physics-based approximate inverse function $a$; integration steps $N \in \mathbb{N}$.
   \STATE Initialize $x_0 = a(y)$.
   \FOR{$n = 0, \dots, N-1$}
      \STATE Let $\Delta t=N^{-1}$ and $t_n = n\Delta t$.
      \STATE Compute $\hat{v}_{t_n}(x_{t_n}; y) = u_\theta(x_{t_n}, t_n; w_\phi(y)) - x_0$.
      \STATE Update $x_{t_{n+1}} = x_{t_n} + \Delta t \,\,\hat{v}_{t_n}(x_{t_n}; y)$.
   \ENDFOR
   \STATE {\bfseries Return:} prediction $x_1$.
\end{algorithmic}
\end{algorithm}

\subsection{Inference-Time Scaling}\label{sec:red_flow}
Inference-time scaling is a common strategy in generative modeling for improving output quality without retraining the generative model.
The strategy allocates additional computational resources at inference time, for example by introducing an additional guidance term or by recasting generation as a search or optimization problem.
Plug-and-play (PnP) algorithms are a prominent example of inference-time scaling which apply guidance to an unconditional diffusion model trained to sample from a prior distribution~\cite{zheng2025inversebench}.
Although often computationally expensive due to repeated evaluations of the forward or adjoint map, similar to classical optimization algorithms, these methods can effectively incorporate physics and measurement information into the generative transport and produce high-quality reconstructions.

Fortunately, the \ourname framework is also compatible with inference-time scaling. 
In particular, physics-based guidance can further improve reconstruction quality and measurement consistency at the expense of increased computational cost.
To this end, we adapt the RED-Diff algorithm \cite{mardani2024reddiff} to the stochastic interpolant framework. 
We refer to the resulting algorithm as \textbf{\ournameRED}.

RED-Diff recasts the generation problem as a variational inference problem. 
Inference is performed by minimizing an objective that balances the data misfit with a denoising-score-matching term that acts as a regularizer.
Following this formulation, we replace the score-based regularization with an analogous flow-matching term which yields the following optimization problem:
\begin{equation} \label{eq:red_flow}
    \min_{m} \quad \underbrace{\|y-\mathcal{F}(m)\|^2}_{\text{data misfit}}
     + \underbrace{\mathbb{E}
    \left[\frac{\lambda_t}{\dot{\beta}_t}
    \textbf{sg}\big[\dot{x}_t - \hat{v}_t(x_t; y)\big]^\top m
    \right]}_{\text{regularization}},
\end{equation}
where $x_t=\alpha_t x_0 + \beta_t m$, $\dot x_t=\dot\alpha_tx_0+\dot\beta_tm$, $\lambda_t$ is a weighting schedule, $\hat{v}_t$ is given by \eqref{eq:factorized_flow}, and \textbf{sg} is the stop-gradient operator.
Since $\dot{x}_t - \hat{v}_t = \dot{\beta}_t(m-\hat{m}_t)$ due to the target-prediction parameterization, the regularization adds a gradient term $\lambda_t(m-\hat{m}_t)$ which pulls the iterate toward the learned reconstruction while the misfit term enforces data consistency.
We initialize the reconstruction $m$ to the approximation $a(y)$, then solve \eqref{eq:red_flow} using Adam~\cite{kingma2015adam}. 

By explicitly incorporating the forward model into inference through the data-misfit term, \ournameRED can improve the physical consistency of the reconstruction, albeit at increased computational cost due to repeated evaluations of $\calF$.
Further details are provided in \ref{app:inference_scaling}, including the full algorithm in Algorithm~\ref{alg:red_flow}.

\section{Experiments}
\label{sec:experiments}

We investigate the performance of \ourname and \ournameRED across four diverse inverse problems from scientific domains, each with their own formulations, modalities, and challenges: Nonlinear Frequency-Domain Inverse Scattering (Section~\ref{sec:nonlin_frequency_inverse_scattering}), Linear Frequency-Domain Inverse Scattering (Section~\ref{sec:lin_frequency_inverse_scattering}), Time-Domain Seismic Imaging (Section~\ref{sec:time_domain_seismic_imaging}), and Black Hole Imaging (Section~\ref{sec:black_hole_imaging}). 

We compare against a range of state-of-the-art methods including classical optimization, deterministic ML models, conditional diffusion models (CDM), and plug-and-play diffusion prior (PnP-DP) methods, evaluating performance on a variety of accuracy metrics as well as on inference cost.
Our evaluation metrics, detailed in~\ref{app:metrics}, aim to match the evaluation criteria of the benchmarks we test against. 
We report quantitative baseline results directly from the respective benchmark papers~\cite{Zhang2025_BPD} or~\cite{zheng2025inversebench} unless otherwise stated. We recompute all reported Sinkhorn divergences using a common library version, since updating the library substantially changed the numerical values. We generate qualitative baseline reconstructions locally using the corresponding implementations and pretrained checkpoints, and plot absolute errors with respect to the ground truth. Unless otherwise stated, \ourname uses 8 forward-Euler steps (Algorithm~\ref{alg:adjointflow_inference}) and CDM models use 256 forward-Euler-Maruyama steps.

\subsection{Nonlinear Frequency-Domain Inverse Scattering}\label{sec:nonlin_frequency_inverse_scattering}
In domains such as medical imaging and nondestructive testing, a common task is to infer the properties, shape, or interior characteristics of an unknown object or medium by analyzing the scattering patterns of impinging waves. To this end, the unknown medium is probed with monochromatic (single-frequency) waves, with the resulting impulse responses, in the form of scattered waves, recorded at the boundary of the domain (see Figure~\ref{fig:imaging_regimes} left). This is known as the \textit{inverse scattering problem}.

We consider the wideband inverse scattering problem in a two-dimensional acoustic setting.
The physical model of time-harmonic wave propagation is provided by the Helmholtz equation,
\begin{equation}\label{eqn:Helmholtz}
\Delta u(x) + \omega^2 n(x) u(x) = 0,
\qquad x\in\mathbb{R}^2,
\end{equation}
where $u$ is the total wave field, $\omega$ is the wavenumber of the probing wave, and $n(x)$ is the squared refractive index which governs the speed at which a wave travels through the medium.
We bound the medium inside a domain $\Omega\subseteq\mathbb{R}^2$ and assume a homogeneous vacuum of $n(x)=1$ as the background.
The natural choice for the media of interest then becomes the refractive index perturbation $m(x) = n(x)-1$ to have $\operatorname{supp}(m) \subseteq \Omega$.

\paragraph{Forward problem}
The medium is illuminated by a monochromatic incident plane wave
\begin{equation}
u^{\text{in}}(x;s) = e^{\mathrm{i}\omega s \cdot x}, \qquad s \in \mathbb{S},
\end{equation}
where $s$ denotes the direction of incidence and $u(\cdot;s)$ denotes a wave field induced by such an incident wave.
The resulting total wave field thus decomposes as $u = u^{\text{in}} + u^{\text{sc}}$, with the scattered wave $u^{\text{sc}}$ satisfying:
\begin{equation}\label{eq:scattered_wavefield}
\begin{cases}
    \Delta u^{\text{sc}}(x) + \omega^2 \big(1+m(x)\big) u^{\text{sc}}(x) = -\omega^2 m(x) u^{\text{in}}(x),\\
    \frac{\partial u^{sc}}{\partial|x|}-i\omega u^{sc}=\mathcal{O}\big(|x|^{-3/2}\big)\,\textrm{ uniformly as }\,|x|\to\infty,
\end{cases}
\end{equation}
where the second equation ensures uniqueness via the Sommerfeld radiation condition.
Measurements are collected from receivers placed at angles $r \in \mathbb{S}$ on a circle of radius $R$.
The recorded far-field measurements at frequency $\omega$ are then denoted as
\begin{equation}
y_\omega(r,s) = u^{\text{sc}}(Rr;s)\,\in\mathbb{C},\qquad (r,s)\in\mathbb{S}\times\mathbb{S} .
\end{equation}
The forward map, which associates a perturbation $m$ with its corresponding scattered field data, is defined as $\mathcal{F}_\omega(m)=y_\omega(r, s)$.

\begin{figure*}[t]
    \centering

    \begin{subfigure}[b]{0.24\textwidth}
        \centering
        \includegraphics[width=\linewidth]{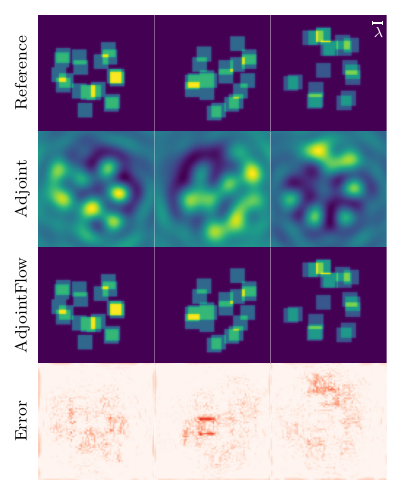}
        \caption{}
        \label{fig:sub-10hsquares}
    \end{subfigure}
    \hfill
    \begin{subfigure}[b]{0.226\textwidth}
        \centering
        \includegraphics[width=\linewidth]{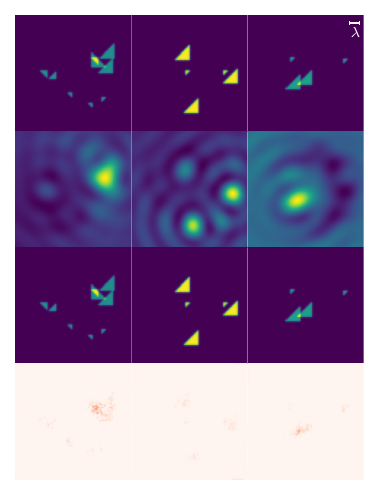}
        \caption{}
        \label{fig:sub-3510tri}
    \end{subfigure}
    \hfill
    \begin{subfigure}[b]{0.226\textwidth}
        \centering
        \includegraphics[width=\linewidth]{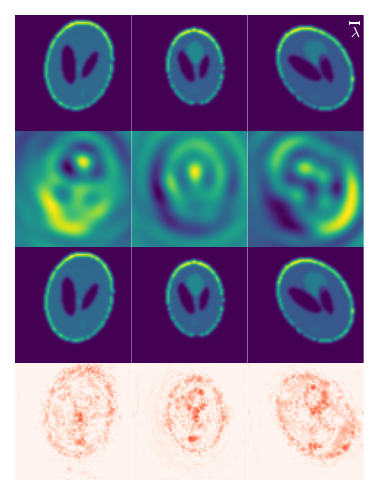}
        \caption{}
        \label{fig:sub-shepp-logan}
    \end{subfigure}
    \hfill
    \begin{subfigure}[b]{0.258\textwidth}
        \centering
        \includegraphics[width=\linewidth]{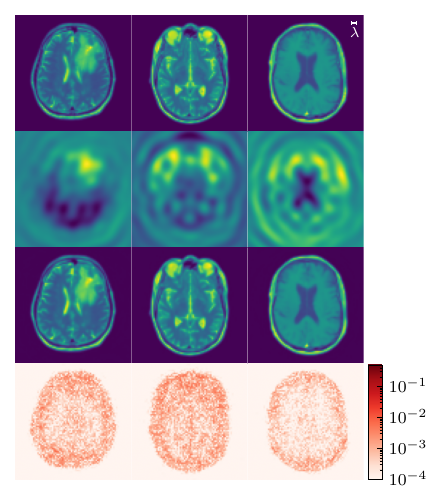}
        \caption{}
        \label{fig:sub-brain-scan}
    \end{subfigure}

    \caption{Qualitative results on nonlinear inverse scattering
    for (a) \textit{10h Overlapping Squares},
    (b) \textit{3--5--10h Triangles},
    (c) \textit{Shepp--Logan}, and (d) \textit{Brain Scans}.
    Within each panel, columns show distinct test examples;
    rows show, from top to bottom, the ground-truth perturbation,
    the adjoint-based initialization, the \ourname reconstruction,
    and its absolute reconstruction error. The wavelength marker corresponds
    to 8 points per wavelength (PPW) in (a)--(c) and 5 PPW in (d).}
    \label{fig:four_panels}
\end{figure*}

\paragraph{Challenges}
A major challenge of the inverse problem is \emph{multiple scattering effects}: 
The solution to the inhomogeneous Helmholtz equation \eqref{eq:scattered_wavefield} admits the Born expansion
\begin{equation}\label{eq:born_series}
    y = A m + B(m,m)+C(m,m,m)+\cdots,
\end{equation}
revealing a series of higher-order nonlinear terms which entangle interactions and make it difficult to know where a signal originated.

As a result, attempting to infer $m$ from single-frequency data is a challenging, highly unstable and nonconvex problem that can lead to \emph{cycle-skipping}: a phenomenon where optimization algorithms become trapped in local minima, yielding spurious, non-physical reconstructions. 
Collecting measurements across a range of frequencies can significantly stabilize the inversion, as each frequency
band probes complementary aspects of the medium.
In practice, reconstructions are carried out by aggregating the information from multiple frequencies, leading to a \emph{wideband} formulation where the forward operator
$\mathcal{F}_{\bar{\Omega}}(m)=\{y_\omega\}_{\omega\in\bar{\Omega}}$
collects data over a set of pre-specified frequencies $\bar{\Omega}$.

\paragraph{Adjoint} 
Instead of using the adjoint of the discretized PDE, we use the far-field asymptotic approximation~\cite{colton1998inverse} of the Green’s function of~\eqref{eqn:Helmholtz} to write (up to a frequency-dependent complex prefactor) the linearized forward operator as a Fourier transform:
\begin{equation}
    Fm\approx \int_\Omega e^{\mathrm{i}\omega(s-r)\cdot x}m(x)dx,
\end{equation}
which then gives a formulation of the far-field adjoint:
\begin{equation}
    (F^*_\omega y)(x)\approx \int_{\mathbb{S}^1\times\mathbb{S}^1} e^{\mathrm{i}\omega(r-s)\cdot x}\,y_\omega(r,s)\,dr\,ds .
\end{equation}
This expression allows us to efficiently apply an approximate adjoint map. Despite not being the exact adjoint, it provides informative back-projections of the measurement data as illustrated in Figure~\ref{fig:four_panels}. 
Since we are in the wideband setting, each frequency admits an adjoint $F^*_\omega$; we initialize \ourname with the real component of the lowest-frequency adjoint because this resolves the large-scale structures of the desired reconstruction and is less prone to high-frequency artifacts.

\subsubsection{Benchmark Dataset}
We use the evaluation suite from the Back-Projection Diffusion (BPD) benchmark~\cite{Zhang2025_BPD}, which contains 4 datasets of varying difficulty, designed to probe distinct challenges in nonlinear inverse scattering:
\begin{enumerate}
    \item The \emph{3--5--10h Triangles} dataset contains small, randomly placed and oriented triangles with sub-wavelength features, testing the ability of the network to recover fine-scale details.
    \item The \emph{10h Overlapping Squares} dataset~\cite{10hsquares_data} consists of multiple strongly overlapping squares, emphasizing nonlinear superposition and multiple scattering effects.
    \item The \emph{Shepp--Logan Phantom} dataset, designed as a simple mimicry of human brains, includes strong discontinuities and resonant cavity effects, producing dominant reflections and masking interior structures.
    \item The \emph{Brain Scans} dataset uses the processed MRI brain perturbations released with BPD~\cite{brain_data}, derived from fastMRI~\cite{fastmri, fastmri2} to evaluate performance on complex, non-synthetic perturbations with rich spatial variability.
\end{enumerate}

\paragraph{Numerical details}
The datasets use a physical domain of $\Omega=[-0.5,0.5]^2$, discretized on an $80 \times 80$ uniform grid. 
Measurements are collected
at 80 equally spaced receivers on a circle of radius
$R=0.5$, using 80 equally spaced incident plane-wave directions, so that the discretized measurement data is represented by a complex matrix $y_\omega\in\mathbb{C}^{80\times 80}$.
Far-field patterns are generated by numerically solving~\eqref{eq:scattered_wavefield} at three dyadically spaced frequencies:
$\omega/(2\pi) \in\{ 2.5, 5,10\}$ with an effective wavelength of 8 points per wavelength (PPW) for all but the \textit{Brain Scans} dataset; and
$\omega/(2\pi) \in\{ 4, 8,16\}$ with an effective wavelength of 5 PPW for \textit{Brain Scans}.
Differential operators are approximated using tensor-product eighth-order finite differences to avoid numerical dispersion, and radiation boundary conditions are enforced via a second-order perfectly matched layer (PML) with intensity~80. 
The resulting sparse linear systems are solved using sparse LU factorization. 

\paragraph{Methods}

\begin{table*}[t]
\caption{We compare performance on three synthetic datasets (\textit{Shepp--Logan, 3--5--10h Triangles, 10h Overlapping Squares}) using RRMSE (\%), MELR $(\times 10^{-2})$, and Sinkhorn divergence (SD) metrics (see~\ref{app:metrics} for definitions, bold indicates best performance for a given metric). All Sinkhorn divergence values, including those of the baselines,
are recomputed using an updated library version
and therefore differ from those reported in~\cite{Zhang2025_BPD}. We find that \ourname outperforms all models, including conditional diffusion, across the board.} \label{table:results_non_linear}
\centering
\scriptsize
\begin{tabular}{lccc ccc ccc}
\toprule
\textbf{} &
\multicolumn{3}{c}{\textbf{Shepp--Logan}} &
\multicolumn{3}{c}{\textbf{3--5--10h Triangles }} &
\multicolumn{3}{c}{\textbf{10h Overlapping Squares }} \\
\cmidrule(lr){2-4}\cmidrule(lr){5-7}\cmidrule(lr){8-10}
\textbf{Method} \hspace{1.5cm} & \textbf{RRMSE} $\downarrow$  & \textbf{MELR}$\downarrow$ & \textbf{SD}$\downarrow$
& \textbf{RRMSE} $\downarrow$ & \textbf{MELR}  $\downarrow$ & \textbf{SD}$\downarrow$
& \textbf{RRMSE}$\downarrow$  & \textbf{MELR}  $\downarrow$ & \textbf{SD} $\downarrow$\\
\midrule
\multicolumn{10}{l}{\textbf{Traditional}} \\
FWI
& 52.041  & 241.893 & 7.658
& 28.637  & 159.894 & 0.809
& 38.777  & 281.057 & 4.773 \\
\midrule
\multicolumn{10}{l}{\textbf{Deterministic}} \\
SwitchNet
& 4.305  & 8.071   & 1.122
& 15.084  & 9.377  & 0.570
& 20.102  & 24.295 & 2.643   \\
WideBNet
& 3.843  & 13.255 & 1.059
& 17.263  & 10.582 & 0.608
& 14.327  & 44.182 & 1.996 \\
B-EquiNet
& 2.022  & 2.906 & 0.805
& 2.944  & 2.010 & 0.301
& 9.484  & 21.434 & 1.456 \\
EquiNet
& 1.693  & 2.827 &  0.758
& 2.741  & 1.734 &  0.298
& 10.891  & 25.906 & 1.614   \\
\midrule
\multicolumn{10}{l}{\textbf{Conditional Diffusion}} \\
B-EquiNet CDM
& 1.406  & 1.757 &  0.722
& 1.657  & 1.318 &  0.263
& 2.046  & 2.683 &  0.607 \\
EquiNet CDM
& 1.323  & 1.564 &   0.717
& 1.590  & 1.385 &   0.260
& 1.744  & 1.979 &   0.578 \\
\midrule
\multicolumn{10}{l}{\textbf{Ours}} \\
\ourname
& \textbf{0.379}  & \textbf{0.474} & \textbf{0.576}
& \textbf{1.108} & \textbf{0.607} & \textbf{0.258}
& \textbf{1.143}  & \textbf{1.401} & \textbf{0.514} \\
\bottomrule
\end{tabular}
\end{table*}

\begin{wraptable}{r}{0.5\textwidth}
\centering
\caption{We compare \ourname and \ournameRED to PnP-DP and CDM baselines for nonlinear inverse scattering on the \textit{Brain Scans} dataset, with bold indicating the best performance on each metric. 
We report RRMSE as a percentage and MELR $\times10^{-2}$.
EquiNet CDM results are obtained from a model we trained using the pipeline from~\cite{Zhang2025_BPD}.
CDM$^\dagger$ indicates that we evaluate the diffusion model at the same number of function evaluations as \ourname (namely, 8 sampling steps), thus demonstrating the advantage of \ourname when compute is limited.}
\label{table:brain_scans}

\footnotesize
\setlength{\tabcolsep}{11.5pt}
\renewcommand{\arraystretch}{1.05}

\begin{tabular}{lccc}
\toprule
&
\multicolumn{3}{c}{\textbf{Brain Scans}} \\
\cmidrule(lr){2-4}
\textbf{Method} & \textbf{RRMSE} $\downarrow$ & \textbf{MELR} $\downarrow$ & \textbf{SD} $\downarrow$ \\
\midrule
\multicolumn{4}{l}{\textbf{PnP Diffusion Prior}} \\
DAPS      & 12.572 & 18.700 & 0.278 \\
RED-Diff  &  2.873 &  7.370 & 0.107 \\
\midrule
\multicolumn{4}{l}{\textbf{Conditional Diffusion}} \\
EquiNet CDM & 5.661 & 3.494 & 0.157 \\
EquiNet CDM$^\dagger$ & 34.990 & 99.337 & 0.667 \\
\midrule
\multicolumn{4}{l}{\textbf{Ours}} \\
\ourname    & 3.421 & 4.482 & 0.136 \\
\ournameRED & \textbf{2.467} & \textbf{3.389} & \textbf{0.099} \\
\bottomrule
\end{tabular}
\end{wraptable}

The BPD benchmark evaluates a range of both classical and machine learning algorithms which includes: 
(a) Full Waveform Inversion (FWI)~\cite{Tarantola:Inversion_of_seismic_reflection_data_in_the_acoustic_approximation}, a classical optimization algorithm;
(b) SwitchNet~\cite{KhooYingSwitchNet:2019}, WideBNet~\cite{li2022butterfly}, EquiNet/B-EquiNet~\cite{zhang2024_equivariant}, four deterministic ML models based on the FBP formula~\eqref{eq:FBP}; and
(c) EquiNet CDM and B-EquiNet CDM\footnote{While the original paper calls these models EquiNet-CNN and B-EquiNet-CNN, we refer to them as EquiNet CDM and B-EquiNet CDM to emphasize the diffusion model formulation.}
~\cite{Zhang2025_BPD}, two FBP-based conditional diffusion models.
Furthermore, we instantiate \ourname with the same EquiNet architecture as EquiNet CDM (the best performing model from the BPD benchmark) to isolate the advantage of our modified SI formulation against a more traditional diffusion model.
For the \textit{Brain Scans} dataset, we add additional comparisons against two state-of-the-art PnP methods: RED-Diff \cite{mardani2024reddiff} and DAPS \cite{zhang2024daps}.
For these PnP methods, we evaluate the forward model using
a JAX-based finite-difference Helmholtz solver\footnote{\url{https://github.com/maguerrap/Inverse_Scattering}} configured to match the measurement data.

\subsubsection{Results}

\begin{figure*}
    \centering
    \includegraphics[width=\linewidth]{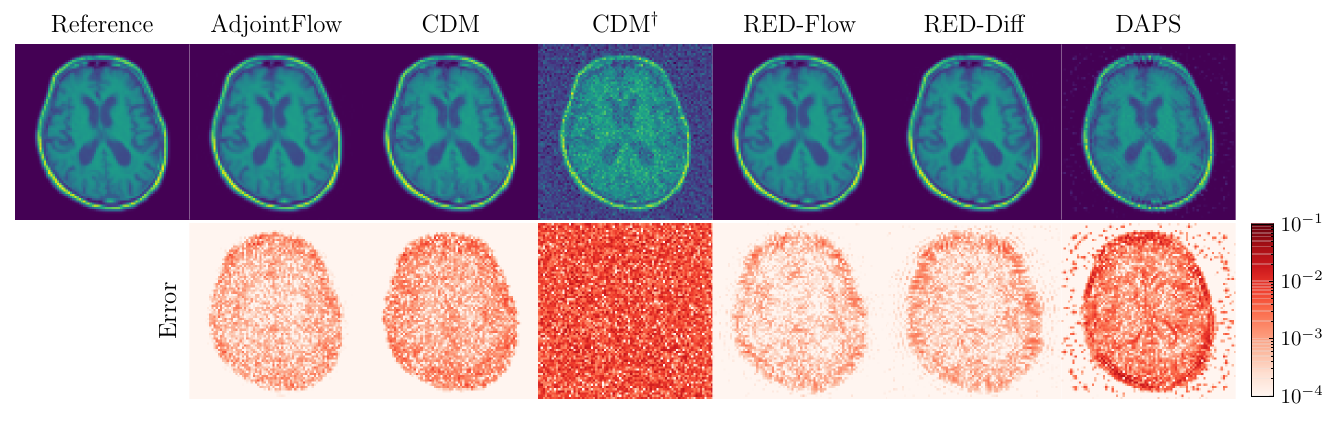}
    \caption{Qualitative comparison of reconstructions on the \textit{Brain Scans} dataset for nonlinear inverse scattering. For CDM$^\dagger$, we evaluate the diffusion model at the same number of NFE (number of function evaluations) as \ourname (namely, we fix 8 sampling steps). \ourname clearly performs far better in this iso-compute comparison, demonstrating its superior efficiency.}
    \label{fig:sub-brain-comparison}
\end{figure*}

We showcase representative reconstructions in~Figure~\ref{fig:four_panels} and~Figure~\ref{fig:sub-brain-comparison}, where we observe that \ourname successfully recovers sub-wavelength structures and handles multiple-scattering effects efficiently.
Quantitative results are given in Tables~\ref{table:results_non_linear} and~\ref{table:brain_scans}.
In addition to improved accuracy, \ourname also provides accelerated inference, requiring only a single call to the adjoint and a small number of integration steps to solve the ODE in~\eqref{eq:flow_ode_adjointflow}.
As a result, it becomes much more efficient than conditional diffusion, which can require hundreds of sampling steps, as well as PnP-DP algorithms, which can require hundreds of function / gradient evaluations.
The result, illustrated in~Figure~\ref{fig:brain_barchart}, reveals an inference acceleration of up to \textit{five orders of magnitude}, while reconstruction accuracy still remains competitive.
We find that inference-time scaling methods like \ournameRED can boost the performance of \ourname even further, albeit at a higher computational cost.

\subsection{Linear Frequency-Domain Inverse Scattering}\label{sec:lin_frequency_inverse_scattering}

Often encountered in optical microscopy~\cite{Matlock:20} due to the weak scattering (little to no multiple internal reflections) of most cells and thin tissue, this inverse problem linearizes the problem presented in~Section~\ref{sec:nonlin_frequency_inverse_scattering}, approximating the impulse response with a linear map $F\approx\calF$. Practically, this allows faster and more stable reconstructions for high-throughput and live-cell imaging.
Here the forward model stems from a first-order Born approximation ~\eqref{eq:born_series} resulting in
\begin{equation}\label{eq:born_approx}
    y\approx F m,
\end{equation}
where $F$ consists of the linear measurement operator composed with the linear approximation
\begin{equation}
    u^\textrm{sc}(x)\approx \omega^2\int G_\omega(x, \xi)\,m(\xi)\,e^{\mathrm{i}\omega s\cdot \xi}d\xi,
\end{equation}
where $G_\omega$ is the Green's function for the homogeneous two-dimensional Helmholtz equation.

\paragraph{Linear inverse problems}
The presence of a linear forward map makes this setup structurally similar to many common inverse problems in image processing and computer vision, such as de-blurring and super-resolution. 
Therefore, it provides a middle ground to test methods like \ourname, specifically designed for scientific inverse problems, with other generative inversion algorithms which were primarily designed with linear image-based methods in mind.
In particular, this includes PnP-DP algorithms which are \emph{only} applicable in a linear regime, for example because they rely on access to a singular value decomposition or pseudoinverse.
Furthermore, linearity significantly affects accuracy-cost tradeoffs, as forward and gradient evaluations are no longer prohibitively expensive; they can now be computed via simple matrix multiplications.

\subsubsection{Benchmark Dataset}
We use the dataset of microscopic cell phantoms (based on HL60 nuclei) generated by InverseBench \cite{zheng2025inversebench,inversebench_data} using the online simulator CytoPacq \cite{wiesner2019cytopacq}.

\paragraph{Numerical details}
The physical domain is a square $\Omega=[-0.09,0.09]^2$ discretized on a $128 \times 128$ uniform grid. 
There are $n_\textrm{src}=20$ uniformly spaced sources at a radius $R=1.6$ from the origin, and a specified number $n_\textrm{rec}\in\{60,180,360\}$ of equally spaced receivers to evaluate performance across different levels of data sparsity.
The simulation uses monochromatic illumination with an effective wavelength of 6 PPW.
Measurements are generated using a linear map constructed via the first Born approximation~\eqref{eq:born_approx}.
The specific unknown used in the benchmark is the scattering potential $\frac{\omega^2}{4\pi}m$ for permittivity $m$.
In keeping with the original benchmark, Gaussian measurement noise $\xi_y\sim\mathcal{N}(0, \sigma_y^2I)$ with strength $\sigma_y=10^{-4}$ is added to the measurement data.

\begin{figure*}[t]
    \centering
    \begin{subfigure}[t]{0.48\textwidth}
        \centering
        \includegraphics[width=\linewidth]{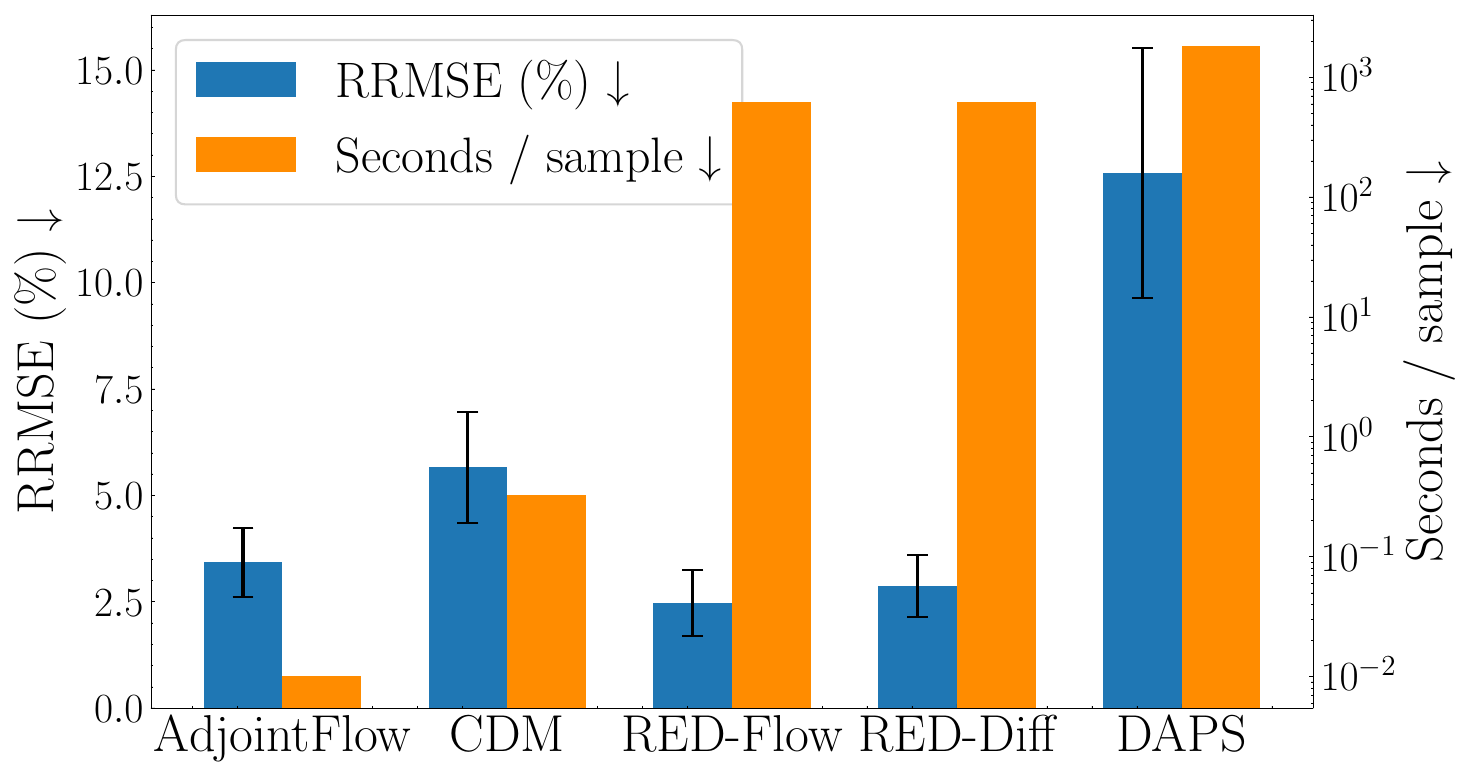}
        \caption{Nonlinear inverse scattering: Brain Scan.}
        \label{fig:brain_barchart}
    \end{subfigure}
    \hfill
    \begin{subfigure}[t]{0.48\textwidth}
        \centering
        \includegraphics[width=\linewidth]{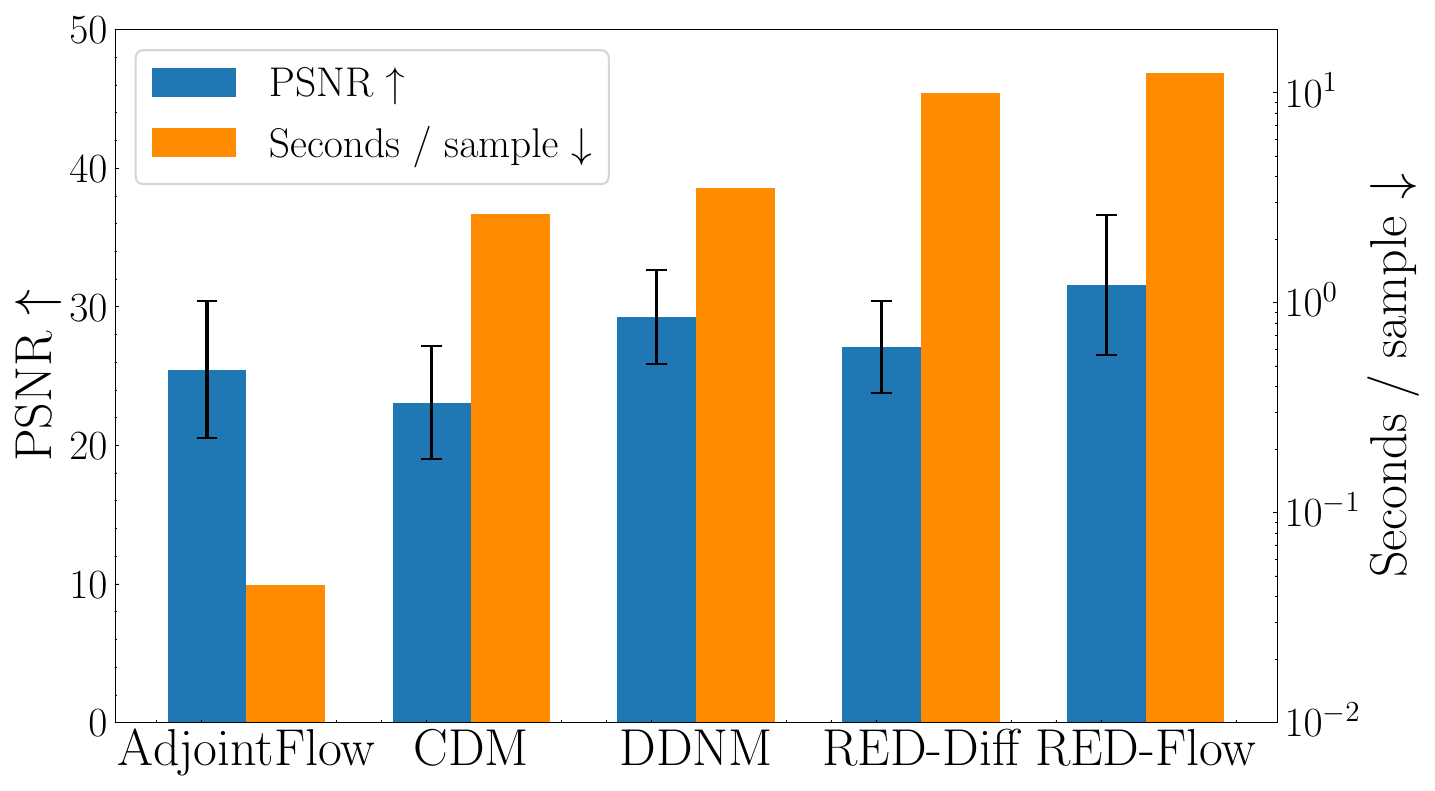}
        \caption{Linear inverse scattering: 60 receivers.}
        \label{fig:lininv_barchart}
    \end{subfigure}
    \caption{Comparison of the efficiency--accuracy tradeoff for nonlinear and linear inverse scattering. All timings were measured on the same NVIDIA RTX A6000 GPU. (a) \ourname achieves a strong balance between performance and efficiency, while RED-Diff and \ournameRED offer small accuracy improvements at substantially greater inference cost; the seconds-per-sample axis uses a logarithmic scale. (b) \ourname achieves competitive accuracy at the lowest cost, though the efficiency gain is smaller than in the nonlinear problem.}
    \label{fig:scattering_efficiency}
\end{figure*}

\paragraph{Methods}
InverseBench evaluates a large number of PnP diffusion algorithms (see~\ref{app:pnp_methods}), all using the same pretrained diffusion prior, as well as the traditional optimization algorithm FISTA-TV~\cite{beck2009fasttv}. 
The results from InverseBench use a larger U-net based architecture since, as a diffusion prior, it has no need to specially process conditional measurement inputs. 
However, given that \ourname uses a conditioned network, we adopt a smaller EquiNet architecture similar to that used in Section~\ref{sec:nonlin_frequency_inverse_scattering}.
We also add a comparison against a CDM with the same EquiNet architecture to separate the benefit of the architecture from the generative model.
Furthermore, for this case, we choose the coupling for \ourname to be with the regularized pseudoinverse, namely, $a(y)=F_\epsilon^\dagger y = (F^* F + \epsilon I)^{-1}F^*y$, with $\epsilon>0$ the regularization coefficient. 
The pseudoinverse is also concatenated as an additional channel to the conditional input to the CDM.

\subsubsection{Results}
\begin{figure*}[!htbp]
    \centering
    \includegraphics[width=\linewidth]{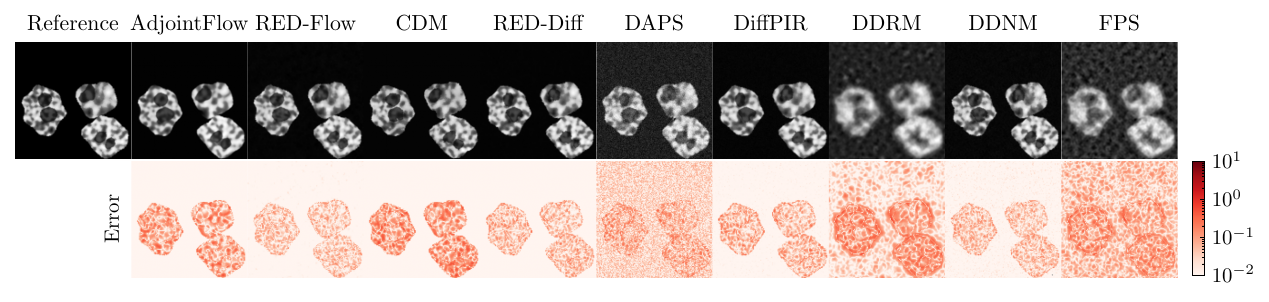}
    \caption{Qualitative comparisons of reconstructions from \ourname and diffusion baselines on linear inverse scattering with 60 receivers.}
    \label{fig:inv_scatter_comparison}
\end{figure*}

Quantitative results are summarized in Table~\ref{table:results_linear_inv_scattering_full}, with a representative example shown in~Figure~\ref{fig:inv_scatter_comparison}.
On all three levels of measurement sparsity, \ournameRED outperforms baselines by a significant margin in reconstruction PSNR, and also surpasses most to all baselines on SSIM.
Therefore, while the linear setting favors PnP-DP methods -- many of which rely on the SVD or pseudoinverse of the forward map -- \ourname still delivers competitive performance while only leveraging the linear structure of the problem via the physics-based initialization.

Figure~\ref{fig:lininv_barchart} again demonstrates that \ourname strongly leads competing methods in terms of speed, albeit with a smaller performance gap when compared to the nonlinear problem (see Figure~\ref{fig:brain_barchart}) due to the reduced cost of running iterative algorithms in the linear setting.
And, with an increased inference-time cost, \ournameRED again produces state-of-the-art reconstruction quality without any retraining of the base model.
Furthermore, while methods which explicitly leverage the linear structure of the forward map -- such as DDNM, FPS, and MCGdiff -- retain the lowest measurement errors, \ournameRED remains competitive with many of these methods, while also significantly improving data consistency over the base \ourname model, especially in the low-data regime.
Our findings thus support that additional sampling guidance via the forward model is especially helpful in producing physically plausible reconstructions.

\begin{table}[t]
\caption{We present the results for linear inverse scattering with three different levels of measurement sparsity and noise level $\sigma_y=10^{-4}$ following~\cite{zheng2025inversebench} (bold result indicates best). 
``Meas err'' quantifies relative agreement between predicted and observed measurements, following InverseBench's normalization convention.
Across all three levels of data sparsity, \ournameRED achieves the best reconstruction PSNR, while also having either the best or close to the best SSIM and competitive data consistency.}
\label{table:results_linear_inv_scattering_full}
\centering
\small
\resizebox{\textwidth}{!}{%
\begin{tabular}{lccc ccc ccc}
\toprule
Number of receivers &
\multicolumn{3}{c}{\textbf{360}} &
\multicolumn{3}{c}{\textbf{180}} &
\multicolumn{3}{c}{\textbf{60}} \\
\cmidrule(lr){2-4}\cmidrule(lr){5-7}\cmidrule(lr){8-10}
\textbf{Method}
& \textbf{PSNR} $\uparrow$ & \textbf{SSIM} $\uparrow$ & \textbf{Meas err} (\%) $\downarrow$
& \textbf{PSNR} $\uparrow$ & \textbf{SSIM} $\uparrow$ & \textbf{Meas err} (\%) $\downarrow$
& \textbf{PSNR} $\uparrow$ & \textbf{SSIM} $\uparrow$ & \textbf{Meas err} (\%) $\downarrow$\\
\midrule
\multicolumn{10}{l}{\textbf{Traditional}} \\
FISTA-TV
& 32.126 (2.139) & 0.979 (0.009) & 1.23 (0.25)
& 26.523 (2.678) & 0.914 (0.040) & 2.65 (0.30)
& 20.938 (2.513) & 0.709 (0.103) & 6.05 (0.65) \\
\midrule
\multicolumn{10}{l}{\textbf{PnP Diffusion Prior}} \\
DDRM
& 32.598 (1.825) & 0.929 (0.012) & 1.04 (0.26)
& 28.080 (1.516) & 0.890 (0.019) & 1.57 (0.39)
& 20.436 (1.210) & 0.545 (0.037) & 3.04 (0.92) \\
DDNM
& 36.381 (1.098) & 0.935 (0.017) & 0.78 (0.22)
& 35.024 (0.993) & 0.895 (0.027) & 0.58 (0.16)
& 29.235 (3.376) & 0.917 (0.022) & 0.28 (0.07) \\
$\Pi$GDM
& 27.925 (3.211) & 0.889 (0.072) & 2.74 (1.23)
& 26.412 (3.430) & 0.816 (0.114) & 3.66 (1.79)
& 20.074 (2.608) & 0.540 (0.198) & 6.90 (3.38) \\
DPS
& 32.061 (2.163) & 0.846 (0.127) & 4.35 (1.19)
& 31.798 (2.163) & 0.862 (0.123) & 4.28 (1.20)
& 27.372 (3.415) & 0.813 (0.133) & 4.53 (1.31) \\
LGD
& 27.901 (2.346) & 0.812 (0.037) & 1.17 (0.20)
& 27.837 (2.337) & 0.803 (0.034) & 1.06 (0.16)
& 20.491 (3.031) & 0.552 (0.077) & 1.45 (0.68) \\
DiffPIR
& 34.241 (2.310) & \textbf{0.988} (0.006) & 1.11 (0.24)
& 34.010 (2.269) & \textbf{0.987} (0.006) & 1.04 (0.23)
& 26.321 (3.272) & 0.918 (0.028) & 1.27 (0.23) \\
PnP-DM
& 33.914 (2.054) & \textbf{0.988} (0.006) & 1.21 (0.25)
& 31.817 (2.073) & 0.981 (0.008) & 1.42 (0.26)
& 24.715 (2.874) & 0.909 (0.046) & 2.20 (0.34) \\
DAPS
& 34.641 (1.693) & 0.957 (0.006) & 1.03 (0.25)
& 33.160 (1.704) & 0.944 (0.009) & 1.11 (0.25)
& 25.875 (3.110) & 0.885 (0.030) & 1.51 (0.25) \\
RED-Diff
& 36.556 (2.292) & 0.981 (0.005) & 0.89 (0.23)
& 35.411 (2.166) & 0.984 (0.004) & 0.87 (0.21)
& 27.072 (3.330) & 0.935 (0.037) & 1.18 (0.23) \\
FPS
& 33.242 (1.602) & 0.870 (0.026) & \textbf{0.70} (0.01)
& 29.624 (1.651) & 0.710 (0.040) & \textbf{0.37} (0.01)
& 21.323 (1.445) & 0.460 (0.030) & \textbf{0.15} (0.02) \\
MCGdiff
& 30.937 (1.964) & 0.751 (0.029) & \textbf{0.70} (0.01)
& 28.057 (1.672) & 0.631 (0.042) & 0.38 (0.01)
& 21.004 (1.571) & 0.445 (0.028) & 0.21 (0.06) \\
\midrule
\multicolumn{10}{l}{\textbf{Conditional Diffusion}} \\
CDM
& 29.921 (3.267) & 0.952 (0.019) & 2.56 (0.45)
& 29.609 (3.289) & 0.964 (0.017) & 2.49 (0.27)
& 23.064 (4.078) & 0.886 (0.070) & 7.48 (0.91) \\
\midrule
\multicolumn{10}{l}{\textbf{Ours}} \\
\ourname
& 32.797 (3.966) & 0.981 (0.013) & 1.86 (0.21)
& 32.459 (3.935) & 0.980 (0.013) & 1.87 (0.20)
& 25.460 (4.969) & 0.920 (0.054) & 4.70 (0.99) \\
\ournameRED
& \textbf{38.651} (2.871) & 0.984 (0.003) & 0.97 (0.32)
& \textbf{37.476} (3.081) & 0.984 (0.003) & 0.87 (0.30)
& \textbf{31.553} (5.028) & \textbf{0.936} (0.032) & 0.66 (0.16) \\
\bottomrule
\end{tabular}
}
\end{table}

\subsection{Time-Domain Seismic Imaging}\label{sec:time_domain_seismic_imaging}
Widely used in geophysics, time-domain inverse scattering reconstructs complex underground structures using broadband seismic waveforms, often with sources and/or receivers limited to the surface.

\paragraph{Forward problem}
The full waveform inversion problem aims to recover the features of sub-surface geophysical structures using the full time-series information of measured waveforms. 
Here, we aim to recover the compressional wavespeed $m(x)=v(x)$, which determines the speed at which sound waves travel through solids like soil and rock. 
This quantity is crucial for subsurface mapping as it varies considerably with material density, porosity, lithology, fracturing, and pressure.

Mathematically, we model the propagation of seismic waves through a heterogeneous medium with the acoustic wave equation
\begin{equation}\label{eq:acoustic_wave_eqn}
    \frac{1}{v^2}\frac{\partial^2 u}{\partial t^2}-\Delta u = q(x, t),
\end{equation}
where $q(x, t)$ is the source wave and $u(x,t)$ is the pressure wavefield. 
This equation is then discretized as
\begin{equation}
    Au=q,
\end{equation}
where $A$ represents the discretization of the operator $\frac{1}{v^2}\frac{\partial^2}{\partial t^2}-\Delta$ and therefore depends on $v$ in a nonlinear fashion. Here $A$ encodes both the time and spatial dimensions, and acts on the pressure field for all discretized times and positions. For explicit time integrators, $A$ usually has a block-lower triangular structure inherited from time causality.  
Finally, measurements are obtained from
\begin{equation}
    y=\calF(m)=Pu,
\end{equation}
where $P$ is the sampling operator for receivers which records the wavefield pressure at specified surface locations over a series of timesteps.

\paragraph{Adjoint}
Using the adjoint-state method to compute the gradient is a classical recipe \cite{Plessix_2006:ajoint_state}. Algorithmically, the adjoint is computed by treating the residual between the observed measurements $y$ and the predicted measurements $\calF(\hat m)$ as a new source. It injects the error as a wave and propagates it backwards in time to get the adjoint wavefield $\delta u$ defined as
\begin{equation}\label{eqn:adjoint_wavefield}
    A^\top\delta u = P^\top\delta y, \quad \text{  where  }\,\, \delta y = y - \calF(\hat{m}).
\end{equation}

Here, $A^\top$ will have a block upper-triangular structure, which requires solving for $\delta u$ backwards in time. Finally, applying~Equation~\eqref{eqn:adjoint_wavefield} allows us to compute the gradient of the data misfit functional $\mathcal{J}(\hat{m})=\tfrac12\|y-\calF(\hat{m})\|^2$ \cite{Plessix_2006:ajoint_state}. 
In particular, the gradient is obtained by zero-lag cross-correlation of the forward wavefield $u$ and the adjoint wavefield $\delta u$:
\begin{equation}
    \nabla_{s\,}\mathcal{J} = \sum_{t=0}^Tu(x, t)\frac{\partial^2}{\partial t^2}\delta u(x, t),
\end{equation}
which can subsequently be used in gradient descent optimization methods.
Here the cross-correlation is expressed in squared-slowness coordinates $s=m^{-2}$; the wavespeed gradient follows by applying the chain rule: $\nabla_{m\,}\mathcal{J}=-2m^{-3}\nabla_{s\,}\mathcal{J}$.

The adjoint-state method thus enables direct computation of the back-projection $-\nabla_{m}\mathcal{J}=F^*(y-\calF(\overline m))$ by simply running the wavefield backward-in-time on a reference medium $\overline{m}$ and cross-correlating. For our case, we take the reference medium to be the mean of the training dataset, which approximates a simple depth-varying gradient. Light additional post-processing is needed to smooth out artifacts near the sources and receivers and render amplitudes consistent across depths.

\paragraph{Challenges}
As measurements are typically only taken at the free surface, the inverse problem is severely ill-posed and can have non-unique solutions.
Classical algorithms that apply the adjoint-state method and cast the inversion as a local optimization problem are often more prone to cycle-skipping than in the frequency-domain inverse scattering problem.
As a result, the success of this method heavily depends on starting from a sufficiently accurate initial model, which is often generated from simpler ray-based methods such as travel-time tomography \cite{Virieux_Operto}.
In addition, many state-of-the-art methods rely on frequency sweeps during the optimization loop~\cite{bunks1995multiscale}, where one applies low-pass filters at increasing frequency cut-offs to regularize the landscape of the loss. This procedure is closely related to the recursive-linearization approach \cite{chen1997inverse} through the Parseval-Plancherel identity, though the two still differ in their update rules and continuation schedules. In this case, however, to remain consistent with InverseBench, we use its FWI setup based on Devito~\cite{louboutin2019devito}, which directly minimizes the full time-domain data misfit without frequency continuation or low-pass filtering.

\subsubsection{Benchmark Dataset}
We use the adapted dataset distributed by
InverseBench~\cite{zheng2025inversebench, inversebench_data}, derived from the CurveFault dataset in OpenFWI~\cite{deng2022openfwi}  composed of velocity profiles typically found in seismic imaging.

\paragraph{Numerical details}
Velocity maps have been resized to $128 \times 128$ from the original CurveFault dataset using bilinear interpolation and anti-aliasing.
Measurements are generated utilizing the Devito software stack~\cite{louboutin2019devito}.
The source is given as a Ricker wavelet with a central frequency of 5 Hz, and the simulation records 1,001 timesteps with a temporal sampling rate of $dt = 0.001$ [s]. 
The acquisition geometry consists of 16 evenly-spaced sources at a depth of 1,270 [m] and 129 evenly-spaced receivers at a depth of 10 [m], mimicking regional earthquake transmission tomography.
As a result, the final measurement data is of shape $y\in\mathbb{R}^{1001\times 129\times 16}$.
All boundaries except the top are set to be absorbing to avoid artificial reflections.

\paragraph{Methods}
We benchmark \ourname against six PnP-DP algorithms evaluated by InverseBench which, as before, all use the same pretrained diffusion prior model based on an ADM U-net architecture. 
We also report InverseBench’s classical optimization baselines (Adam, L-BFGS)~\cite{zheng2025inversebench}; rows marked $\dagger$ consider a ``best-case'' scenario where the optimization is initialized with a smooth background model of the target medium, while the unmarked Adam row is initialized at a constant velocity of 3 km/s.

Seismic imaging FWI does not carry the same symmetries as full-aperture frequency-domain inverse scattering; therefore, we instantiate \ourname with a comparable ADM U-net architecture~\cite{dhariwal2021diffusion} to that used by InverseBench. 
To process the conditional inputs, we first take a Fourier transform of the raw wavefield and select a subset of coefficients based on their energy content, reducing the measurement dimensionality from $16\times129\times1001$ to $16\times33\times20$.
From there, we add a Fourier positional embedding~\cite{tancik2020fourier} to encode the acquisition geometry of the sources and receivers and then process the data with a small MLP to obtain a latent context vector of dimension $16\times33\times D_\textrm{emb}$ (we choose $D_\textrm{emb}=512$), which is then injected into the U-net via added cross-attention blocks.
Finally, for training we set the target to be the velocity perturbations relative to the reference medium $\overline m$.
We again include a comparison against a CDM with the same architecture as \ourname.

\subsubsection{Results}
\begin{figure*}[t]
    \centering
    \includegraphics[width=\linewidth]{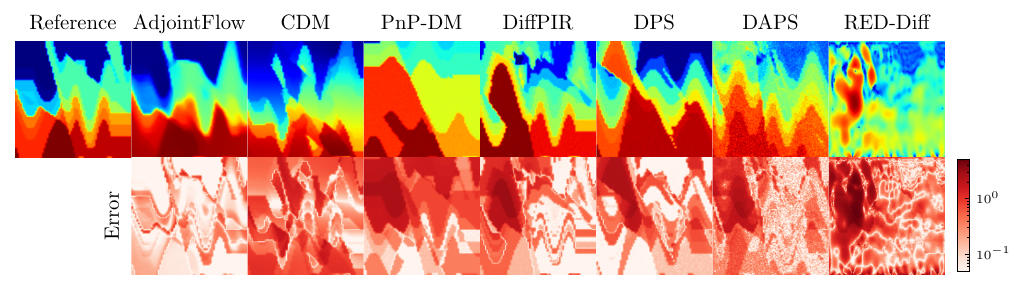}
    \caption{Comparisons of \ourname against different diffusion-based methods on the seismic imaging task.}
    \label{fig:fwi_comparison}
\end{figure*}

Results are summarized in Table~\ref{table:time_domain_results}, which demonstrates that \ourname outperforms all non-oracle baselines on all metrics but data consistency. 
Notably, we observe that for image-reconstruction metrics, generative models which approximate the conditional distribution \emph{directly} (\ourname, CDM) -- thus leveraging the knowledge encoded in paired data -- mostly outperform PnP-DP approaches which attempt to approximate this distribution via a Bayesian framework at inference time.
On the other hand, PnP-DP approaches often provide better data consistency due to their direct use of the forward model.

However, given the reliance of PnP methods on minimizing a data misfit functional at inference time, we find that they are prone to suffer from cycle-skipping in this setting because the functional is highly nonconvex (see Figure~\ref{fig:fwi_extended}).
The significantly higher variance exhibited by almost all PnP-DP methods in Table~\ref{table:time_domain_results} further evidences the instability.

Minimizing the data misfit functional is also computationally demanding, as calculating its derivatives requires intensive PDE simulations. 
PnP reconstructions inherit this high cost, and thus often take hours to generate a single result.
Furthermore, while InverseBench~\cite{zheng2025inversebench} notes that the performance of PnP methods on seismic imaging can be highly sensitive to hyperparameter choices, \ourname works out-of-the-box and circumvents the need for tedious tuning procedures.
Thus, \ourname avoids these inherent limitations of PnP methods and produces significantly more efficient and stable reconstructions of the wavespeed. 
Figure~\ref{fig:fwi_comparison} illustrates a qualitative comparison of the results, where the relative stability of \ourname compared to other methods is quite apparent.

\begin{table}[!htbp]
\caption{We present the results for seismic imaging from InverseBench~\cite{zheng2025inversebench} (bold result indicates best non-oracle method). 
``Data misfit'' quantifies agreement between predicted and observed measurements, following InverseBench's normalization convention.
For \ourname we report performance using 1 sampling step, as we found this to give the best performance based on validation data. $\dagger$ indicates ``oracle models'' initialized from ground-truth data blurred with a Gaussian filter ($\sigma=20$ pixels), which requires access to the unknown ground truth and therefore serves only as an oracle baseline.
}
\label{table:time_domain_results}
\centering
\footnotesize
\setlength{\tabcolsep}{3.5pt}    
\renewcommand{\arraystretch}{1.05} 
\begin{tabular}{lcccc}
\toprule
\textbf{Method} & \textbf{RRMSE}  $\downarrow$ & \textbf{PSNR}$\uparrow$ & \textbf{SSIM}$\uparrow$ & \textbf{Data misfit}$\downarrow$ \\
\midrule
\multicolumn{5}{l}{\textbf{Traditional}} \\
Adam     & 33.3 (8.6) &  9.968 (2.083) & 0.305 (0.120) & 115.14 (52.10) \\
Adam$^\dagger$  & 8.9 (2.1) & 21.273 (2.045) & 0.679 (0.073) &  15.89 (10.16) \\
L-BFGS$^\dagger$ & 7.0 (2.3) & 23.398 (2.749) & 0.704 (0.077) &   9.18 (6.47) \\
\midrule
\multicolumn{5}{l}{\textbf{PnP Diffusion Prior}} \\
DPS      & 25.0 (15.4) & 14.111 (6.820) & 0.491 (0.161) & 155.08 (92.17) \\
LGD      & 24.4 (2.4) & 12.288 (0.889) & 0.341 (0.047) & 258.47 (26.40) \\
DiffPIR  & 20.4 (12.9) & 16.113 (6.962) & 0.554 (0.191) & \textbf{88.53} (56.91) \\
DAPS$^\dagger$  & 20.1 (10.3) & 14.914 (4.184) & 0.321 (0.067) & 111.13 (71.33) \\
PnP-DM   & 25.9 (7.5) & 11.983 (2.269) & 0.431 (0.073) & 308.84 (26.34) \\
RED-Diff  & 31.9 (10.2) & 10.372 (2.650) & 0.280 (0.108) &  94.67 (41.33) \\
\midrule
\multicolumn{5}{l}{\textbf{Conditional Diffusion}} \\
CDM
& 17.6 (2.1) & 15.074 (1.298) & 0.425 (0.064) &  243.25 (35.69) \\
\midrule
\multicolumn{5}{l}{\textbf{Ours}} \\
\ourname
& \textbf{9.7} (1.9) &\textbf{ 20.530} (1.516) & \textbf{0.676} (0.062) & 115.63 (30.89) \\
\bottomrule
\end{tabular}
\vspace{-1.5mm}
\end{table}

\subsection{Black Hole Imaging}\label{sec:black_hole_imaging}
According to the Rayleigh criterion (or diffraction limit) the resolution of a telescope is inversely proportional to the diameter of its aperture. Thus, to detect objects with small angular scales, such as black holes, one would desire a telescope as large as possible. While we cannot feasibly construct an \emph{Earth-sized telescope},
it is still possible to obtain sparse measurements corresponding to such a telescope using an array of linked telescopes across the globe via a technique known as Very Long Baseline Interferometry (VLBI).

\paragraph{Forward problem}
We wish to measure the brightness distribution of a black hole $m=m(\vartheta)$, where $\vartheta\in\Omega_{\mathrm{FOV}}\subset\mathbb{R}^2$ denotes angular sky coordinates supported in the field of view $\Omega_{\mathrm{FOV}}$.
VLBI does not directly construct an image but instead measures complex-valued sparse samples of the image’s 2D Fourier transform (often called \textit{visibilities}). 

Physically, a pair of telescopes acts as an interferometer: radiation arriving from the direction $\vartheta$ reaches the two sites with a small geometric path delay that depends on the physical separation vector between the telescopes. 
This delay induces a relative phase shift between received waveforms that varies sinusoidally across the sky, projecting an interference pattern onto the field of view. 
Correlating the recorded electric fields therefore yields one complex number per telescope pair and time: an inner product of the sky brightness and a complex sinusoid whose spatial frequency is set by the separation between telescopes.

Let $i$ and $j$ index two telescopes observed at time $t$, and denote by $w_{(i,j)}^t=(u_{(i,j)}^t,v_{(i,j)}^t)\in\mathbb{R}^2$ the \emph{dimensionless baseline}: the separation vector between the telescopes projected onto the 2D angular plane of the sky. Then to obtain a measurement, we use the van Cittert--Zernike theorem~\cite{vancittert1934wahrscheinliche, zernike1938concept} which states that the \emph{ideal visibility} $I_{(i,j)}^t(m)$ corresponding to this time-dependent baseline is the spatial Fourier coefficient of $m$ at that baseline:
\begin{equation}\label{eq:vcz_thm}
    I_{(i,j)}^t(m)
    =
    \int_{\Omega_{\mathrm{FOV}}}
    m(\vartheta)\,
    e^{-\mathrm{i}2\pi \vartheta\,\cdot\,w_{(i,j)}^t}
    \, d\vartheta .
\end{equation}
Thus, each telescope pair and time provides one data point $I_{(i,j)}^t(m)\in\mathbb{C}$ in the $(u,v)$-plane. As the Earth rotates, the projected baselines change with $t$, yielding additional Fourier samples, though the $(u,v)$ coverage remains highly sparse relative to the number of image pixels.

\paragraph{Challenges}
In practice, the measured complex visibility is corrupted by telescope-dependent calibration errors and thermal noise \cite{ehtc2019third, sun2024provable}. A common measurement model for \emph{observed visibilities} is
\begin{equation}\label{eq:corrupted_vis}
    V_{(i,j)}^t
    =
    g_i^t g_j^t \, e^{-\mathrm{i}(\phi_i^t-\phi_j^t)}\, I_{(i,j)}^t(m)
    + \eta_{(i,j)}^t ,
\end{equation}
for three sources of error: amplitude gains $g_i^t,g_j^t$, phase errors $\phi_i^t,\phi_j^t$ (e.g., due to atmospheric turbulence and clock offsets), and additive thermal Gaussian noise $\eta_{(i,j)}^t$. Because the per-telescope phase factors can dominate the visibility phase, direct use of $V_{(i,j)}^t$ is often unreliable. Instead, VLBI imaging commonly relies on \emph{closure quantities} that cancel telescope-dependent errors \cite{blackburn2020closure}.

For any telescope triple $(i,j,k)$, the \emph{closure phase} is defined as
\begin{equation}\label{eq:closure_phase}
    y_{\mathrm{cp}}^{t,(i,j,k)}
    =
    \calF^{t,(i,j,k)}_{\mathrm{cp}}(m)
    =
    \angle\!\left(
    V_{(i,j)}^t\, V_{(j,k)}^t\, V_{(k,i)}^t
    \right) \in \mathbb{R}.
\end{equation}
Similarly, for a telescope quadruple $(i,j,k,l)$, the \emph{log closure amplitude} is
\begin{equation}\label{eq:log_closure_amp}
    y_{\mathrm{logca}}^{t,(i,j,k,l)}
    =
    \calF^{t,(i,j,k,l)}_{\mathrm{logca}}(m)
    =
    \log\!\left(
    \frac{|V_{(i,j)}^t|\,|V_{(k,l)}^t|}
         {|V_{(i,k)}^t|\,|V_{(j,l)}^t|}
    \right)\in \mathbb{R},
\end{equation}
which removes the unknown gains $\{g_\cdot^t\}$ under mild assumptions. Finally, to anchor the overall intensity scale, we constrain the \emph{total flux}
\begin{equation}\label{eq:flux}
    y_{\mathrm{flux}} = \calF_{\mathrm{flux}}(m)
    = \int_{\Omega_{\mathrm{FOV}}} m(\vartheta)\, d\vartheta\in\mathbb{R}.
\end{equation}

Collecting these observables over available times and telescope combinations yields the nonlinear forward map
\begin{equation}\label{eq:bh_forward}
    y = \calF(m) = \left(
    \left\{\calF^{t,(i,j,k)}_{\mathrm{cp}}(m)\right\},\, 
    \left\{\calF^{t,(i,j,k,l)}_{\mathrm{logca}}(m)\right\},\, 
    \calF_{\mathrm{flux}}(m)
    \right).
\end{equation}
Since measurements are nonlinear closure quantities of $V_{(i,j)}^t$, the inverse problem $m \approx \calF^{-1}(y)$ is inherently nonconvex and can admit multiple plausible solutions, especially under sparse $(u,v)$ coverage and super-resolution requirements.

\paragraph{Adjoint}
Due to the underlying Fourier fringe kernels and sparse $(u,v)$ sampling, gradients computed on closure residuals generally appear as global oscillatory patterns rather than image-like features, even when they provide correct descent directions. 
Therefore, to facilitate successful reconstructions with \ourname, we choose the initializer $x_0=a(y)\equiv\tilde{\mu}_m$, where $\tilde{\mu}_m$ is the standardized per-pixel mean of the training dataset (namely, re-normalized to have zero spatial mean and unit spatial variance). 
While a simple, \textit{data-independent} initialization, it nonetheless provides an inductive bias toward physically plausible black-hole image manifolds by encoding a strong ring-shaped prior. 

To demonstrate a hypothetical setting in which we have access to a more informative data-dependent coupling, we also consider an ``oracle'' initialization obtained by Gaussian-blurring the ground-truth medium:
\begin{equation}\label{eqn:smooth_blackhole}
    \rho(x_0\mid x_1) = \delta_{\left(x_1 \,*\, \mathcal{G}_\sigma\right)},
\end{equation}
where $\mathcal{G}_\sigma = \mathcal{N}(0,\sigma^2 I)$ is an isotropic Gaussian kernel and $*$ denotes convolution. 
Here $\sigma$ controls the spatial smoothing level, and we choose $\sigma=3$ (for images of resolution $64\times 64$), which removes fine-scale details while preserving the large-scale structure.

\subsubsection{Benchmark Dataset}
Since the General Relativistic Magnetohydrodynamic
(GRMHD) simulated black hole images training dataset used by InverseBench is not publicly available, we generate our own dataset of the same size -- 50,000 black holes at resolution $64\times 64$ -- using the pretrained diffusion prior made available by InverseBench. However, we evaluate our models using the validation and test sets provided by InverseBench~\cite{zheng2025inversebench,inversebench_data} (which were also generated with a pretrained diffusion prior). 

\paragraph{Methods}
InverseBench evaluates two classical optimization algorithms (SMILI~\cite{Akiyama_2017}, EHT-Imaging~\cite{Chael_2016,Chael_2018}) and six PnP-DP methods on this problem.
Their diffusion priors use a pretrained NCSN++ U-net~\cite{song2021scorebased}; however, we find that \ourname achieves superior results even when instantiated with a smaller convolutional architecture, similar to that used in the frequency-domain inverse scattering problems.
We also find that the gradient of the closure quantities, though generally appearing as a global oscillatory pattern, still provides useful physical information to the network and empirically helps reduce the $\chi^2$ data misfit statistics. Therefore, we concatenate the gradient $\nabla_{x_0}\chi^2(x_0)$ to the network as an additional input channel. Measurement data is pre-processed with a small MLP and then embedded to modulate the network features via FiLM conditioning \cite{perez2018film}.

\subsubsection{Results}
\begin{figure*}
    \centering
    \includegraphics[width=\linewidth]{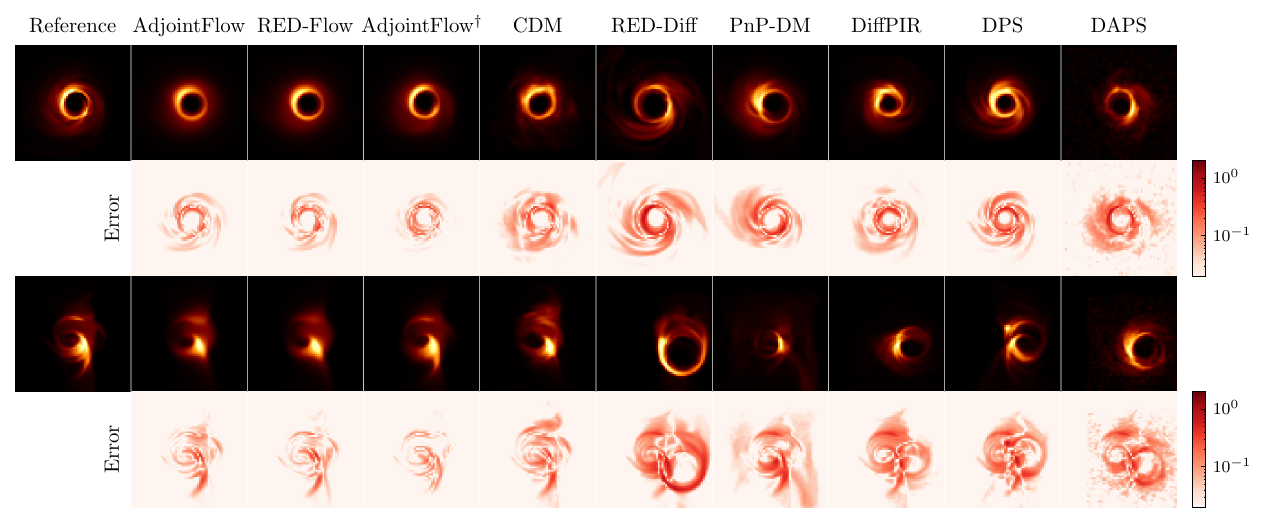}
    \caption{Comparisons of \ourname against different diffusion-based methods for black hole imaging with 3\% observation ratio. Even with the simple ring-shaped initialization, \ourname properly matches the shape of both simpler ring-like black holes as well as those with a more exotic accretion disk.}
    \label{fig:blackhole_comparison}
\end{figure*}

\begin{table}
\caption{We present the results for black hole imaging from InverseBench~\cite{zheng2025inversebench} with visibility noise (gain, phase, and thermal) added via EHT library. \ourname$^\dagger$ indicates that we use as the physics-based approximation the ground-truth image blurred with a Gaussian filter ($\sigma=3$). While the baseline \ourname improves reconstruction accuracy, it struggles to achieve high data consistency. This issue is resolved by either (a) inference-time guidance via \ournameRED, or (b) a more informative coupling with \ourname$^\dagger$, both of which significantly improve $\chi^2$ data consistency to be competitive with state-of-the-art. The top panel reports reconstruction quality and the bottom panel data consistency. Bold indicates best performance of non-oracle models.}
\label{table:blackhole_results}
\centering
\small
\setlength{\tabcolsep}{1.8pt}
\textit{Reconstruction quality}\\[2pt]
\begin{tabular}{lcc cc cc}
\toprule
Observation time ratio &
\multicolumn{2}{c}{\textbf{3\%}} &
\multicolumn{2}{c}{\textbf{10\%}} &
\multicolumn{2}{c}{\textbf{100\%}} \\
\cmidrule(lr){2-3}\cmidrule(lr){4-5}\cmidrule(lr){6-7}
\textbf{Method}
  & \textbf{PSNR} $\uparrow$ & \textbf{Blur PSNR} $\uparrow$ & \textbf{PSNR} $\uparrow$ & \textbf{Blur PSNR} $\uparrow$ & \textbf{PSNR} $\uparrow$ & \textbf{Blur PSNR} $\uparrow$ \\
\midrule
\multicolumn{7}{l}{\textbf{Traditional}} \\
SMILI
  & 18.51 (1.39) & 23.08 (2.12) & 20.85 (2.90) & 25.24 (3.86) & 22.67 (3.13) & 27.79 (4.02) \\
EHT-Imaging
  & 21.72 (3.39) & 25.66 (5.04) & 22.67 (3.46) & 26.66 (3.93) & 24.28 (3.63) & 28.57 (4.52) \\
\midrule
\multicolumn{7}{l}{\textbf{PnP Diffusion Prior}} \\
DPS
  & 24.20 (3.72) & 30.83 (5.58) & 24.36 (3.72) & 30.79 (5.75) & 25.86 (3.90) & 32.94 (6.19) \\
LGD
  & 22.51 (3.76) & 28.50 (5.49) & 22.08 (3.75) & 27.48 (5.09) & 21.22 (3.64) & 26.06 (4.98) \\
RED-Diff
  & 20.74 (2.62) & 26.10 (3.35) & 22.53 (3.02) & 27.67 (4.53) & 23.77 (4.13) & 29.13 (6.22) \\
PnP-DM
  & 24.25 (3.45) & 30.49 (4.93) & 24.57 (3.47) & 30.80 (5.22) & 26.07 (3.70) & 32.88 (6.02) \\
DAPS
  & 23.54 (3.28) & 29.48 (4.88) & 23.99 (3.56) & 30.10 (5.13) & 25.60 (3.64) & 32.78 (5.68) \\
DiffPIR
  & 24.12 (3.25) & 30.45 (4.88) & 23.84 (3.59) & 30.04 (5.03) & 25.01 (4.64) & 31.86 (6.56) \\
\midrule
\multicolumn{7}{l}{\textbf{Conditional Diffusion}} \\
CDM
  & 26.86 (3.60) & 35.43 (5.32) & 26.98 (3.67) & 35.96 (5.76) & 26.98 (3.61) & 35.56 (5.39) \\
\midrule
\multicolumn{7}{l}{\textbf{Ours}} \\
\ourname
  & \textbf{28.78} (3.44) & \textbf{38.28} (5.23) & 28.35 (3.38) & \textbf{37.81} (5.42) & 29.12 (3.47) & 38.91 (5.25) \\
\ournameRED
  & 28.54 (3.81) & 37.62 (5.98) & \textbf{28.55} (3.72) & 37.68 (6.08) & \textbf{29.63} (3.50) & \textbf{39.69} (5.51) \\
\ourname$^\dagger$
  & 31.28 (2.37) & 42.96 (2.20) & 31.20 (2.38) & 42.94 (2.24) & 31.26 (2.37) & 42.92 (2.25) \\
\bottomrule
\end{tabular}
 
\vspace{8pt}
\textit{Data consistency}\\[2pt]
\begin{tabular}{lcc cc cc}
\toprule
Observation time ratio &
\multicolumn{2}{c}{\textbf{3\%}} &
\multicolumn{2}{c}{\textbf{10\%}} &
\multicolumn{2}{c}{\textbf{100\%}} \\
\cmidrule(lr){2-3}\cmidrule(lr){4-5}\cmidrule(lr){6-7}
\textbf{Method}
  & $\tilde{\chi}^2_{cp}$ $\downarrow$ & $\tilde{\chi}^2_{\mathrm{logca}}$ $\downarrow$ & $\tilde{\chi}^2_{cp}$ $\downarrow$ & $\tilde{\chi}^2_{\mathrm{logca}}$ $\downarrow$ & $\tilde{\chi}^2_{cp}$ $\downarrow$ & $\tilde{\chi}^2_{\mathrm{logca}}$ $\downarrow$ \\
\midrule
\multicolumn{7}{l}{\textbf{Traditional}} \\
SMILI
  & \textbf{1.478} (0.428) & 4.348 (3.827) & 1.209 (0.169) & 21.788 (12.491) & 1.878 (0.952) & 17.612 (10.299) \\
EHT-Imaging
  & 1.507 (0.485) & 1.695 (0.539) & 1.166 (0.156) & \textbf{1.240} (0.205) & 1.251 (0.250) & 1.259 (0.316) \\
\midrule
\multicolumn{7}{l}{\textbf{PnP Diffusion Prior}} \\
DPS
  & 8.024 (24.336) & 5.007 (5.750) & 13.052 (43.087) & 6.614 (26.789) & 8.759 (37.784) & 5.456 (24.185) \\
LGD
  & 15.825 (16.838) & 12.862 (12.663) & 10.775 (21.684) & 13.375 (56.397) & 13.239 (17.231) & 13.233 (39.107) \\
RED-Diff
  & 6.713 (6.925) & 9.128 (19.052) & 2.488 (2.925) & 4.916 (13.221) & 1.853 (0.938) & 2.050 (2.361) \\
PnP-DM
  & 2.201 (1.352) & 1.668 (0.551) & 1.433 (0.417) & 1.336 (0.478) & 1.311 (0.195) & 1.199 (0.221) \\
DAPS
  & 3.647 (3.287) & 2.329 (1.354) & 1.545 (0.705) & 2.253 (9.903) & 1.300 (0.324) & 1.229 (0.532) \\
DiffPIR
  & 14.085 (14.105) & 10.545 (8.860) & 5.374 (3.733) & 5.205 (5.556) & 3.271 (1.623) & 2.970 (1.202) \\
\midrule
\multicolumn{7}{l}{\textbf{Conditional Diffusion}} \\
CDM
  & 21.811 (46.440) & 13.202 (31.235) & 16.200 (25.382) & 13.978 (23.813) & 19.241 (23.680) & 13.531 (12.885) \\
\midrule
\multicolumn{7}{l}{\textbf{Ours}} \\
\ourname
  & 8.118 (18.273) & 7.868 (20.607) & 15.743 (33.948) & 18.372 (71.714) & 11.734 (16.559) & 16.302 (49.840) \\
\ournameRED
  & 1.483 (0.545) & \textbf{1.539} (0.726) & \textbf{1.155} (0.143) & 1.302 (0.254) & \textbf{1.130} (0.123) & \textbf{1.196} (0.177) \\
\ourname$^\dagger$
  & 1.515 (0.520) & 1.668 (0.630) & 1.293 (0.253) & 1.302 (0.295) & 1.294 (0.241) & 1.371 (0.363) \\
\bottomrule
\end{tabular}
\end{table}

Results are presented in~Table~\ref{table:blackhole_results}, with illustrative examples shown in~Figure~\ref{fig:blackhole_comparison}. 
We find that over all levels of measurement sparsity, \ourname performs substantially better than previous methods in terms of PSNR reconstruction quality. 
However, we also find that the base \ourname model using data-independent initialization $x_0\equiv\tilde{\mu}_m$ markedly struggles to produce reconstructions consistent with the measurement data, as evidenced by large $\tilde\chi^2$ metrics.
CDM also outperforms PnP-DP methods on reconstruction quality, though not to the extent of \ourname, yet similarly struggles with data consistency.

To overcome this challenge, we demonstrate that adding inference-time guidance from the forward model via \ournameRED is sufficient to improve data consistency. 
In fact, \ournameRED achieves the best data consistency in all but two instances, for which it is a close second. 
Furthermore, using the blurred-ground-truth initialization~\eqref{eqn:smooth_blackhole} produces high-quality \textit{and} data-consistent reconstructions \textit{without} added inference-time costs. 
However, we recall that this is an oracle diagnostic unavailable from the observation alone and is therefore not a deployable initialization in practice. 
Therefore, it serves primarily as a demonstration of the possible gains from a more informative initializer, lending support to the empirical benefits of choosing a physically informative initialization $a(y)$.

\section{Ablations}
\label{sec:ablations}
To verify the design of \ourname, we ablate the choices we made given the motivation of~Section~\ref{sec:method}. We focus on the nonlinear inverse scattering problem -- which supplies four distinct datasets covering a range of difficulties and challenges -- as a strong canonical baseline on which to test sampling methods, interpolant designs, reference distributions, and network parameterizations.
To isolate the choice of initialization for \ourname, we additionally consider the linear inverse scattering problem, as it admits inexpensive physics-based approximations of varying accuracy (namely, both an adjoint and a regularized pseudoinverse).

\paragraph{Sampling methods and learned flow}
As discussed in Section~\ref{sec:stochastic_interpolants}, the SI framework provides flexibility with several modeling choices: how to add noise, how to generate samples (deterministically with ODEs~\eqref{eq:flow_ode} or stochastically with SDEs~\eqref{eq:flow_sde}), and how to schedule the added noise.

We test how ODE and SDE sampling methods, different learned fields, as well as different choices for the noise bridge $\gamma_t$ affect performance, especially when using the physics-based reference distribution $\rho_0=\delta_{a(y)}$. 

Specifically, as enabled by the velocity decomposition~\eqref{eq:velocity_decomposition}, we evaluate both learning the standard SI velocity field $b_t$ directly, or learning $v_t$ and $\mathcal{D}_t$ and then combining them to form the standard velocity.
We also investigate three choices of the noise-bridge term: $\gamma_t=\sqrt{t(1-t)}$, $\gamma_t=\sin^2(\pi t)$, and $\gamma_t=1-t$.
For the schedule labeled $\gamma_t=1-t$ the interpolant is
$x_t=(1-t)x_0+tx_1$ with
$x_0\sim\mathcal{N}(a(y),\sigma^2I)$, namely, a Gaussian centered at the physics-based initialization.
The standard SI ODE and SDE use this Gaussian initialization, whereas
the noise-free-velocity ODE still starts deterministically at $x_0=a(y)$.

\begin{figure*}[t]
    \centering
    \includegraphics[width=\linewidth]{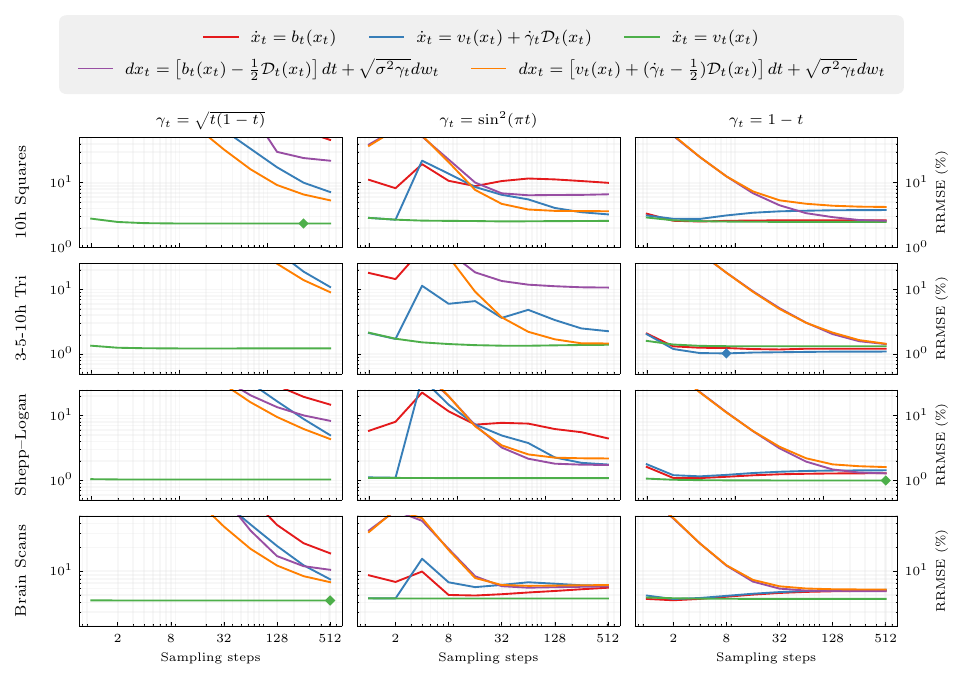}
    \caption{We compare accuracy (RRMSE) and efficiency (number of sampling steps) of different stochastic interpolants on the four nonlinear inverse scattering datasets. The best performance on each dataset is labeled with a diamond. We do not use TP parameterizations for this ablation. For SDE sampling we choose $\epsilon_t=\sigma^2\gamma_t/2$ in~\eqref{eq:flow_sde} to cancel the explicit $1/\gamma_t$ factor in the score correction. For $\gamma_t=1-t$, the noise-free-velocity ODE uses a deterministic initialization, while the other samplers use a Gaussian initialization. We observe that sampling with $\dot{x}_t=v_t(x_t)$ performs consistently well across datasets and only requires a small number of sampling steps for near-optimal performance.}

    \label{fig:sampling_methods}
\end{figure*}

We evaluate performance across a range of sampling timesteps and noise-bridge scale $\sigma\in(0,2]$, and present the accuracy-efficiency curve for the best performing choice of $\sigma$ on each dataset in~Figure~\ref{fig:sampling_methods}.
Among the evaluated configurations, the noise-free-velocity ODE achieves highest accuracy on three of
the four datasets.
Notably, the noise-free velocity's performance is largely decoupled from the choice of both noise bridge $\gamma_t$ and number of sampling steps, enabling more flexibility in interpolant design and accelerated sampling.

\paragraph{Initial distribution and parameterization}
We compare the standard flow-matching interpolant $I_t = (1-t)x_0 + tx_1$ with reference distribution $\rho_0=\calN(0, I)$ (an ``uncoupled SI'') with \ourname's physics-based initialization $\rho_0=\delta_{a(y)}$ (a ``coupled SI'').
We also evaluate both interpolant designs with and without their respective TP parameterizations -- \eqref{eq:uncoupled_tp} for the uncoupled SI, and~\eqref{eq:factorized_flow} for the coupled SI. 
Following~\cite{li2025letdenoise}, we clip the denominator of \eqref{eq:uncoupled_tp} when it goes below 0.05 to avoid numerical instability during training.

We train all interpolants with the same architecture and training procedure, and evaluate them on the four nonlinear inverse scattering datasets.
For all models, inference is conducted with a fixed 8 sampling steps.

We present results in~Table~\ref{table:algo_comp_main}, which illustrates a crucial interaction between the construction of reference distribution and network parameterization. 
Without TP, the coupled SI has lower accuracy than the uncoupled SI on three of four datasets.
With TP, this comparison reverses: the coupled model achieves higher accuracy on every dataset, and outperforms the base uncoupled SI on all but \textit{3--5--10h Triangles}.
The effect of TP likewise depends on the initialization: on all four datasets, it improves accuracy for the coupled SI  but worsens it for the uncoupled SI. 
Crucially, the DD-TP which improves the coupled SI so much is \textit{only possible} with the singular reference distribution used by \ourname; the standard TP from~\eqref{eq:uncoupled_tp} includes a singular denominator and thus requires additional truncation to remain stable, yet rendering it less effective.

\begin{table}[!htbp]
\caption{Comparison of reference distribution and model parameterization on the four nonlinear inverse scattering datasets, with mean and standard deviation of RRMSE across the test set shown. We fix the number of sampling steps at 8 for all models. Without TP, we find that a coupled interpolant does not meaningfully improve results on these datasets, demonstrating the practical necessity of the DD-TP parameterization. Bold indicates the best result on each dataset.}
\label{table:algo_comp_main}
\resizebox{\textwidth}{!}{
\begin{tabular}{lcccc}
\toprule
& \multicolumn{4}{c}{\textbf{RRMSE} (\%) $\downarrow$} \\
\cmidrule(lr){2-5}
\textbf{Method} & \textbf{Shepp--Logan} & \textbf{3--5--10h Triangles} & \textbf{10h Overlapping Squares} & \textbf{Brain Scans} \\
\midrule
Uncoupled SI + $b_t$
& 0.972 (0.395) & \textbf{0.590} (1.599) & 1.808 (2.403) & 4.711 (1.250) \\
Uncoupled SI + $b_t$ + TP
& 1.097 (0.504) & 1.356 (2.871) & 2.649 (2.537) & 4.887 (1.282) \\
Coupled SI + $v_t$
& 1.031 (0.265) & 1.228 (1.473) & 2.345 (2.159) & 4.233 (1.020) \\
Coupled SI + $v_t$ + DD-TP (\textbf{\ourname})
& \textbf{0.379} (0.158) & 1.108 (2.861) & \textbf{1.143} (1.686) & \textbf{3.421} (0.809) \\
\bottomrule
\end{tabular}
}
\vskip -0.1in
\end{table}

Together, the physics-based initialization and DD-TP achieve the best accuracy on three of the four datasets.
These results suggest that the benefits of the method depend on the \textit{combined construction of initialization and network parameterization}, rather than either modification in isolation.
Namely, the known physics-based coupling allows DD-TP to express the velocity as a residual between the predicted reconstruction and the prescribed initialization. 
Incorporating this known contribution analytically may therefore help explain why TP is particularly effective in the coupled setting. 

We note that the \textit{3--5--10h Triangles} dataset is the one case in both ablation studies in which the noise-free velocity does not achieve the best performance.
This suggests that the adjoint-based initialization may not be as informative for this dataset as for the others.
Qualitatively, one may note that the back-projections from the low-frequency adjoint in~Figure~\ref{fig:sub-3510tri} fail to fully capture high-frequency details of the smallest triangles. 
Therefore, a more informative initialization for \textit{3--5--10h Triangles} may reduce the observed performance gap; we leave this exploration to future work.

\begin{table}[!htbp]
\caption{We present the results of \ourname for linear inverse scattering with four choices of initialization $a(y)$, three different levels of measurement sparsity, and noise level $\sigma_y=10^{-4}$ following~\cite{zheng2025inversebench} (bold result indicates best for each receiver count). All models share the architecture and training procedure of~Section~\ref{sec:lin_frequency_inverse_scattering} and differ only in $a(y)$; the pseudoinverse corresponds to \ourname in~Table~\ref{table:results_linear_inv_scattering_full}. Within each block, the first two initializations are data-independent and the last two are data-dependent. $\mathcal W_2$ is the empirical transport distance $\mathcal W_2(\delta_{a(y)},\,\delta_m)=\|a(y) - m\|_2$ between singular endpoints (\ref{app:transport_distances}), averaged over the test set.}
\label{table:init_ablation}
\centering
\small
\begin{tabular}{lcccc}
\toprule
\textbf{Initialization}
& {\boldmath$\mathcal W_2$} $\downarrow$ & \textbf{PSNR} $\uparrow$ & \textbf{SSIM} $\uparrow$ & \textbf{Meas err} (\%) $\downarrow$ \\
\midrule
\multicolumn{5}{l}{\textbf{60 receivers}} \\
Constant
& 127.0 (16.3) & 13.007 (2.830) & 0.436 (0.208) & 38.05 (8.67) \\
Training mean
& 157.1 (20.9) & 17.025 (2.675) & 0.631 (0.156) & 14.74 (2.31) \\
Adjoint
& 110.2 (18.4) & 25.341 (6.155) & 0.910 (0.066) & 5.50 (1.85) \\
Pseudoinverse
& \textbf{53.5} (10.2) & \textbf{25.460} (4.969) & \textbf{0.920} (0.054) & \textbf{4.70} (0.99) \\
\midrule
\multicolumn{5}{l}{\textbf{180 receivers}} \\
Constant
& 127.0 (16.3) & 13.233 (2.667) & 0.518 (0.204) & 42.87 (10.26) \\
Training mean
& 157.1 (20.9) & 18.687 (2.506) & 0.701 (0.127) & 11.13 (1.72) \\
Adjoint
& 98.6 (19.7) & 29.913 (4.819) & 0.966 (0.024) & 2.76 (0.50) \\
Pseudoinverse
& \textbf{23.8} (4.9) & \textbf{32.459} (3.935) & \textbf{0.980} (0.013) & \textbf{1.87} (0.20) \\
\midrule
\multicolumn{5}{l}{\textbf{360 receivers}} \\
Constant
& 127.0 (16.3) & 14.314 (2.565) & 0.531 (0.199) & 33.09 (6.46) \\
Training mean
& 157.1 (20.9) & 19.546 (2.554) & 0.735 (0.115) & 9.64 (1.44) \\
Adjoint
& 97.6 (19.1) & 30.056 (4.831) & 0.966 (0.024) & 2.74 (0.49) \\
Pseudoinverse
& \textbf{22.6} (4.7) & \textbf{32.797} (3.966) & \textbf{0.981} (0.013) & \textbf{1.86} (0.21) \\
\bottomrule
\end{tabular}
\end{table}

\paragraph{Choice of initialization}
The \textit{3--5--10h Triangles} results suggest that the benefit of \ourname depends on how informative its initialization is. To test this directly, we train \ourname on the linear inverse scattering problem of~Section~\ref{sec:lin_frequency_inverse_scattering} with four choices of $a(y)$, holding the architecture, training procedure, and sampler fixed. Two initializations are data-dependent: the adjoint $a(y)=F^*y$ and the regularized pseudoinverse $a(y)=F_\epsilon^\dagger y$ used in the main experiments. The other two initializations are data-independent: the constant $a(y) = 0$, and the standardized training mean $a(y)=\tilde{\mu}_m$ equivalent to the initialization used for black hole imaging. We report empirical transport distance $\mathcal W_2$ (\ref{app:transport_distances}) computed over the test sets alongside reconstruction metrics in~Table~\ref{table:init_ablation}.
 
Data-dependent initializations demonstrate a clear advantage, vastly outperforming data-independent ones across all metrics: for example, they improve reconstructions by 8 to 19\,dB in PSNR at every receiver count. Consistent with~Proposition~\ref{prop:reduced_transport_case_linear}, the adjoint also lies closer to the target than the constant initialization at every receiver count. The pseudoinverse initialization further reduces transport distance between endpoints and achieves the best performance at every receiver count.
 
Between the two data-independent initializations, however, $a(y)=\tilde\mu_m$ improves performance over $a(y)=0$ even though it lies farther from the target (a consequence of standardization of the unscaled training mean), demonstrating that transport distance by itself is not entirely predictive of performance. 
We posit that, despite the increased transport distance, the resulting spatial structure of $\tilde\mu_m$ may nonetheless be easier for the network to refine than a constant image. 

Together, these results support choosing a physically informative initialization: incorporating the measurements has the largest effect, and between the data-dependent initializations, a more accurate one further improves reconstruction.

\paragraph{Additional ablations} 
Additional ablations and tests covering performance in noisy or uncertain environments, sample complexity, empirical transport distances, and training techniques are covered in~\ref{app:ablations}.

\section{Related Work}
\label{sec:related}

\subsection{Classical Approaches}
Optimization-based reconstruction has long been a standard technique for solving many classical scientific inverse problems.
In this setting, one seeks to minimize a data misfit together with regularization to stabilize the inversion in ill-posed and noisy regimes~\cite{tikhonov1977solutions,engl1996regularization}.
We focus on this optimization viewpoint and refer the reader to~\cite{cakoni2005qualitative,colton2003linear,colton1998inverse} for reviews of qualitative and analysis-driven approaches.

In particular, wave-based inverse problems constitute an important and challenging subclass, where the goal is to reconstruct medium or object properties -- such as the refractive index, permittivity, or wavespeed -- from boundary or diffraction measurements of scattered wave fields~\cite{colton1998inverse,Virieux_Operto}.

With improved numerical solutions of the underlying PDEs, full waveform inversion (FWI) emerged in the 1980s~\cite{Tarantola:Inversion_of_seismic_reflection_data_in_the_acoustic_approximation} and became a central paradigm in seismic imaging~\cite{Pratt:1999,virieux2014introduction}.
FWI recasts inversion as a PDE-constrained optimization problem, seeking parameters that minimize the mismatch between measured data and synthetic data generated by solving the governing PDE.
Its primary advantage over ray-based methods is its ability to model the full wavefield, thereby capturing finite-frequency, diffraction, and scattering effects neglected by geometric-optics approximations.
This increased physical fidelity comes at the expense of substantial computational and optimization challenges.

Despite its success, particularly in exploration geophysics, FWI faces several well-known obstacles.
First, the per-iteration cost is dominated by repeated forward and adjoint PDE solves, making high-resolution inversion computationally expensive even with modern fast wave solvers such as sweeping and domain-decomposition-style methods~\cite{EngquistYing:Sweeping_PML,ZepedaDemanet:the_method_of_polarized_traces}.
Second, the objective is highly nonconvex, and convergence to spurious local minima (cycle-skipping) is common, especially in the absence of low-frequency data~\cite{Leeuwen_Herrmann:2013}.
Common mitigation strategies include adding regularization~\cite{Symes_Carazzone:1991,Leeuwen_Herrmann:2013}, employing multiscale or frequency-continuation strategies~\cite{bunks1995multiscale,chen1997inverse}, and extrapolating or interpolating missing low frequencies~\cite{li2016full}.
Third, recovery of fine-scale structure is constrained by wavelength, acquisition geometry, and diffraction, even under favorable noise levels and sampling~\cite{Fichtner2011}.

\subsection{Deterministic ML Approaches}
Motivated by the limitations of classical optimization-based methods, a large body of work has explored learning-based solvers that replace or accelerate inversion by learning explicit mappings from measurements to reconstructions using supervised regression.
Representative approaches include learned representation transforms~\cite{mccann2017convolutional} and model-based deep learning architectures such as MoDL, which combines a learned CNN prior with alternating data-consistency updates in an unrolled iterative reconstruction framework~\cite{Aggarwal2019MoDL}.

In the context of wave-based inverse scattering, deterministic ML approaches typically employ supervised neural networks to regress unknown parameters directly from measured scattered fields~\cite{Chen2020Review}.
Learned primal-dual methods unroll gradient-like optimization schemes while replacing hand-crafted update rules with learned operators that receive data-misfit information~\cite{Adler_2017,Adler_2018}.
Other architectures explicitly seek to exploit structure and symmetries in the wave-based inverse operator itself, such as SwitchNet~\cite{KhooYingSwitchNet:2019}, WideBNet~\cite{li2022butterfly}, EquiNet, and B-EquiNet~\cite{zhang2024_equivariant}.

\subsection{Transport-Based Generative Models}
Bayesian formulations provide a principled framework for uncertainty quantification by modeling the posterior distribution of the unknown parameters conditioned on the measurements~\cite{stuart2010bayesian,kaipio2006statistical}.
Posterior inference can be carried out using sampling methods such as Markov chain Monte Carlo or Langevin-type dynamics, or through variational approximations, with the forward operator incorporated through the likelihood.
However, repeated likelihood evaluations can become computationally expensive when the forward model requires solving a large-scale PDE.

Generative models provide an alternative by shifting much of this computational burden to an offline training stage and learning a distribution over reconstructions rather than individual point estimates.
In particular, conditional diffusion models aim to sample from the unknown posterior directly, and have demonstrated strong performance in applications such as MRI, elastography, and computational imaging~\cite{Dasgupta:2026,Zhang2025_BPD,korkmaz2024selfsupervised,Liu_2026,Dasgupta_2025}.

Diffusion models are defined by a stochastic process which transforms samples from a simple base distribution -- commonly an isotropic Gaussian -- into samples from the target conditional distribution~\cite{song2019generative,ho2020ddpm}.
By approximating the score function of a reverse-time denoising process~\cite{ANDERSON1982313}, diffusion models allow generation of new samples via numerically solving the reverse-time SDE.
However, diffusion samplers can require hundreds of network evaluations for high-quality reconstructions, which can be restrictive when inference-time efficiency is important~\cite{song2021scorebased,ho2020ddpm}.
It was subsequently realized that the sampling process could be reframed as solving a reverse-time ODE, providing substantial speedups with little loss of accuracy compared to SDEs~\cite{song2021ddim}.
Taking inspiration from optimal transport, flow matching~\cite{lipman2023flow,liu2023reflow, albergo2023si} expands upon this approach by directly learning transport velocities along prescribed paths which efficiently transport mass between endpoint distributions.

Stochastic interpolants (SI)~\cite{albergo2023si, albergo2025stochastic} offer a unifying framework for such transport-based generative models.
The SI framework additionally permits general, potentially dependent couplings between its endpoint distributions~\cite{albergo2024couplings}, providing flexibility beyond the conventional independent Gaussian source used by standard diffusion and flow models.

Ensuing work has aimed to further improve generative efficiency by constructing straighter or otherwise better-conditioned transport trajectories, thereby reducing the numerical cost of sampling even further~\cite{liu2023reflow,lee2024improving}.
Distillation provides another route to few-step inference by compressing a trained diffusion or flow sampler into a student requiring fewer network evaluations~\cite{salimans2022progressive,zhao2024cosign,boffi2025buildconsistencymodellearning}.
Our approach is complementary in that we shorten and structure the transport path using information from the governing inverse problem itself.

Flow matching has also been applied to scientific inverse problems such as seismic FWI~\cite{paulwitz2026probabilistic}, including within online inversion~\cite{huang2026selfflow} and to refine an existing FWI reconstruction~\cite{zhang2026postfwi}.
A recent study~\cite{dasgupta2026conditional} examines the behavior of conditional flow matching in the physics-constrained, finite-training-data regime characteristic of scientific inversion, and compares Gaussian sources with sources sampled from the parameter prior.
More closely related to our setting, DAWN-FM~\cite{dawnfm2026} applies flow matching to inverse problems by conditioning the learned velocity field on a physics-based data embedding, such as a back-projection.
However, the physics-based approximation enters DAWN-FM as conditioning information while the transport itself retains the standard Gaussian reference distribution.

\subsection{Alternative Endpoints, Couplings, and Bridges in Generative Models}
A related line of work replaces the standard noise-to-image paradigm by learning transport directly between informative endpoint distributions.
Within image restoration and translation, I$^2$SB~\cite{liu2023i2sb} and BBDM~\cite{li2023bbdm} formulate generation as a stochastic bridge between degraded and clean image distributions.
InDI~\cite{delbracio2023inversion} instead learns a direct iterative corrupted-to-clean inversion map rather than a conditional denoising process initialized from pure noise.
Mean-reverting SDE models retain a stochastic diffusion formulation but center the forward process around the degraded observation itself~\cite{luo2023meanreverting}.
DBIM~\cite{zheng2024dbim} adapts design principles of denoising diffusion implicit
models~\cite{song2021ddim} to further accelerate the inference speed of diffusion bridge models on image restoration tasks.

Informative-endpoint constructions also extend to inverse problems whose measurements and unknowns occupy different mathematical spaces, where inexpensive approximate reconstructions -- for example via back-projection or linearized adjoints -- can provide useful parameter-space information.
Jiang et al.~\cite{jiang2024embridge} employ diffusion Schr\"odinger bridges to transport between electromagnetic property maps and channel representations, using a learned autoencoder network to rectify the dimensional mismatch.
Projection Embedded Diffusion Bridge initializes CT reconstruction from a filtered back-projection image and incorporates the projection measurements into the reverse dynamics to improve data consistency~\cite{wang2025projectionembeddeddiffusionbridge}.
Related work also considers learning from indirect corrupted measurements through self-consistent stochastic interpolants, using a black-box forward simulator to enforce consistency between generated samples and observed data~\cite{Han2025SI}.
\ourname extends this perspective to nonlinear scientific inverse problems and analyzes how a physics-based initialization interacts with the choice of reconstruction dynamics, particularly when the conditional posterior is highly concentrated.

Singular endpoint distributions have also been considered in recent works.
floq~\cite{agrawalla2026floq} employs flow matching with a Dirac target distribution in the context of Q-learning. They demonstrate the benefits of iterative computation with dense supervision brought by a generative framework, a perspective supported by~\cite{pan2026adonoisingdispellingmyths}, \emph{even when the target is deterministic}.
Conversely, Chen and Vanden-Eijnden~\cite{chen2026variationaloptimalityfollmerprocesses} study how stochastic dynamics can generate a nondegenerate target distribution from a deterministic initial state.
\ourname uses a singular physics-based reference distribution motivated by inverse problems with highly concentrated posteriors. 
We thus analyze the limiting setting in which \textit{both} endpoints become concentrated and characterize the regularity of the standard and noise-free velocity fields used for reconstruction.

\subsection{Plug-and-Play Priors}
Another approach, complementary to sampling from the conditional posterior directly, is to learn an unconditional prior over the unknown parameters and incorporate the measurement information only during inference.
Plug-and-play (PnP) methods originally introduced this idea by replacing an explicit regularizer with an external denoising prior~\cite{venkatakrishnan2013plug}.
More recently, diffusion and score models have been leveraged as powerful learned priors, while information from the likelihood or forward model is incorporated through inference-time guidance~\cite{song2021scorebased,jalal2021robust,song2022solving,chung2023diffusion,song2023pigdm,wang2023ddnm}.
PnP algorithms with learned priors have also been developed for scientific applications such as electromagnetic scattering~\cite{guo2026scorefield,guo2025pnplatent} and seismic FWI~\cite{wang2023priorfwi,taufik2025dpsfwi}.
While successful in many imaging applications, these methods typically require repeated forward-model evaluations and, depending on the approach, adjoint, pseudoinverse, or Jacobian-based computations throughout inference.
For wave-based problems, this can yield inference costs comparable to classical iterative inversion algorithms.
The InverseBench benchmark~\cite{zheng2025inversebench} provides a systematic comparison of these approaches across a range of scientific inverse problems.
\section{Conclusion}\label{sec:conclusion}

We introduce \ourname, a physics-based framework for scientific inverse problems motivated by two limitations of conventional Gaussian-to-data transport: an uninformative initialization leaves available physics-based approximations unused, and transport to sharply concentrated posteriors can produce ill-conditioned velocity fields. 
\ourname addresses these limitations by reconsidering two design axes: the source distribution and the velocity used for reconstruction.

First, we exploit paired measurement-reconstruction data and the governing physics to construct an informative initialization.
An adjoint or approximate inverse function maps measurements into parameter space, providing a physically meaningful initialization associated with each target reconstruction.
We identify conditions under which this initialization reduces transport distance and show that posterior concentration favors concentrating the source as well.
Second, we introduce a noise-free velocity formulation whose regularity properties in the concentrated-endpoint regime support stable learning and numerical integration.
Together, these choices yield a natural and practically effective data-dependent target-prediction parameterization that incorporates the known initialization analytically and aligns the learning objective with the desired reconstruction.

Across four scientific inverse problems, \ourname matches or even improves reconstruction accuracy on five of seven datasets, while requiring substantially fewer integration steps than diffusion-based baselines. 
By employing additional inference-time scaling, \ournameRED further improves physical consistency and achieves state-of-the-art reconstruction accuracy on every dataset where we apply it.
Our results indicate that the design of generative models for scientific inverse problems should account for the available physics \emph{in the generative transport itself}, replacing generic noise-to-data transport with physics-informed refinement.

\paragraph{Limitations and future work}
Several limitations of \ourname motivate directions for future work.

First, \ourname benefits most when the initialization is informative. Many PDE-constrained inverse problems admit useful adjoints or approximate adjoints, but in settings where the adjoint is unavailable, inaccurate, computationally expensive, or weakly correlated with the target, the advantage of the method may diminish. Heuristic physical priors, truncated classical inversions, or learned approximate inverse maps can be substituted, but their effectiveness is problem-dependent and may require additional validation or inference-time refinement. 

Second, our current implementations assume paired measurement-reconstruction data and, in several experiments, fixed acquisition geometries. Extending the method to variable acquisition geometries such as changing sensor placements or numbers of sensors, as well as to unknown acquisition protocols or strongly distribution-shifted data, is likely possible, but would require additional architecture modifications and broader training distributions beyond the scope of this work. 

Third, our best-performing variant of \ourname falls outside the standard assumptions of existing stochastic-interpolant theory in two places by using a singular initial distribution and training against the noise-free velocity. We therefore do not inherit the marginal-correctness guarantees typical of stochastic interpolants when the conditions of the singular-endpoint limit are not met. 

Finally, although we include robustness experiments, many benchmarks remain synthetic or semi-synthetic. Deployment in high-stakes scientific or engineering pipelines will require further testing under real measurement noise, model mismatch, calibration error, and out-of-distribution physical regimes.

\section*{Impact Statement}
This work aims to improve machine-learning methods for scientific inverse problems. By enabling efficient, high-quality reconstructions, \ourname may lower the computational cost and energy usage of generative inverse solvers, especially in settings where repeated forward and adjoint PDE solves are expensive. More efficient reconstruction tools could make advanced imaging and simulation-based inference more accessible to scientific laboratories, medical researchers, and engineering groups with limited computational resources.

At the same time, inverse-problem solvers can affect consequential decisions when used in domains such as healthcare, geophysics, infrastructure inspection, or remote sensing. Reconstructions produced by \ourname are learned approximations and may be unreliable under certain conditions, an important consideration in settings where false positives/negatives could influence medical diagnosis, safety assessment, resource exploration, or scientific interpretation.


\section*{CRediT Authorship Contribution Statement}

\textbf{Ethan Hanold:} Conceptualization, Methodology, Software, 
Investigation, Formal analysis, Validation, Writing -- original draft, Writing -- review \& editing, Visualization.
\textbf{Nicholas M.\ Boffi:} Conceptualization, Methodology, Supervision, Writing -- original draft,
Writing -- review \& editing.
\textbf{Qin Li:} Conceptualization, Methodology, Supervision, Formal analysis, Writing -- original draft,
Writing -- review \& editing.
\textbf{Leonardo Zepeda-N\'u\~nez:} Conceptualization, Methodology, Supervision, Writing -- original draft,
Writing -- review \& editing.

\section*{Data Availability}

Much of the data used in this study is made available by the respective benchmarks: Back-Projection Diffusion~\cite{Zhang2025_BPD} provides the \textit{10hSquares}~\cite{10hsquares_data} and \textit{Brain Scans}~\cite{brain_data} nonlinear inverse scattering datasets, and InverseBench~\cite{zheng2025inversebench} provides the cell phantom and FWI datasets, as well as validation and test sets for black holes~\cite{inversebench_data}. 
The \textit{3--5--10h Triangles} and \textit{Shepp--Logan} datasets were provided
directly by the authors of Back-Projection Diffusion~\cite{Zhang2025_BPD}.
Code for generating these datasets is publicly available at
\url{https://github.com/borongzhang/back_projection_diffusion}.
The pretrained diffusion prior model checkpoint and generation script for generating a black hole training dataset are publicly available at \url{https://github.com/devzhk/InverseBench}.
Code for \ourname will be made available upon reasonable request. 

\section*{Declaration of Generative AI and AI-Assisted Technologies in the Manuscript Preparation Process}

During the preparation of this work the authors used ChatGPT, Gemini and Claude to write code and improve language and formatting. After using these tools, the authors reviewed and edited the content as needed and take full responsibility for the content of the published article.

\bibliography{ref}


\appendix
\renewcommand{\theHequation}{appendix.\thesection.\arabic{equation}}
\section{Proofs}\label{app:proofs}

\subsection{Proof of Proposition~\ref{prop:reduced_transport_case_linear}}
\begin{proof}
Let $r=\|F^*F-I\|_{\mathrm{op}}<1$. Then $F^*F\succ0$, and completing the square up to an $m$-independent factor gives
\begin{equation}\label{eq:linear_posterior}
\begin{aligned}
    p(m\mid y) \,\propto\, & p(y\mid m)p(m) \\
    \,\propto\, & \operatorname{exp}\left(-\frac{1}{2\sigma_y^2}\|y-Fm\|_2^2\right) \\
    \,=\, & \exp\left(-\frac{1}{2\sigma_y^2}\left(y^*y - 2m^*F^*y + m^*F^*Fm\right)\right) \\
    \,\propto\, & \exp\left(-\frac{1}{2\sigma_y^2}\left(m - (F^*F)^{-1}F^*y\right)^*(F^*F)\left(m - (F^*F)^{-1}F^*y\right)\right) \\
    \,=\, & \exp\left(-\frac1{2\sigma_y^2}(m-\mu)^*F^*F(m-\mu)\right).
\end{aligned}
\end{equation}
We now apply the decomposition~\eqref{eq:wasserstein_source_translation}, using the fact that $\overline\rho_0^\mathrm{adj}=\overline\rho_0=\rho_0$ since they only differ by a shift:
\begin{equation}\label{eq:wasserstein_transport_analysis_linear}
\begin{aligned}
    \mathcal W_2^2(\rho_0,\rho_1)
    - \mathcal W_2^2(\rho_0^{\mathrm{adj}},\rho_1)
    &= \|\mu\|_2^2 + \mathcal W_2^2(\rho_0,\overline\rho_1)
    - \left(\|F^*y-\mu\|_2^2
    + \mathcal W_2^2(\rho_0,\overline\rho_1)\right) \\
    &= \|\mu\|_2^2 - \|F^*y-\mu\|_2^2 \\
    &= \|\mu\|_2^2 - \|(F^*F-I)\mu\|_2^2 \\
    &\geq (1-r^2)\|\mu\|_2^2.
\end{aligned}
\end{equation}
This is nonnegative, and positive when $F^*y\ne0$. If $F^*y=0$, both base distributions coincide.
\end{proof}

\subsection{Proof of Proposition~\ref{prop:reduced_transport_case_nonlinear}}
\begin{proof}
Given the decomposition~\eqref{eq:wasserstein_source_translation}, we aim to show $\|F^*\widetilde y - \mu\|_2 < \|\mu\|_2$.
By the triangle inequality, $\|F^*\widetilde y - \mu\|_2 \le \|F^*\widetilde y - \delta^\star\|_2 + \tau$, and by the reverse triangle inequality, $\|\mu\|_2 \ge \|\delta^\star\|_2 - \tau$. 
Comparing them gives a difference
\begin{equation}\label{eq:transport_distances_nonlinear}
    \|\mu\|_2 - \|F^*\widetilde y - \mu\|_2 
    \geq \|\delta^\star\|_2 - \|F^*\widetilde y - \delta^\star\|_2 - 2\tau,
\end{equation}
so we look to show that the right-hand side is positive.
By Taylor's theorem, with an $L$-Lipschitz Jacobian, 
\begin{equation}
    \mathcal F(m^\star) = \mathcal F(\overline m) + F\delta^\star + R
\end{equation}
with $\|R\|_2 \le \tfrac L2\|\delta^\star\|_2^2$. Therefore,
\begin{equation}
    \widetilde y 
    = y - \mathcal F(m^\star) + \mathcal F(m^\star) - \mathcal F(\overline m) 
    = e + F\delta^\star + R
\end{equation}
and
\begin{equation}\label{eq:residual_bound}
    F^*\widetilde y - \delta^\star = (F^*F - I)\delta^\star + F^*R + \left(F - \tfrac{\partial\mathcal F}{\partial m}(m^\star)\right)^* e,
\end{equation}
where we use $\big(\tfrac{\partial\mathcal F}{\partial m}(m^\star)\big)^*e=0$ because $m^\star$ minimizes the least-squares misfit.
Then, taking the norm of the above equation gives
\begin{equation}
    \|F^*\widetilde y - \delta^\star\|_2 \le r\|\delta^\star\|_2 + \tfrac L2\|F\|_{\mathrm{op}}\|\delta^\star\|_2^2 + L\|\delta^\star\|_2\|e\|_2,
\end{equation}
where $\|F - \tfrac{\partial\mathcal F}{\partial m}(m^\star)\|_{\mathrm{op}} \le L\|\delta^\star\|_2$ because the Jacobians are $L$-Lipschitz.
Finally, applying assumption~\eqref{eq:nonlinear_assumption} and plugging back into~\eqref{eq:transport_distances_nonlinear} gives the desired result.
\end{proof}

\subsection{Proof of Proposition~\ref{prop:analytic_b}}\label{sec:analytic_b}
\begin{proof}
Since $\dot{I}_t$ and $I_t$ are jointly Gaussian, the conditional expectation formula for jointly Gaussian random variables gives
\begin{equation}\label{eq:joint_conditional_expectation}
    b_t(x)\,=\,
    \mathbb{E}\left[\dot{I}_t\right] + \operatorname{Cov}(\dot{I}_t, I_t)\operatorname{Cov}(I_t)^{-1}\left(x - \mathbb{E}\left[I_t\right]\right).
\end{equation}
Define
\begin{equation}\label{eq:analytic_b_components}
    \mu_t := \alpha_t\mu_0 + \beta_t\mu_1 = \mathbb{E}\left[I_t\right], \qquad
    K_b(t) := \tfrac{1}{2}\dot{C}_tC^{-1}_t, \qquad
    C_t := \alpha^2_t \sigma_0^2I + \beta^2_t \sigma_1^2\Sigma_1 + \gamma^2_t \sigma^2I.
\end{equation}
If $x_0, x_1$, and $z$ are independent, then $\operatorname{Cov}(\dot I_t,I_t)=\tfrac12\dot C_t$ and
\begin{equation}\label{eq:b_final}
b_t(x)=\dot\mu_t+K_b(t)(x-\mu_t).
\end{equation}
If the target endpoint is singular, i.e. $\sigma_1=0$, we may write $C_t=c_tI$ with $c_t=\alpha_t^2\sigma_0^2+\gamma_t^2\sigma^2$, and then
\begin{equation}
    K_b(t) = \left(\tfrac{1}{2}\tfrac{d}{dt}\log c_t\right)I.
\end{equation}
Then, interpolant boundary conditions dictate that $c_t\to0$ as $t\uparrow1$, and thus $K_b(t)$ must become unbounded as $t\uparrow1$.

Similarly, if $\sigma_0=0$, we may apply Jacobi’s formula for differentiating a determinant to get
\begin{equation}
    \operatorname{tr} K_b(t) = \tfrac12\tfrac{d}{dt}\log\det C_t.
\end{equation}
The interpolant boundary conditions imply $\det C_t\to0$ as $t\downarrow0$, so $|\log\det C_t|\to\infty$ and $K_b$ once again becomes unbounded as we approach the singular endpoint.
\end{proof}

For the linear schedule with $\sigma_0>0$, $\sigma_1=\sigma=0$,
$K_b=-I/(1-t)$ and $b_t(I_t)=\mu_1-x_0$ along exact interpolant trajectories.

For the variance-preserving schedule $\alpha_t=\sqrt{1-t^2}$ and $\beta_t=t$, one instead obtains
$K_b=-tI/(1-t^2)$ and $b_t(I_t)=\mu_1-tx_0/\sqrt{1-t^2}$, which diverges in norm for $x_0\ne0$.

\subsection{Proof of Corollary~\ref{cor:analytic_v}}
\begin{proof}
The same Gaussian-conditioning formula, now for $\dot\alpha_tx_0+\dot\beta_tx_1$, gives
\begin{equation}\label{eq:kv}
    v_t(x)=\dot\mu_t+K_v(t)(x-\mu_t),\qquad
    K_v(t) := \bigg(\dot{\alpha}_t \alpha_t \sigma_0^2I + \dot{\beta}_t \beta_t \sigma_1^2\Sigma_1\bigg)\, C^{-1}_t.
\end{equation}
Crucially, $K_v$ vanishes if $\sigma_0=\sigma_1=0$.

More generally, suppose $\sigma_0=0,\sigma>0$, and the target distribution has a positive-definite covariance $\Sigma_1\succ0$ (here we absorb any scaling $\sigma_1$ into $\Sigma_1$).
Then, under the schedules $\alpha_t=1-t$, $\beta_t=t$, and $\gamma_t=\sqrt{t(1-t)}$ we have 
\begin{equation}
    C_t=t\left(t\Sigma_1+(1-t)\sigma^2I\right) \implies K_v=\Sigma_1\left(t\Sigma_1+(1-t)\sigma^2I\right)^{-1}.
\end{equation}

$K_v$ has eigenvalues $\lambda/[t\lambda+(1-t)\sigma^2]>0$ for $\lambda>0$ since $\Sigma_1$ is positive definite, and thus extends continuously to $[0,1]$.
As a result,
\begin{equation}
    \|v_t(x)-v_t(y)\|
    =\|K_v(t)(x-y)\|
    \leq \|K_v(t)\|_{\mathrm{op}}\|x-y\|,
\end{equation}
so $v_t$ is uniformly Lipschitz in $x$, giving uniqueness
for the noise-free ODE.
Finally, we set $e_t=x_t-\mu_t$, with $\dot e_t=v_t(x_t) - \dot \mu_t = K_v(t)e_t$. Then initial condition $e_0=0$ and uniqueness give $e_t=0$, hence $x_1=\mu_1$.
\end{proof}

\section{Baselines}\label{app:baselines}
\subsection{Plug-and-Play Methods}\label{app:pnp_methods}
Since the posterior can be expressed as $p(x\mid y)\propto p(y\mid x)p(x)$, plug-and-play diffusion prior (PnP-DP) methods leverage a pretrained diffusion model which can sample from $p(x)$, and then incorporate data-misfit information from the forward model to approximate the likelihood $p(y\mid x)$.
PnP-DP methods have recently emerged as a powerful paradigm for image-based inverse problems due to the existence of large-scale diffusion prior models for natural images~\cite{dhariwal2021diffusion}, and have recently begun to see use in scientific applications as well~\cite{wang2023priorfwi,guo2026scorefield,guo2025pnplatent,taufik2025dpsfwi}.
These methods fall into a number of distinct categories, as delineated by InverseBench~\cite{zheng2025inversebench}, which we summarize now for completeness:

\paragraph{Likelihood-based guidance} Guidance methods 
\cite{song2023pigdm, wang2023ddnm, kawar2022denoising, rout2023solving, chung2023diffusion, song2023lgd} 
apply Bayes rule to rewrite the posterior score function as
\begin{equation}
    \nabla_{x_t}\log p(x_t\mid y) = \nabla_{x_t}\log p(y\mid x_t) + \nabla_{x_t}\log p(x_t),
\end{equation}
and approximate the likelihood score while using a pretrained diffusion model to approximate the prior score.

\paragraph{Variable splitting} Borrowing from variable-splitting techniques in optimization \cite{vono2019split}, these methods divide inference into two alternating steps 
\cite{zhu2023denoising, li2024decoupled, song2024solving, zhang2024daps, xu2024provably}.
One step uses the forward model to sample the posterior in the neighborhood of the current $x_t$ (for example, using Langevin Monte Carlo) to reduce data misfit, and the second step uses the diffusion model to apply a prior-based denoising update.

\paragraph{Variational inference} These methods aim to approximate the posterior distribution $p(x\mid y)$ with a simpler, easy-to-sample \textit{variational} distribution $q(x\mid y)$ via the KL-minimization objective
\begin{equation}
    \underset{q}{\min}\,D_{KL}\big(q(x\mid y) \,\| \,p(x\mid y)\big),
\end{equation}
thus transforming the standard sampling procedure into an optimization problem \cite{mardani2024reddiff, feng2023score, feng2024variational}.

\paragraph{Sequential Monte Carlo} 
SMC~\cite{delmoral2006sequential} augments standard Markov Chain Monte Carlo algorithms -- which can struggle in inverse problems due to abrupt transitions from prior to posterior -- and has recently been adapted to the sampling process of diffusion models \cite{wu2023practical, trippe2023diffusion,cardoso2024monte,dou2024diffusion}.
Such algorithms ease the transition from prior to posterior by introducing a smooth sequence of intermediate distributions that converge to the target distribution.
Intermediate distributions are represented by a population of particles with importance weights that are evolved through iterative reweighting and resampling.

\subsection{Deterministic Baselines}
Two types of deterministic baselines are considered: classical numerical inversion and optimization algorithms and deterministic point-reconstruction ML models.

For nonlinear inverse scattering and seismic imaging, full waveform inversion (FWI)~\cite{Tarantola:Inversion_of_seismic_reflection_data_in_the_acoustic_approximation} is a physics-based baseline that reconstructs the perturbation by minimizing the nonlinear data misfit while updating the background model. 
Optimization is conducted with momentum techniques such as Adam or quasi-Newton methods such as L-BFGS \cite{liu1989limited}. 
In addition, for seismic imaging we consider the best-case scenario for FWI, where we set the initial condition of the model to be a smoothed version of the ground-truth.
This is unrealistic as computing this initial condition is still an area of active research~\cite{li2016full}, and is usually performed by other modalities, which are lower-resolution but are less prone to local minima.

For nonlinear inverse scattering we also consider four state-of-the-art deterministic neural operators. 
Building on the filtered back-projection (FBP)~\cite{colton1998inverse} framework, these networks aim to approximate both the back-scattering and filtering operators with structured neural networks.
All four methods represent the filtering operator using a translationally equivariant CNN. 
They differ primarily in how the back-scattering operator is parameterized. 
In particular, SwitchNet~\cite{KhooYingSwitchNet:2019} and WideBNet~\cite{li2022butterfly} employ butterfly-based factorizations to obtain low-complexity representations of the scattering operator, with WideBNet further incorporating FFT-inspired structure. 
EquiNet~\cite{zhang2024_equivariant} reformulates the scattering integral via a change of variables to construct a rotationally equivariant network. 
Building on this, B-EquiNet~\cite{zhang2024_equivariant} compresses the EquiNet architecture via a butterfly factorization. 
Deterministic models are trained end-to-end by minimizing mean squared error between predicted and ground-truth perturbations.

For linear inverse scattering, we consider FISTA-TV~\cite{beck2009fasttv}, which reconstructs the scattering potential by minimizing a least-squares data misfit with total variation (TV) regularization.
The data misfit uses the linear forward operator obtained under the first Born approximation, while the TV penalty promotes piecewise-constant reconstructions with sharp interfaces.
Optimization combines gradient steps on the data misfit with proximal updates for the TV regularizer and a momentum-based extrapolation that accelerates convergence.

For black hole imaging, we consider the classical reconstruction algorithms SMILI~\cite{Akiyama_2017} and EHT-Imaging~\cite{Chael_2016, Chael_2018}.
Both implement regularized imaging methods that balance agreement with interferometric measurements against explicitly prescribed image priors.
For the closure-based formulation considered here, reconstruction involves fitting closure phases and log closure amplitudes, together with the total flux constraint.
These methods optimize the image directly, using regularization to constrain the reconstruction under sparse Fourier coverage.

\section{Inference-Time Scaling}\label{app:inference_scaling}

\begin{algorithm}[tb]
\caption{\ournameRED: learned-field     data-consistency refinement}\label{alg:red_flow}
\begin{algorithmic}
\STATE \textbf{Input:} $y$, initialization $a$, target-prediction network $\hat m_t$, forward map $\calF$, steps $n$, time-sampling distribution $p_t$, weight $\lambda_t$, and optimizer.
\STATE Set $x_0=a(y)$ and $m^{(0)}=x_0$; initialize optimizer state.
\FOR{$k=0,\ldots,n-1$}
\STATE Draw $t\sim p_t$; set $x_t^{(k)}=\alpha_tx_0+\beta_tm^{(k)}$.
\STATE Calculate gradient $g_k=\nabla_m\|y-\calF(m)\|^2|_{m=m^{(k)}}+
\lambda_t\operatorname{sg}[m^{(k)}-\hat m_t(x_t^{(k)};y)]$.
\STATE Take one optimizer step using $g_k$, producing $m^{(k+1)}$.
\ENDFOR
\STATE \textbf{Return:} $m^{(n)}$.
\end{algorithmic}
\end{algorithm}

RED-Diff~\cite{mardani2024reddiff} is a variational approach to inference that bears resemblance to the regularization-by-denoising (RED) optimization method proposed in~\cite{romano2017red}. We adapt this method to the stochastic interpolant framework and call the resulting method \ournameRED.

Following the framework of RED-Diff, we aim to approximate the intractable posterior distribution $p(x\mid y)$ with a simpler Gaussian \textit{variational distribution} $q(x\mid y) = \mathcal{N}(x; m, \sigma^2I)$ that seeks to minimize the KL divergence
\begin{equation}\label{eqn:variational_obj}
    \underset{m,\sigma}{\text{min}}\,\, KL\left(
    q(x\mid y)\|p(x\mid y)
    \right),
\end{equation}
essentially aiming to match the dominant mode of the posterior. Then, by allowing $\sigma\to0$, the objective can be reduced to solving for the MAP estimator $m^\mathrm{MAP}$~\cite{mardani2024reddiff} which satisfies:
\begin{equation}\label{eq:reddiff_final_obj}
    \underset{m}{\text{min}}\,\,
    \underbrace{\|y-\calF(m)\|^2}_{\text{data misfit}} +
    \underbrace{\mathbb{E}
    \left[\lambda_t(\textbf{sg}[m-\hat{m}_t])^\top m\right]}_{\text{regularization}},
\end{equation}
where \textbf{sg} is the stop-gradient operator, $\lambda_t$ is a time-dependent weight scaling to balance the two terms, and $\hat{m}_t$ is the minimum MSE target-prediction estimator produced by a diffusion model (with the expectation taken over time $t$ and the noise initialization $x_0\sim\calN(0,I)$ of the diffusion model).

We now consider the coupled stochastic interpolant with a learned noise-free velocity $\hat{v}_t(x; y)=\dot{\alpha}_ta(y)+ \dot{\beta}_t\hat{m}_t$.
With the residual of the corresponding flow-matching loss, we have
\begin{equation}
    \mathbb{E}
    \left[\lambda_t
    \big((\dot{\alpha_t}x_0 + \dot{\beta_t}m)
    - \hat{v}_t(x_t; y)\big)
    \right]\\
    =\mathbb{E}
    \left[\lambda_t\dot{\beta}_t(m-\hat{m}_t)
    \right],
\end{equation}
where the $x_0$ terms cancel due to the data-dependent target-prediction parameterization.
We therefore recover an analogue of the original regularization term in~\eqref{eq:reddiff_final_obj} (up to a factor of $\dot{\beta}_t$), leading us to propose the \ournameRED objective
\begin{equation}
    \,\underset{m}{\min}\,\,
    \underbrace{ \|y-\calF(m)\|^2}_{\text{data misfit}} +
    \underbrace{\mathbb{E}
    \left[\frac{\lambda_t}{\dot{\beta}_t}
    \textbf{sg}[\dot{x}_t - \hat{v}_t(x_t; y)]^\top m
    \right]}_{\text{regularization}},
\end{equation}
with $x_t=\alpha_ta(y) + \beta_t m$. 
We then initialize $m^{(0)}=a(y)$ with the physics-based initializer and run an off-the-shelf optimizer such as Adam~\cite{kingma2015adam} to solve the stochastic optimization problem.
The full algorithm is covered in Algorithm~\ref{alg:red_flow}.

We mention that unlike in RED-Diff, here $x_t$ carries no injected noise.
The noise in RED-Diff is a sample from its Gaussian source, whereas our source $a(y)$ is deterministic (the bridge noise $\gamma_t z$ serves only as training-time augmentation and we do not include it for inference).
As a result, the only source of randomness for \ournameRED is the interpolant time $t$.

In practice, we find it sufficient to parameterize the time-dependent weighting as $\lambda_t=\lambda^a_0t^a + \lambda_0^b(1-t)^b$ for hyperparameters $\lambda^a_0, \lambda_0^b, a, b$. 
We also found it beneficial to use a U-shaped time-sampling distribution~\cite{lee2024improving} which biases the endpoints of $[0,1]$:
\begin{equation}\label{eq:u_shaped_time}
    p_{\mathrm{U}}(t;\xi)
    \propto \exp\!\big(\xi(2t-1)\big) + \exp\!\big(-\xi(2t-1)\big)
    = 2\cosh\!\big(\xi(2t-1)\big),
    \qquad t\in[0,1],
\end{equation}
where $\xi$ is a hyperparameter that controls how strongly samples concentrate near $t\approx 0$ and $t\approx 1$.
For these, as well as the optimizer's base learning rate $\eta$, we perform a hyperparameter sweep over a validation set with Bayesian optimization to select the final hyperparameters used for this paper's results.

We implement \ournameRED for both linear and nonlinear inverse scattering, as well as black hole imaging.
We find that it consistently leads to improved data consistency -- and often reconstruction quality as well -- at the cost of additional inference-time compute (for example, as shown in~Figure~\ref{fig:brain_barchart}).
Most notable is the impact on black hole imaging in~Table~\ref{table:blackhole_results}: \ourname struggles to achieve competitive $\chi^2$ metrics but \ournameRED fully resolves this issue, producing some of the best $\chi^2$ values of any method while using the same base model.

Therefore, if higher accuracy is desired and real-time reconstruction is not required, \ournameRED can improve reconstruction quality of a pretrained \ourname model with no additional retraining.

\section{Metrics}\label{app:metrics}
We evaluate reconstruction accuracy using the following metrics, which encompass standard image-based analyses, distributional distances, and problem-specific metrics.
For stochastic methods, we generate one reconstruction per measurement $y$, average metrics across the entire evaluation dataset, and report the means (and for some cases standard deviations).
All reported inference costs correspond to this single-realization setting.
We find that averaging multiple reconstructions per measurement does not substantially affect accuracy results, yet greatly increases inference cost, so we do not focus on this case.

\paragraph{Relative root mean square error (RRMSE)}
\begin{equation}
    \text{RRMSE}(\hat{m}, m)=\frac{\|\hat{m}-m\|_2}{\|m\|_2}
\end{equation}
is a common metric used to quantify pixel-wise reconstruction fidelity in PDE problems.

\paragraph{Peak signal-to-noise ratio (PSNR)}
\begin{equation}
    \text{PSNR}(\hat{m}, m)=10\log_{10}\left(\frac{1}{\text{MSE}(\hat{m}, m)}\right),
\end{equation}
where $\text{MSE}(\hat{m}, m)=\frac{1}{n^2}\|\hat{m}-m\|_F^2$ is the Mean-Squared Error for $n\times n$ images. PSNR gives the ratio between the maximum possible signal strength of the image and the power of the noise corrupting the reconstruction, and is represented as a logarithmic quantity using the decibel scale.
For black-hole imaging, Blur PSNR evaluates reconstruction accuracy after blurring to the telescope’s target resolution~\cite{zheng2025inversebench}.

\paragraph{Structural similarity index measure (SSIM)}
\begin{equation}\label{eq:ssim}
    \text{SSIM}(\hat{m}, m) = l(\hat{m}, m)^\alpha\cdot c(\hat{m}, m)^\beta \cdot s(\hat{m}, m)^\gamma
\end{equation}
is a metric used to quantify perceptual similarity between two images by taking into account the comparative measures of luminance ($l$), contrast ($c$), and structure ($s$)~\cite{wang2004image}.

\paragraph{Sinkhorn divergence (SD)}
Sinkhorn Divergence, an entropically regularized optimal-transport-based distance between distributions~\cite{cuturi2013sinkhorn, peyre2019computational}, is a metric to measure distributional alignment. We compute SD using the JAX-based Optimal Transport Tools~\cite{cuturi2022optimal} version 0.5.1, updated from the version used in~\cite{Zhang2025_BPD}.

\paragraph{Mean energy log ratio (MELR)}
To quantify how well reconstructions capture spectral consistency across spatial scales, we use MELR, which compares the energy spectra of generated and ground-truth samples in the Fourier domain.

We define the energy spectrum, which conveys how much energy an image $m\in\R^{n\times n}$ has in a given radial frequency, as
\begin{equation}
    E_k(m) = 
    \sum_{|\underbar{k}|=k}\left|
    \sum_{x,y}
    m(x, y)e^{\frac{-2\pi i}{n}\underbar{k}\cdot (x,y)}
    \right|^2,
\end{equation}
where $k$ is the magnitude of the 2D wave-vector $\underbar{k}$ and $x,y=0,1,\dots,n-1$ represent the underlying spatial grid.

Then to assess the overall consistency of the spectrum of the predicted image, we compute the scalar quantity
\begin{equation}
    \text{MELR}(\hat{m}, m)=\frac{1}{N_k}\sum_{k}\left|
    \log\left(\frac{E_k(\hat{m})}{E_k(m)}\right)
    \right|.
\end{equation}

\paragraph{$\chi^2$ statistics}
For the problem of black hole imaging, a standard data-fidelity objective uses a (normalized) $\chi^2$ misfit\footnote{
We refer to this weighted squared-residual objective as $\chi^2$, following standard convention in VLBI imaging. 
Under an idealized model where the (whitened) measurement residuals are independent and drawn from $\mathcal{N}(0,1)$, the sum of their squares follows a $\chi^2$ distribution; consequently, a reduced $\chi^2$ (normalized by the number of data points) of $\approx 1$ indicates a fit at the nominal noise level. 
For closure quantities -- which are nonlinear functions of visibilities -- this probabilistic interpretation is approximate, and thus we employ $\chi^2$ primarily as a standardized measure of data misfit to align with~\cite{zheng2025inversebench}.
}
on the closure quantities and flux:
\begin{equation}\label{eq:chi2}
\begin{split}
    \mathcal{L}(y\mid m) &= \chi^2_{\mathrm{cp}}(m) + \chi^2_{\mathrm{logca}}(m) + \chi^2_{\mathrm{flux}}(m) \\
    &= 
    \frac{1}{N_{\mathrm{cp}}}\left\|
       \frac{\calF_{\mathrm{cp}}(m)-y_{\mathrm{cp}}}{\sigma_{\mathrm{cp}}}
       \right\|_2^2 
       + \frac{1}{N_{\mathrm{logca}}}\left\|
       \frac{\calF_{\mathrm{logca}}(m)-y_{\mathrm{logca}}}{\sigma_{\mathrm{logca}}}
       \right\|_2^2
     + \left\|
       \frac{\calF_{\mathrm{flux}}(m)-y_{\mathrm{flux}}}{\sigma_{\mathrm{flux}}}
       \right\|_2^2 ,
\end{split}
\end{equation}
with $\sigma_{\mathrm{cp}},\sigma_{\mathrm{logca}},\sigma_{\mathrm{flux}}$ representing estimated standard deviations and $N_{\mathrm{cp}},N_{\mathrm{logca}}$ the counts of closure measurements.

As a $\chi^2$ value closer to 1 indicates a better fit, we follow~\cite{zheng2025inversebench} and define a unified $\chi^2$ metric
\begin{equation}
    \tilde{\chi}^2 = \chi^2\cdot\mathbf{1}\{\chi^2\geq1\} + \frac{1}{\chi^2}\cdot\mathbf{1}\{\chi^2<1\}
\end{equation}
to encompass both underfitting ($\chi^2>1$) and overfitting ($\chi^2<1$) cases.

\section{Additional Ablations and Experiments} \label{app:ablations}
Here, we extend the exploration of model design choices discussed in~Section~\ref{sec:ablations} of the main paper.

\subsection{Robustness}
In our experiments, the forward operator is exactly known and perfectly generates the synthetic data we train on.
Yet, real-world measurement data often contains uncertainty due to factors such as imperfect sensing devices or even epistemic uncertainty in the system we are modeling. 
We should therefore \textit{not} expect any guarantee that real-world data perfectly matches the training distribution on which \ourname was trained. 
In fact, both benchmarks we compare against~\cite{Zhang2025_BPD,zheng2025inversebench} explicitly state the importance of robustness to distributional shifts due to noisy or mismatched forward models.
Therefore, to assess the robustness of \ourname to real-world measurement uncertainties, we evaluate performance in two scenarios: noisy measurements and inverse crimes.

\subsubsection{Noisy Measurement Data}
Following the evaluation criteria of InverseBench~\cite{zheng2025inversebench}, the linear inverse scattering dataset already contains Gaussian noise and the black-hole dataset already contains Gaussian thermal noise in addition to phase and amplitude errors.
In both cases, \ourname still performs well compared to other methods (see~Tables~\ref{table:results_linear_inv_scattering_full} and~\ref{table:blackhole_results}).

\begin{figure*}[!htbp]
    \centering

    \begin{subfigure}[b]{0.45\textwidth}
        \centering
        \includegraphics[width=\linewidth]{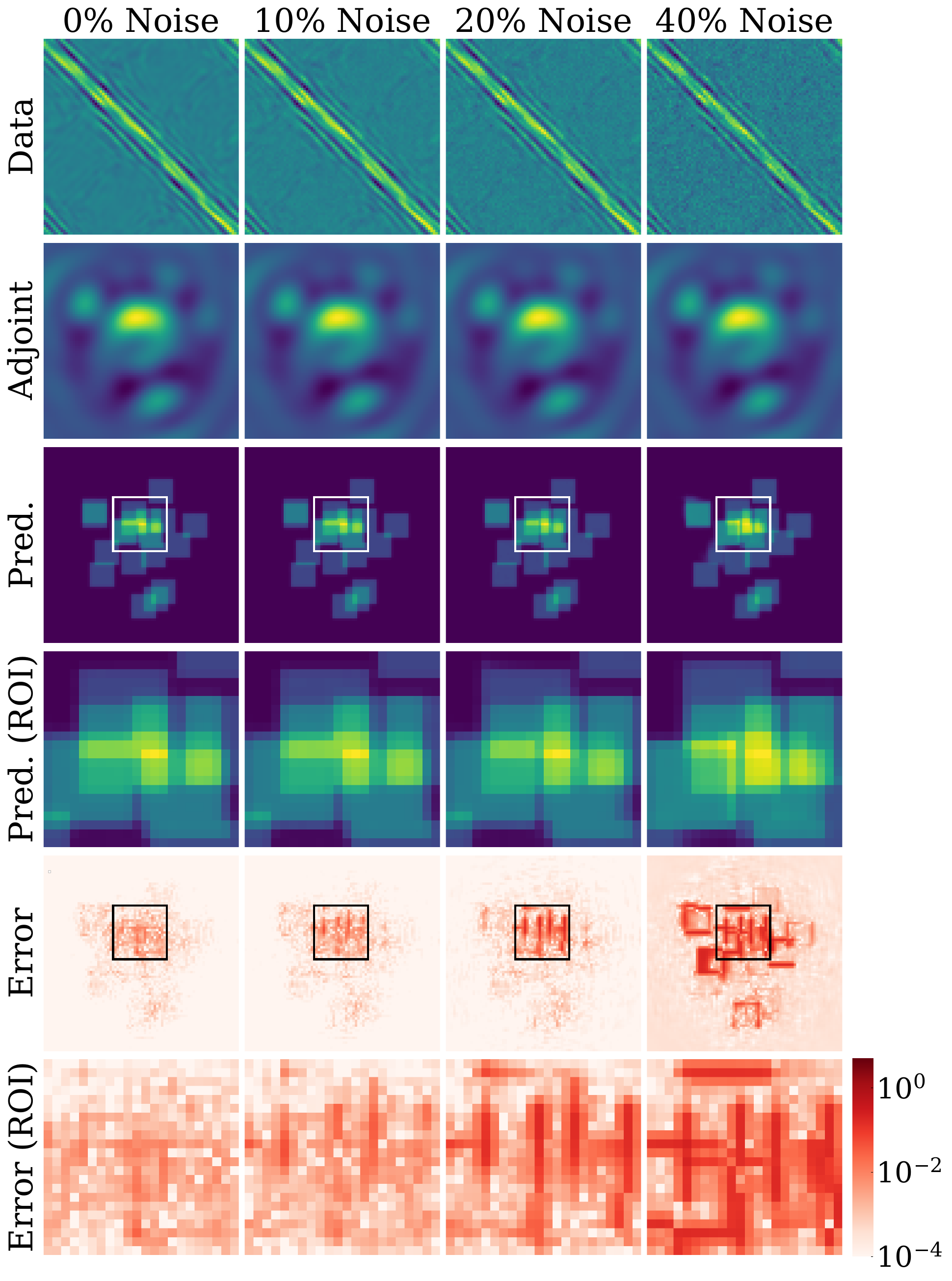}
    \end{subfigure}
    \hfill
    \begin{subfigure}[b]{0.45\textwidth}
        \centering
        \includegraphics[width=\linewidth]{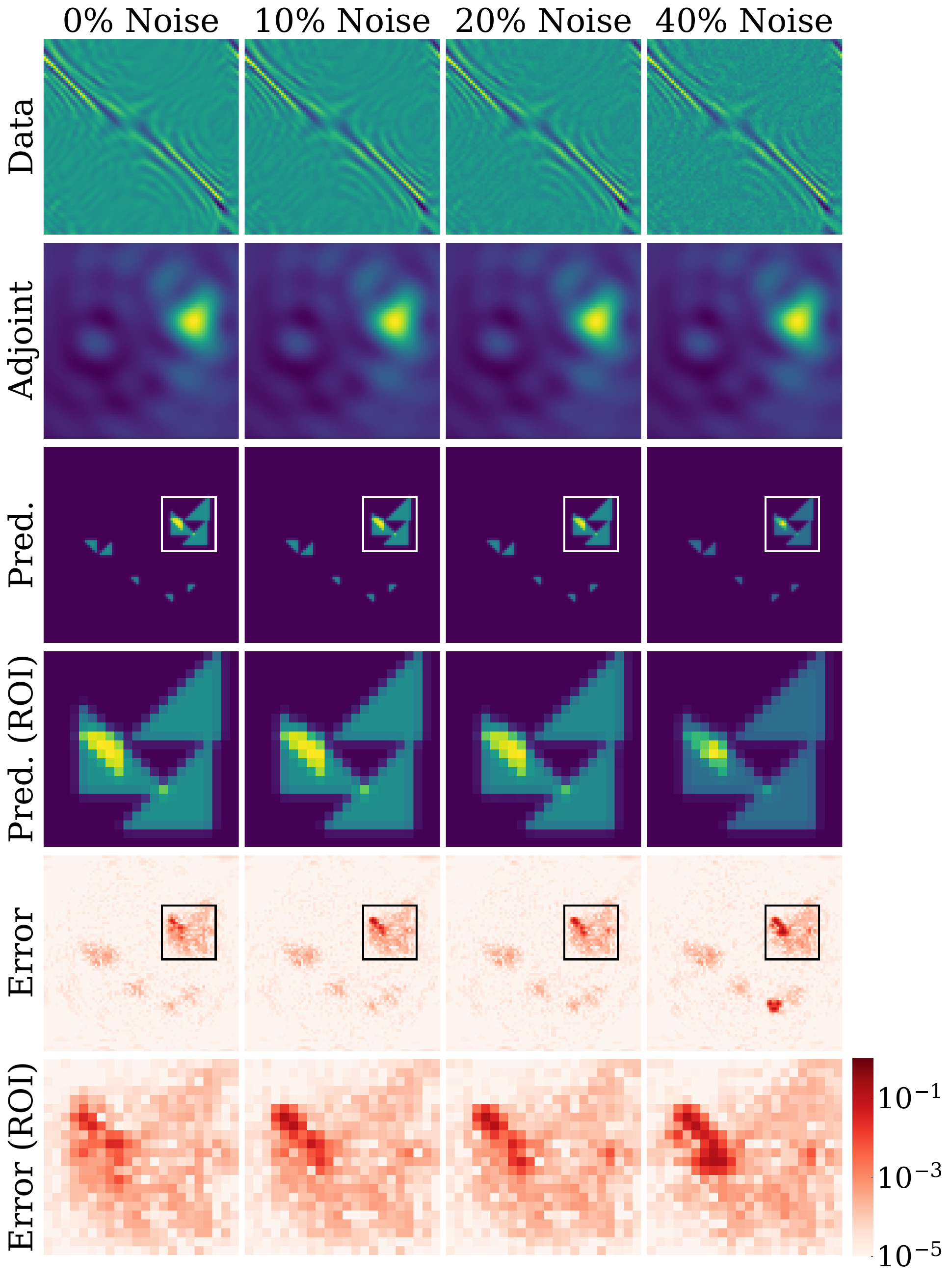}
    \end{subfigure}

    \vspace{0.5em}

    \begin{subfigure}[b]{0.45\textwidth}
        \centering
        \includegraphics[width=\linewidth]{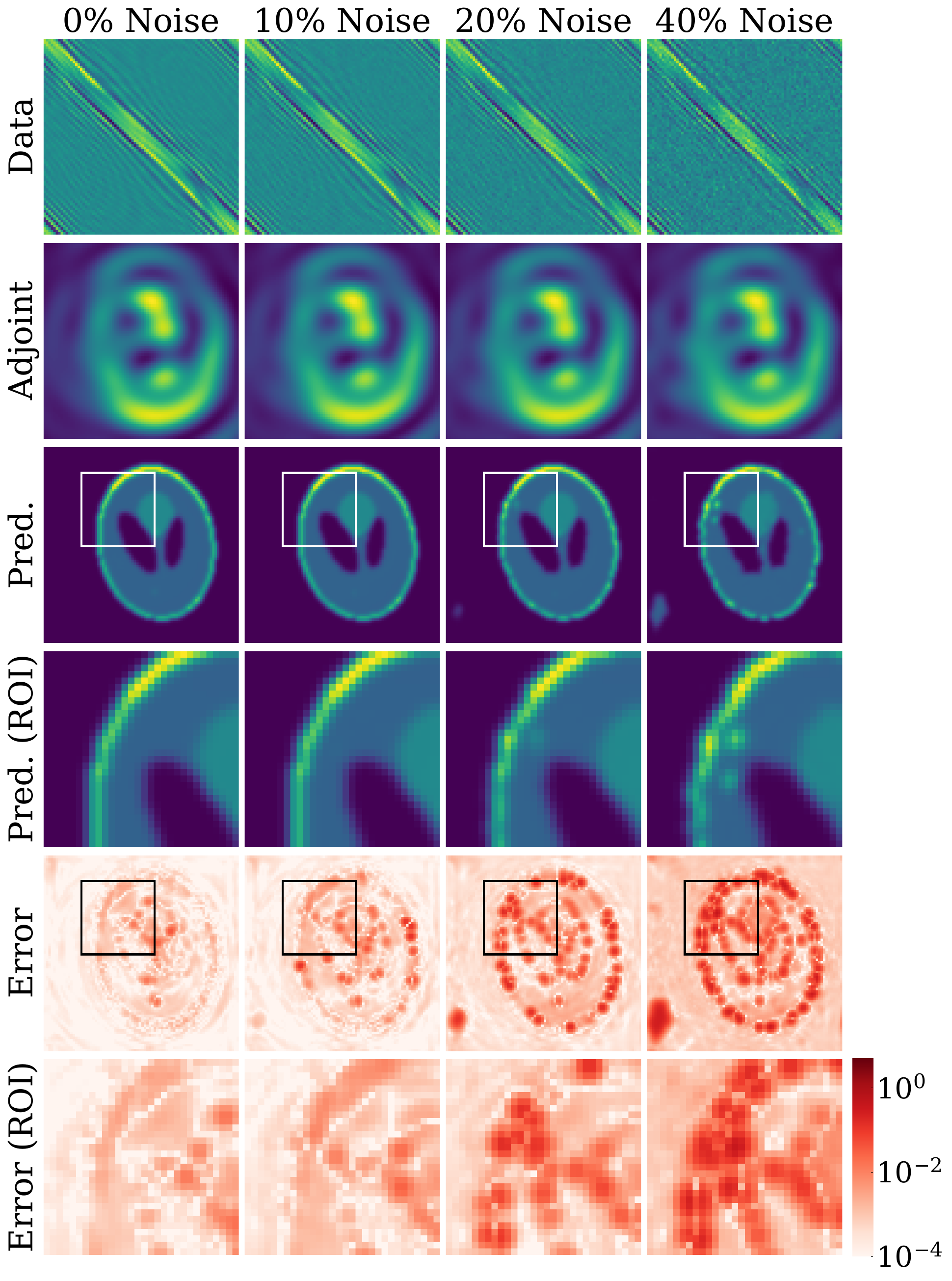}
    \end{subfigure}
    \hfill
    \begin{subfigure}[b]{0.45\textwidth}
        \centering
        \includegraphics[width=\linewidth]{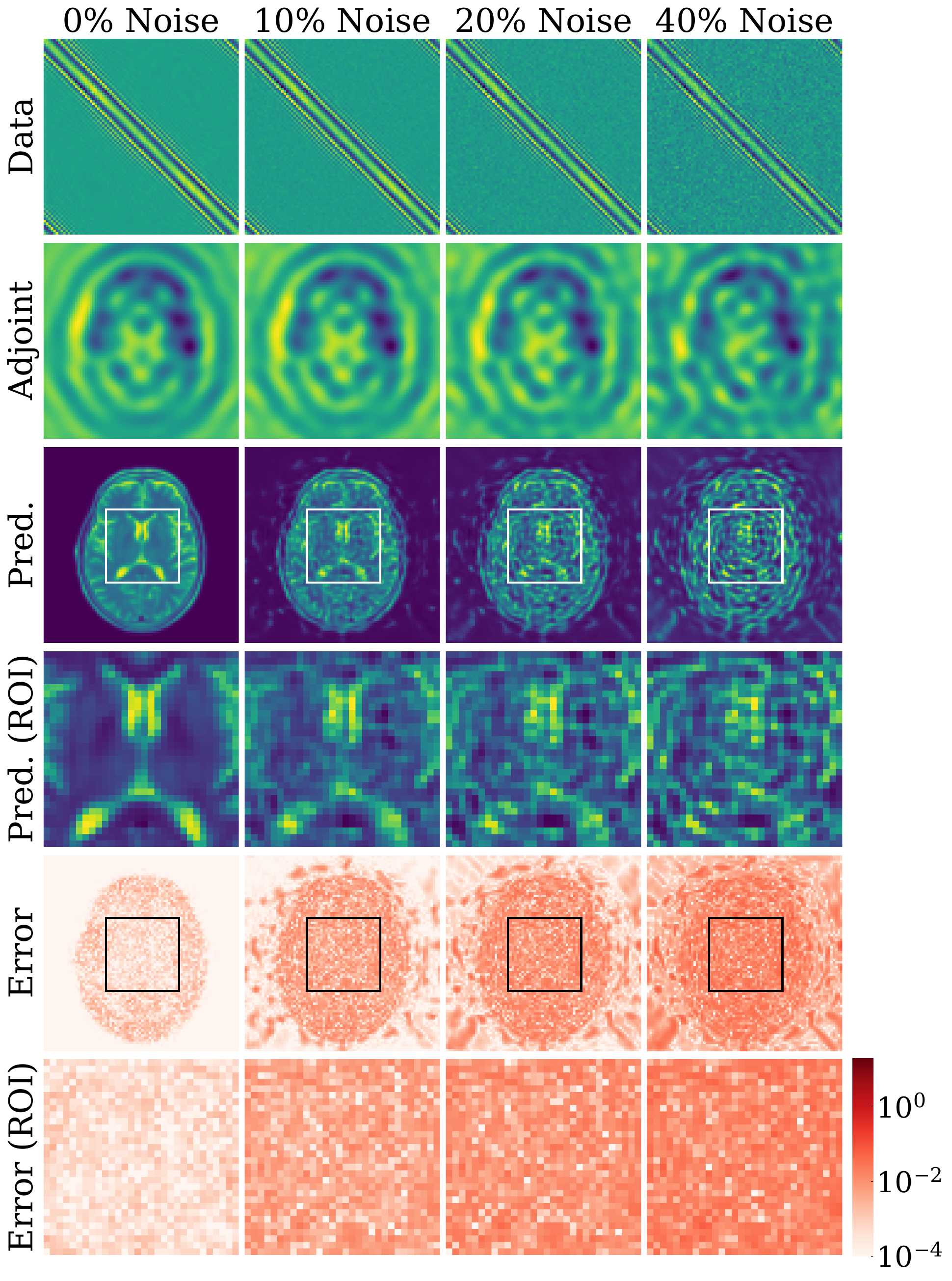}
    \end{subfigure}

    \caption{Effect of measurement noise on nonlinear inverse
    scattering. The panels show \textit{10h Overlapping Squares}
    (top left), \textit{3--5--10h Triangles} (top right),
    \textit{Shepp--Logan} (bottom left), and \textit{Brain Scans}
    (bottom right). Within each panel, columns correspond to
    0\%, 10\%, 20\%, and 40\% added noise, as defined in the text.
    Rows show the measurement data, adjoint-based initialization,
    \ourname reconstruction, reconstruction within the boxed
    region of interest (ROIs), reconstruction error, and error within
    that region.}
    \label{fig:noisy_measurements}
\end{figure*}

To further investigate the effect of noisy measurements on \ourname, we perform a systematic sweep over varying levels of measurement noise for each of the four nonlinear inverse scattering datasets.
Specifically, we artificially corrupt the measurement data with 10\%, 20\%, and 40\% additive Gaussian noise, where corrupting the data with $k$\% noise is defined as
\begin{equation}
    y_\text{noisy} = y_\text{normalized} + \frac{k}{100}\epsilon,\qquad \epsilon\sim\calN(0,I).
\end{equation}
We present the results, alongside a comparison with EquiNet CDM, in Table~\ref{table:results_noisy} with examples displayed in Figure~\ref{fig:noisy_measurements}. 
For the three synthetic datasets, \ourname remains accurate even in the presence of large amounts of measurement noise and successfully reconstructs most large, prominent features; however, reconstructions are heavily corrupted even at 10\% noise for brain scans.
Interestingly, even though \ourname is deterministic at inference time, it continues to perform on par with the probabilistic CDM when observations are noisy.

\begin{table*}[t]
\caption{Model performance on the four nonlinear inverse scattering datasets with increasing levels of noise added to the data, from 10\% to 40\%. We fix the number of sampling steps for \ourname at 8, and compare to the results of the EquiNet CDM from~\cite{Zhang2025_BPD} with 256 sampling steps. EquiNet CDM results for brain scans are obtained from a model we trained using the pipeline from~\cite{Zhang2025_BPD}.} \label{table:results_noisy}
\centering
\small
\begin{tabular}{llcccc}
\toprule
& & \multicolumn{4}{c}{\textbf{RRMSE} (\%) $\downarrow$} \\
\cmidrule(lr){3-6}
\textbf{Noise Level} & \textbf{Method} & \textbf{Shepp--Logan} & \textbf{3--5--10h Triangles} & \textbf{10h Overlapping Squares} & \textbf{Brain Scans} \\
\midrule
\multirow{2}{*}{0\% Noise}
& \ourname
& 0.379
& 1.108
& 1.143
& 3.421 \\
& EquiNet CDM
& 1.323
& 1.590
& 1.744
& 5.661 \\
\midrule
\multirow{2}{*}{10\% Noise}
& \ourname
& 2.844
& 1.141
& 2.015
& 21.598 \\
& EquiNet CDM
& 3.355
& 1.564
& 2.675
& 21.889 \\
\midrule
\multirow{2}{*}{20\% Noise}
& \ourname
& 7.800
& 1.677
& 5.763
& 38.648 \\
& EquiNet CDM
& 7.422
& 1.618
& 8.464
& 38.805 \\
\midrule
\multirow{2}{*}{40\% Noise}
& \ourname
& 15.827
& 3.524
& 17.842
& 65.231 \\
& EquiNet CDM
& 14.827
& 3.522
& 24.761
& 65.949 \\
\bottomrule
\end{tabular}
\end{table*}

\subsubsection{Inverse Crimes}
An \textit{inverse crime} is where the same mathematical model is used to generate the synthetic data for training and to test the inversion algorithm. We now test how well \ourname responds to out-of-distribution data generated by a different numerical method. 
Specifically, we follow~\cite{Zhang2025_BPD} and assess performance on measurement data generated using lower-order solvers. 
We evaluate the effects of mismatched forward models on nonlinear inverse scattering and seismic imaging because the numerical implementations of the forward models we use for these problems have easily configurable finite-difference stencil orders.

\begin{figure*}[!htbp]
    \centering

    \begin{subfigure}[b]{0.75\textwidth}
        \centering
        \includegraphics[width=\linewidth]{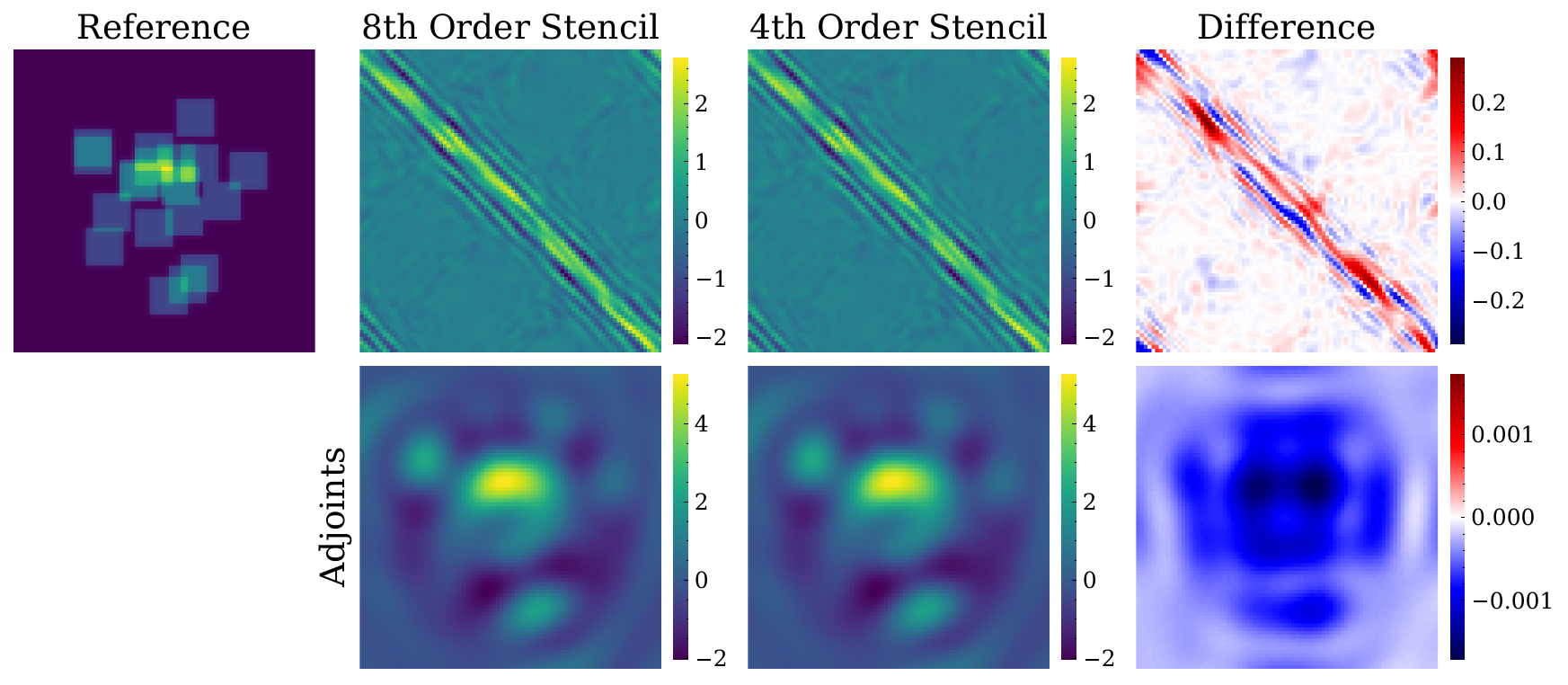}
        \caption{}
    \end{subfigure}

    \vspace{0.75em}

    \begin{subfigure}[b]{0.75\textwidth}
        \centering
        \includegraphics[width=\linewidth]{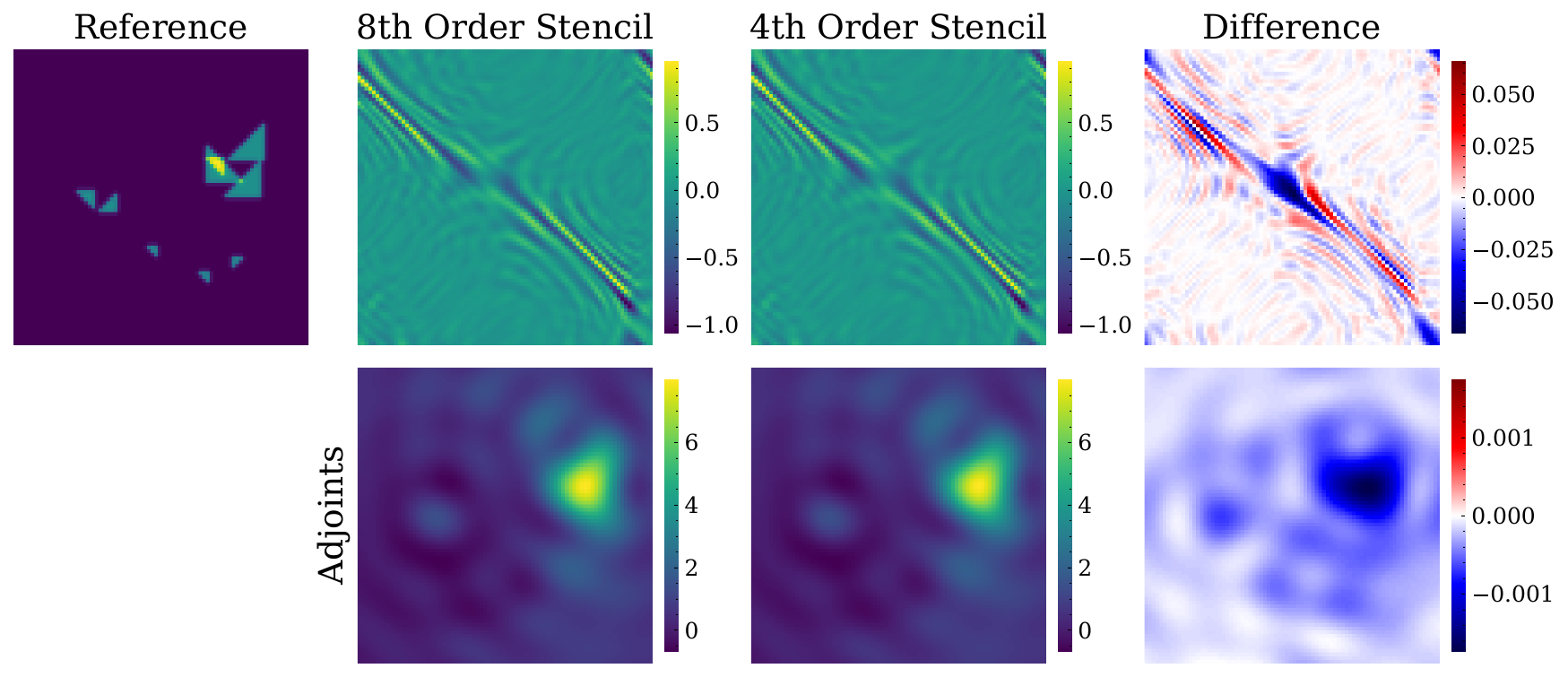}
        \caption{}
    \end{subfigure}

    \vspace{0.75em}

    \begin{subfigure}[b]{0.75\textwidth}
        \centering
        \includegraphics[width=\linewidth]{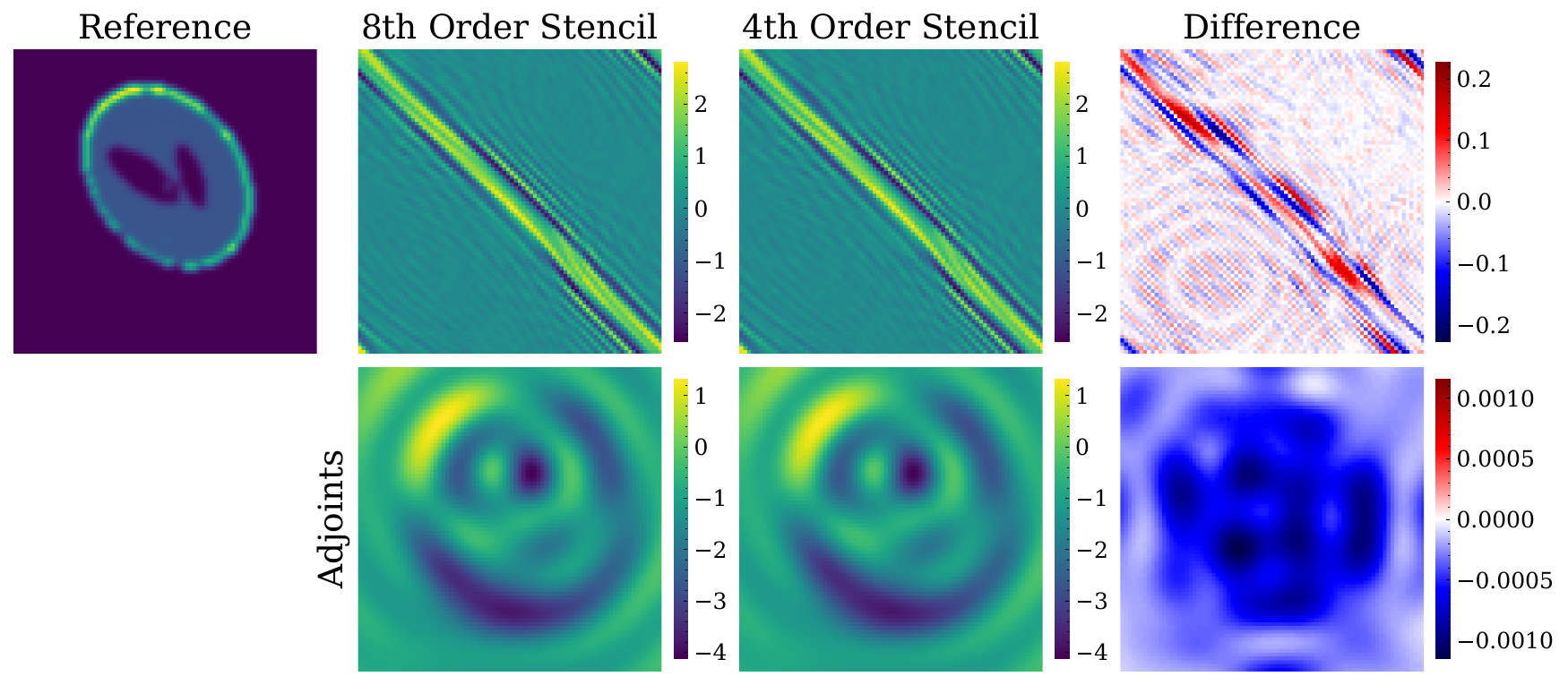}
        \caption{}
    \end{subfigure}

    \caption{Effect of finite-difference stencil order on nonlinear
    inverse scattering for (a) \textit{10h Overlapping Squares},
    (b) \textit{3--5--10h Triangles}, and
    (c) \textit{Shepp--Logan}. Each panel shows the ground-truth
    perturbation at left. The remaining columns compare the
    8th-order and 4th-order results and their signed difference:
    measurement data at the highest probing frequency in the
    top row, and the corresponding adjoint back-projections
    in the bottom row.}
    \label{fig:distributional_mismatch}
\end{figure*}

For nonlinear inverse scattering, the training data was generated using an 8th-order finite-difference scheme, and we test performance on measurements generated with 4th- and 6th-order solvers for the three synthetic inverse scattering datasets.
We visualize the distributional mismatch in Figure~\ref{fig:distributional_mismatch}, and observe that while the scattering data can differ by up to 10\%, the adjoints remain essentially unchanged. 
Results are given in~Table~\ref{table:results_stencil_order} and demonstrate that \ourname is robust to these distributional shifts, with minimal change in performance except for Shepp--Logan with a 4th order stencil.

\begin{table*}[ht]
\caption{\ourname performance on the three synthetic inverse scattering datasets when the test data is generated using different finite-difference stencil orders (training data was generated with an 8th order stencil). 
We report RRMSE as a percentage and MELR $\times10^{-2}$, and fix the number of sampling steps for \ourname at 8.}\label{table:results_stencil_order}
\centering
\small
\resizebox{\textwidth}{!}{%
\begin{tabular}{lccc ccc ccc}
\toprule
\textbf{} &
\multicolumn{3}{c}{\textbf{Shepp--Logan}} &
\multicolumn{3}{c}{\textbf{3--5--10h Triangles }} &
\multicolumn{3}{c}{\textbf{10h Overlapping Squares }} \\
\cmidrule(lr){2-4}\cmidrule(lr){5-7}\cmidrule(lr){8-10}
\textbf{Stencil Order} \hspace{1.5cm} & \textbf{RRMSE} $\downarrow$  & \textbf{MELR}$\downarrow$ & \textbf{SD}$\downarrow$
& \textbf{RRMSE} $\downarrow$ & \textbf{MELR}  $\downarrow$ & \textbf{SD}$\downarrow$
& \textbf{RRMSE}$\downarrow$  & \textbf{MELR}  $\downarrow$ & \textbf{SD} $\downarrow$\\
\midrule
4th Order Stencil
& 0.726  & 0.788 & 0.624
& 1.112  & 0.617 & 0.258
& 1.242  & 1.506 & 0.525 \\
6th Order Stencil
& 0.382  & 0.479 & 0.576
& 1.122  & 0.626 & 0.258
& 1.150  & 1.411 & 0.515 \\
8th Order Stencil
& 0.379  & 0.474 & 0.576
& 1.108  & 0.607 & 0.258
& 1.143  & 1.401 & 0.514 \\
\bottomrule
\end{tabular}
}
\end{table*}

\begin{wraptable}{r}{0.6\linewidth}
\caption{\ourname performance on seismic imaging using
test data generated with 2nd and 4th order spatial finite-difference
stencils (training data was generated with a 4th order stencil). We report performance using 1 sampling step, as we found this to give the best performance based on validation data.}
\label{table:fwi_stencil_order}
\centering
\footnotesize
\setlength{\tabcolsep}{3.5pt}
\renewcommand{\arraystretch}{1.05}
\begin{tabular}{lcccc}
\toprule
\textbf{Stencil Order}
& \textbf{RRMSE (\%)} $\downarrow$
& \textbf{PSNR} $\uparrow$
& \textbf{SSIM} $\uparrow$
& \textbf{Data misfit} $\downarrow$ \\
\midrule
2nd order & 9.7 (1.9) & 20.515 (1.509) & 0.676 (0.062) & 112.39 (29.17) \\
4th order & 9.7 (1.9) & 20.530 (1.516) & 0.676 (0.062) & 115.63 (30.89) \\
\bottomrule
\end{tabular}
\end{wraptable}

\begin{figure*}[t]
    \centering
    \includegraphics[width=0.8\linewidth]{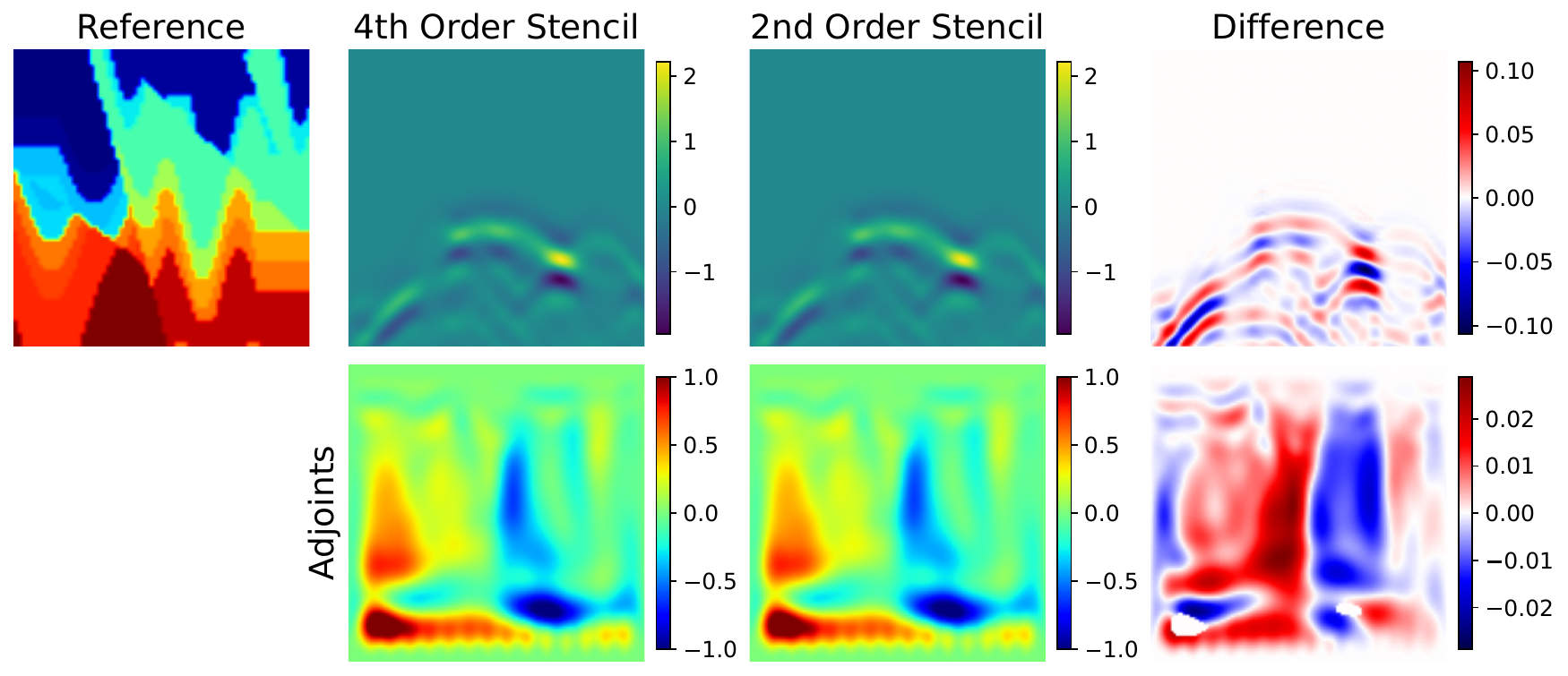}
    \caption{Effect of spatial finite-difference stencil order
    on seismic imaging. The ground-truth wavespeed map is shown
    at left. The remaining columns compare the 4th-order and
    2nd-order results and their signed difference. The top row
    shows single-shot receiver measurements, and the bottom row
    shows the corresponding adjoint-based initializations.
    Each color bar gives the scale of its associated image.}
    \label{fig:fwi_mismatch}
\end{figure*}

For seismic imaging, the training data was generated using a 4th-order spatial finite-difference scheme, and we test performance on measurements generated with 2nd- and 4th-order solvers.
We visualize the distributional mismatch in Figure~\ref{fig:fwi_mismatch}, and again note that the distributional shift has very little impact on the adjoint. 
The results in~Table~\ref{table:fwi_stencil_order} show that the different solver order has negligible impact on performance, further supporting the robustness of \ourname to distributional shifts.

\subsection{Sample Complexity}
We compare the training-data efficiency of \ourname and EquiNet CDM by varying the number of training samples from \(1{,}000\) to \(21{,}000\), while fixing their respective sampling budgets at 8 and 256 steps.
Results are presented in~Table~\ref{table:results_sample_complexity}, and show that \ourname consistently achieves lower RRMSE across all four datasets and all training-set sizes.
The improvement is already evident with only \(1{,}000\) samples and persists as additional training data are introduced.
Notably, \ourname requires far fewer training samples to achieve comparable accuracy to EquiNet CDM.
These results indicate that the performance gains of \ourname are not restricted to the large-data regime and suggest improved training-data efficiency relative to EquiNet CDM.

\begin{table*}[ht]
\caption{Model performance on four inverse scattering datasets as the number of training samples increases from 1{,}000 to 21{,}000. We fix the number of sampling steps for \ourname at 8, and compare to the results of the EquiNet CDM from~\cite{Zhang2025_BPD} with a fixed budget of 256 sampling steps. EquiNet CDM results for brain scans are obtained from a model we trained using the pipeline  from~\cite{Zhang2025_BPD}.}\label{table:results_sample_complexity}
\centering
\small
\resizebox{\textwidth}{!}{%
\begin{tabular}{llcccc}
\toprule
\textbf{Training Samples} & \textbf{Method} &
\textbf{Shepp--Logan} &
\textbf{3--5--10h Triangles} &
\textbf{10h Overlapping Squares} &
\textbf{Brain Scans} \\
\cmidrule(lr){3-6}
& & \textbf{RRMSE} $\downarrow$ & \textbf{RRMSE} $\downarrow$ & \textbf{RRMSE} $\downarrow$ & \textbf{RRMSE} $\downarrow$ \\
\midrule
\multirow{2}{*}{1{,}000}
& \ourname
& 1.912
& 4.356
& 7.933
& 5.476 \\
& EquiNet CDM
& 2.085
& 8.475
& 12.454
& 9.485 \\
\midrule
\multirow{2}{*}{2{,}000}
& \ourname
& 1.242
& 3.136
& 5.856
& 4.829 \\
& EquiNet CDM
& 1.644
& 5.454
& 7.893
& 7.995 \\
\midrule
\multirow{2}{*}{4{,}000}
& \ourname
& 0.853
& 2.582
& 4.014
& 4.331 \\
& EquiNet CDM
& 1.379
& 3.525
& 4.896
& 7.120 \\
\midrule
\multirow{2}{*}{8{,}000}
& \ourname
& 0.612
& 1.647
& 2.424
& 4.004 \\
& EquiNet CDM
& 1.376
& 2.824
& 2.756
& 6.291 \\
\midrule
\multirow{2}{*}{16{,}000}
& \ourname
& 0.426
& 1.529
& 1.481
& 3.568 \\
& EquiNet CDM
& 1.281
& 1.957
& 2.184
& 5.833 \\
\midrule
\multirow{2}{*}{21{,}000}
& \ourname
& 0.379
& 1.108
& 1.143
& 3.421 \\
& EquiNet CDM
& 1.323
& 1.590
& 1.744
& 5.661 \\
\bottomrule
\end{tabular}
}
\end{table*}

\subsection{Empirical Transport Distances}
\label{app:transport_distances}

To illustrate the geometry of the initializations in the concentrated-posterior regime, we approximate the conditional target for each measurement--reconstruction pair $(y,m)$ by $\delta_m$.
Writing $a(y)$ for the initialization used by \ourname and $d$ for the dimension of the reconstruction, we apply~\eqref{eq:wasserstein2_gaussian} to explicitly compute
\begin{equation}\label{eq:empirical_W2}
    \mathcal{W}_2\big(\mathcal{N}(0,I),\delta_m\big)
    = \sqrt{\|m\|_2^2+d},\qquad
    \mathcal{W}_2\big(\delta_{a(y)},\delta_m\big)
    = \|a(y)-m\|_2.
\end{equation}
We report the mean and standard deviation of these distances across the training examples in each dataset. Table~\ref{table:wasserstein} shows that the concentrated, physics-based \ourname initialization $\delta_{a(y)}$ reduces transport distance compared to a standard Gaussian on every dataset we test.
While our transport-distance analysis provides strong geometric motivation for the \ourname initialization, it does not necessarily establish a causal relationship with observed model performance.
Additionally~\eqref{eq:empirical_W2} provides a useful estimate when the training dataset is finite and discrete, but does not capture posterior variability that may exist in real-world settings.

\begin{table}[ht]
\caption{Comparison of Wasserstein-2 distances estimated numerically between the reference and target distributions when the base distribution is a standard Gaussian $x_0\sim\calN(0, I)$ vs. when the base distribution $\rho_0=\delta_{a(y)}$ is concentrated at the physics-based approximation. We report the averages across each training dataset with standard deviation shown in parentheses.}
\label{table:wasserstein}
\centering
\begin{small}
\resizebox{\textwidth}{!}{%
\begin{tabular}{lccccccc}
\toprule
\textbf{Reference Distribution} \hspace{0.2cm} & \textbf{Shepp--Logan} & \textbf{3-5-10h Tri} & \textbf{10h Squares} & \textbf{Brain Scans} & \textbf{Cells} ($N_\text{rec}=60/180/360$) & \textbf{Seismic Imaging} & \textbf{Black Holes}\\
\midrule
Gaussian & 113.1 (3.0) & 111.6 (18.7) & 113.1 (3.8) & 112.8 (8.6) & 180.7 (11.4) & 180.1 (18.3) & 89.7 (11.9)\\
Physics-based & 94.1 (16.0) & 77.4 (22.9) & 68.6 (7.2) & 95.7 (24.3) & 53.6 (10.1) / 24.0 (4.8) / 22.8 (4.7) & 169.6 (29.0) & 48.9 (11.3)\\
\bottomrule
\end{tabular}
}
\end{small}
\vskip -0.1in
\end{table}

\subsection{Auxiliary Training Techniques}
We observed modest gains for \ourname in some settings from flow-matching training techniques adapted from Lee et al.~\cite{lee2024improving}. These fall outside our main scope but suggest directions for future work.

Adapting~\cite{lee2024improving}, we replace the squared $\ell_2$ loss in~\eqref{eq:loss_cfm} with a Pseudo-Huber loss combined with a time-weighted, SSIM-based~\eqref{eq:ssim} perceptual term:
\begin{equation}
    \mathcal{L}_{\mathrm{ssim\text{-}hub}}[\hat{v}]
    = \int_0^1 \mathbb{E}\!\left[
    t \cdot \Big(\sqrt{\|x_1-x_0-\hat{v}_t(x_t; y)\|_2^2+c^2}-c\Big)
    - \frac{1}{1-t}\,\mathrm{SSIM}(\hat{x}_1, x_1)
    \right] \, dt,
\end{equation}
where $c=0.00054d$ for an image $x_t\in\mathbb{R}^d$, and $\hat{x}_1 = x_t + (1-t)\hat{v}_t(x_t;y)$.
One may also use $\hat{x}_1$ to add a data consistency term to the training loss, but we generally found this to be either computationally infeasible and/or ineffective in our limited experiments.

Additionally, one may optimize a reweighted objective.
The U-shaped time-weighting scheme~\eqref{eq:u_shaped_time} of Lee et al.~\cite{lee2024improving}, with their choice $\xi=4$, biases the endpoints of $[0,1]$ where learning may be more difficult.

\section{Implementation Details}\label{app:model_details}
We implemented all models using JAX\footnote{\url{https://github.com/jax-ml/jax}}, FLAX\footnote{\url{https://github.com/google/flax}} and swirl-dynamics\footnote{\url{https://github.com/google-research/swirl-dynamics}}, and trained on NVIDIA Quadro RTX 6000, RTX A6000, L40S, and H100 GPUs.
We use Adam~\cite{kingma2015adam} with linear warm-up from $10^{-5}$ to $10^{-3}$ over the first 5\% of training, followed by cosine annealing to $10^{-6}$. We targeted the benchmark models' epoch counts, reducing batch sizes when GPU memory required it.
We use InverseBench's publicly available implementations, pretrained diffusion prior checkpoints, and hyperparameter choices\footnote{\url{https://github.com/devzhk/InverseBench}} for linear inverse scattering, FWI, and black hole imaging.
For our own PnP implementations, we follow InverseBench's tuning procedure~\cite{zheng2025inversebench} with comparable hyperparameter ranges and tuning iterations, using Optuna~\cite{akiba2019optuna} for Bayesian optimization. For \ournameRED we use 500 optimization iterations for nonlinear inverse scattering and 1000 for other problems. Tables~\ref{table:model_params} and~\ref{table:dataset_params} report parameter counts and dataset details.

\begin{table}[!htbp]
\caption{Neural network parameter counts by problem. As in Algorithm~\ref{alg:adjointflow_train}, $\theta$ denotes the main score/velocity/target-prediction network parameters and $\phi$ the conditional branch parameters. PnP models have no conditional branch ($|\phi|=0$).}
\label{table:model_params}
\centering
\begin{small}
\begin{tabular}{ll cc}
\toprule
\textbf{Model} & \textbf{Problem} & \textbf{$|\phi|$} & \textbf{$|\theta|$}\\
\midrule
EquiNet-CNN & Nonlinear inverse scattering & $156k$ & $375k$ \\
ADM (PnP) & Nonlinear inverse scattering & 0 & $26.8M$ \\
EquiNet-CNN & Linear inverse scattering ($N_\text{rec}$=60/180/360) & $20k/145k/549k$ & $2.73M$ \\
ADM (PnP) & Linear inverse scattering & 0 & $26.8M$ \\
ADM (PnP) & Seismic Imaging & 0 & $26.8M$ \\
ADM (conditional) & Seismic Imaging & $550k$ & $27.8M$ \\
CNN & Black hole imaging (obs. ratio=3\%/10\%/100\%) & $128k/263k/862k$ & $1.89M$ \\
NCSN++ (PnP) & Black hole imaging & 0 & $20.0M$ \\
\bottomrule
\end{tabular}
\end{small}
\vskip -0.1in
\end{table}

\begin{table}[!htbp]
\caption{Dataset parameters and forward/gradient evaluation timings (medians of 50 evaluations after warm-up). FWI measurements are Fourier-transformed and sub-sampled to $16\times33\times20$ (Section~\ref{sec:time_domain_seismic_imaging}). Linear inverse scattering uses $N_\text{rec}\in\{60,180,360\}$ receivers; black hole observation ratios are $3\%$, $10\%$, and $100\%$. Inverse scattering uses one Quadro RTX 6000 GPU; seismic and black hole imaging use Intel Xeon Gold 6248R CPUs on the same node, with 16 single-threaded workers and one CPU thread, respectively, following~\cite{zheng2025inversebench}.}
\label{table:dataset_params}
\centering
\begin{small}
\resizebox{\textwidth}{!}{%
\begin{tabular}{l cccccc}
\toprule
\textbf{Problem} \hspace{1cm} & \textbf{$N_{\text{train}}$} & \textbf{$N_{\text{test}}$} & \textbf{Dimension of $m$} & \textbf{Dimension of $y$} & \textbf{Forward Eval. (sec)} & \textbf{Gradient Eval. (sec)} \\
\midrule
Nonlinear inverse scattering & 21000 & 500 & $80\times80$ & $80\times80\times2\times3$ & 2.310 & 2.455 \\
Linear inverse scattering & 10000 & 100 & $128\times128$ & $20\times N_{\text{rec}}\times2$& 0.004 / 0.005 / 0.006 & 0.011 / 0.012 / 0.015 \\
Seismic Imaging & 50000 & 10 & $128\times128$ & $16\times129\times1001$ & $4.004$ & $6.261$ \\
Black hole imaging & 50000 & 100 & $64\times64$ & $185/885/4007$ & 0.001 / 0.002 / 0.008 & 0.005  / 0.019 / 0.111\\
\bottomrule
\end{tabular}
}
\end{small}
\vskip -0.1in
\end{table}

\clearpage
\section{Additional Results}

\begin{figure*}[!htbp]
    \centering
    \includegraphics[width=5in]{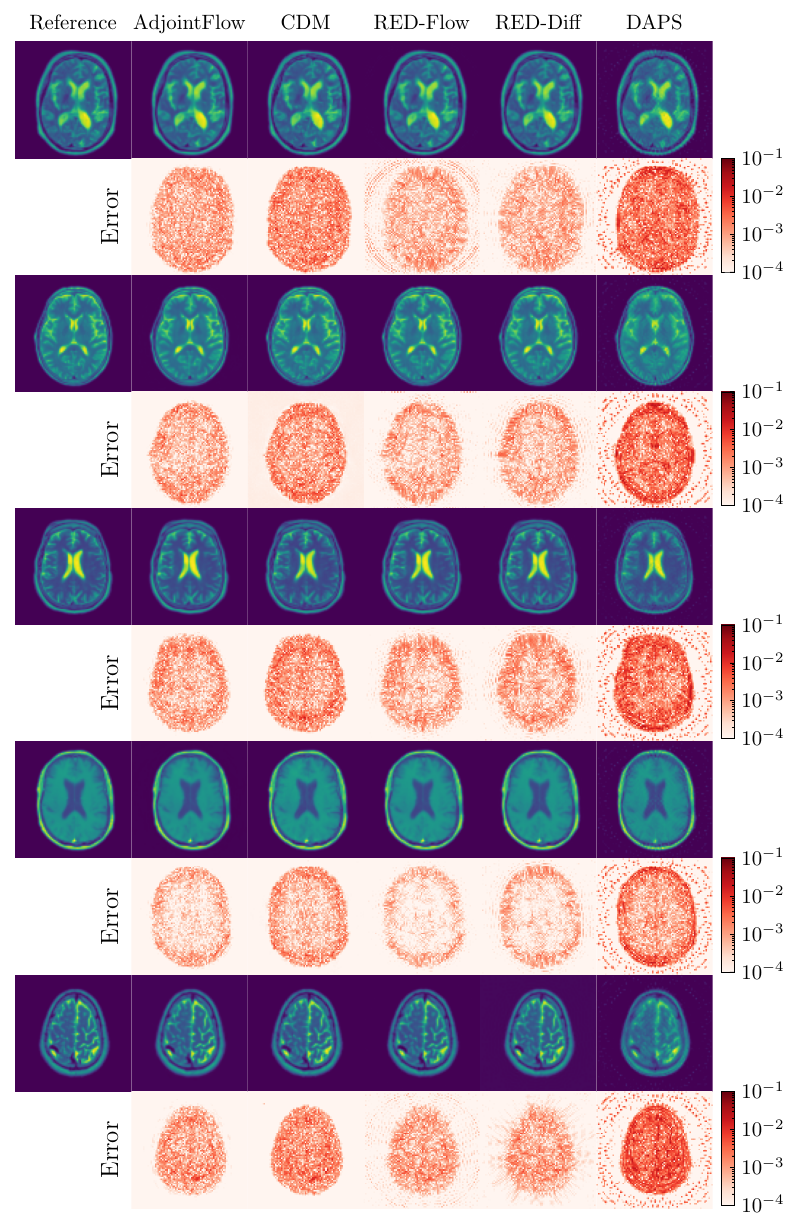}
    \caption{Further qualitative comparisons of \ourname with different diffusion-based methods on nonlinear inverse scattering for the \textit{Brain Scans} dataset.}
    \label{fig:brain_extended}
\end{figure*}

\begin{figure*}[!htbp]
    \centering
    \includegraphics[width=\linewidth]{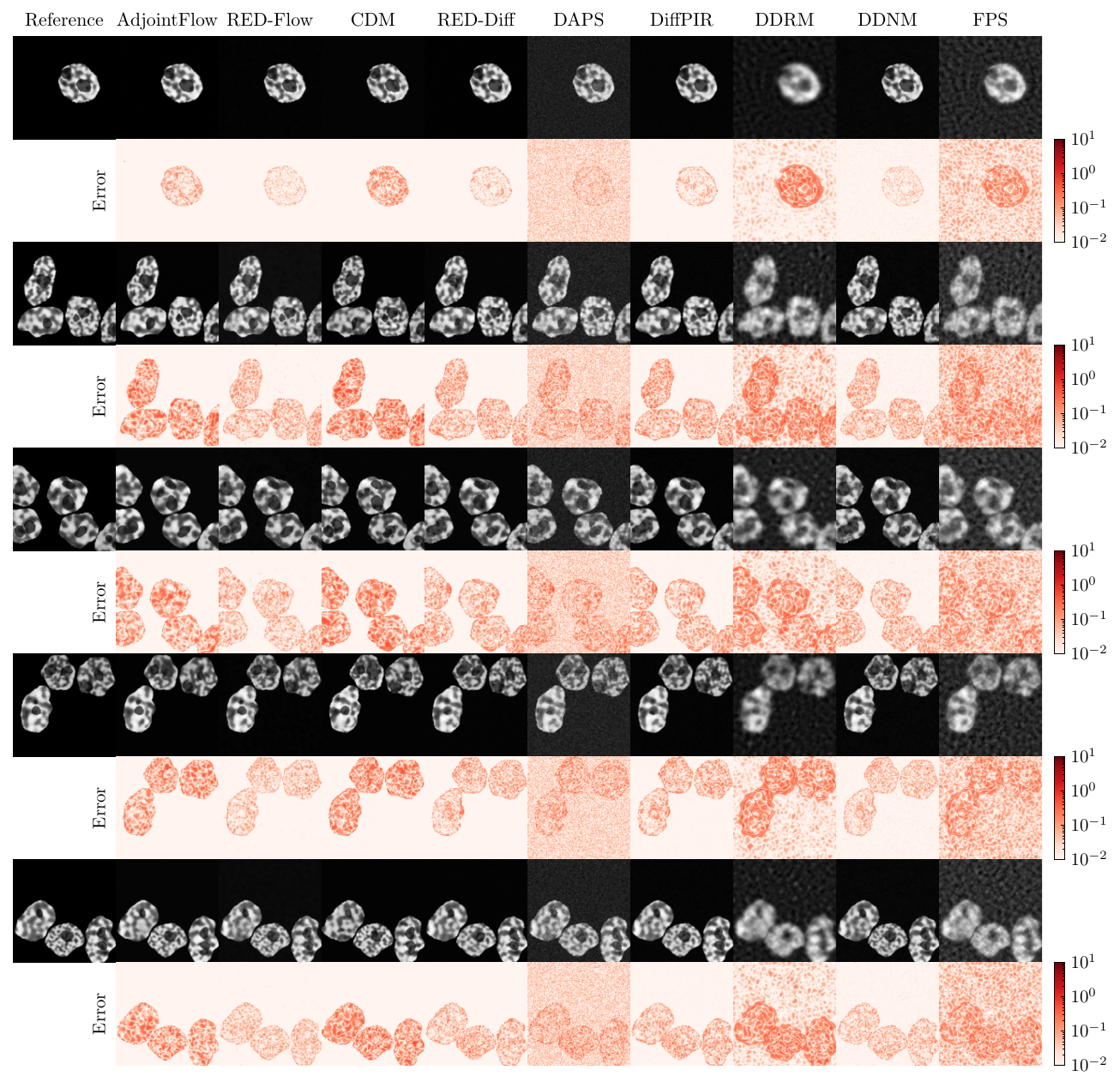}
    \caption{Further qualitative comparisons of \ourname against diffusion baselines for linear inverse scattering with 60 receivers.}
    \label{fig:inv_scatter_extended}
\end{figure*}

\begin{figure*}[!htbp]
    \centering
    \includegraphics[width=6in]{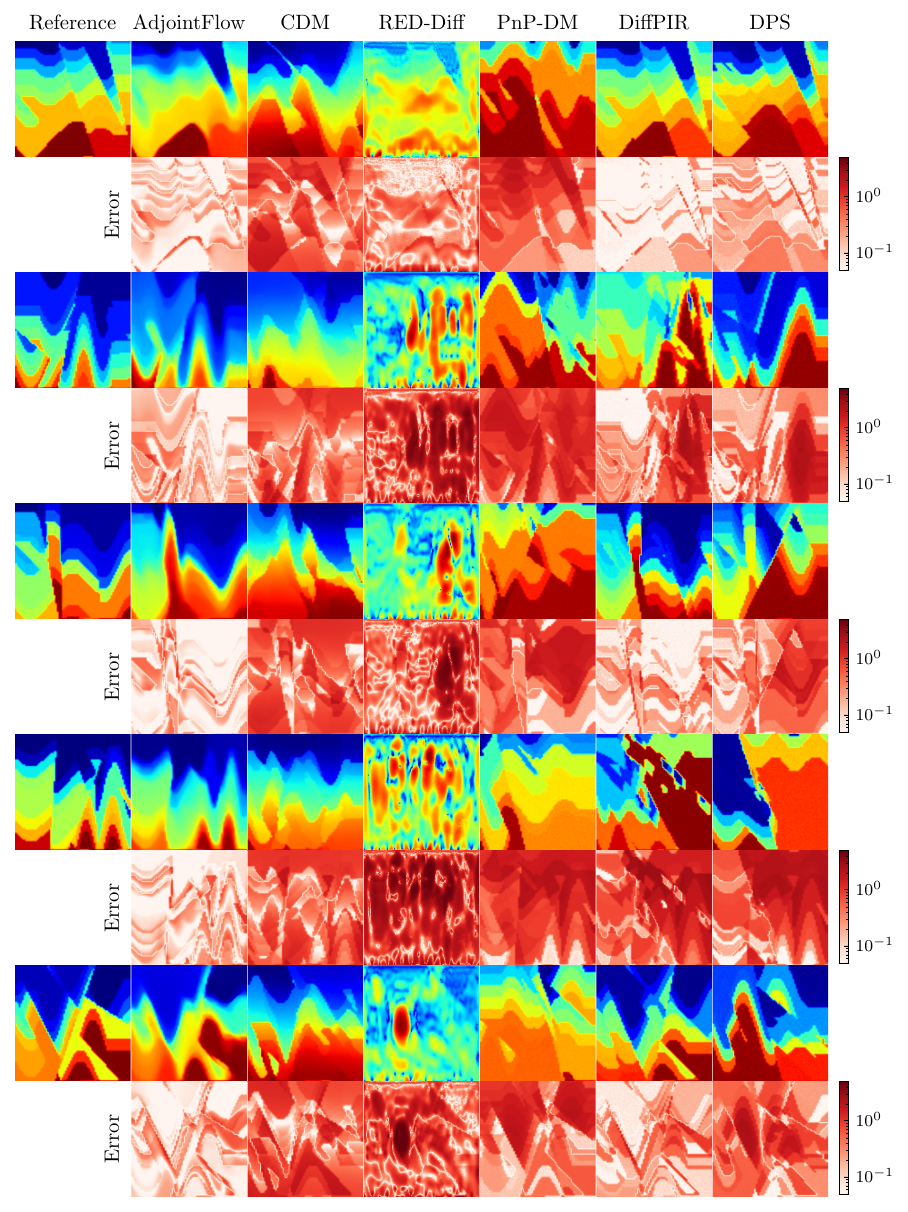}
    \caption{Further qualitative comparisons of \ourname against diffusion baselines for seismic imaging. Note that while other methods occasionally produce accurate reconstructions, only \ourname consistently avoids cycle-skipping.}
    \label{fig:fwi_extended}
\end{figure*}

\begin{figure*}[!htbp]
    \centering
    \includegraphics[width=\linewidth]{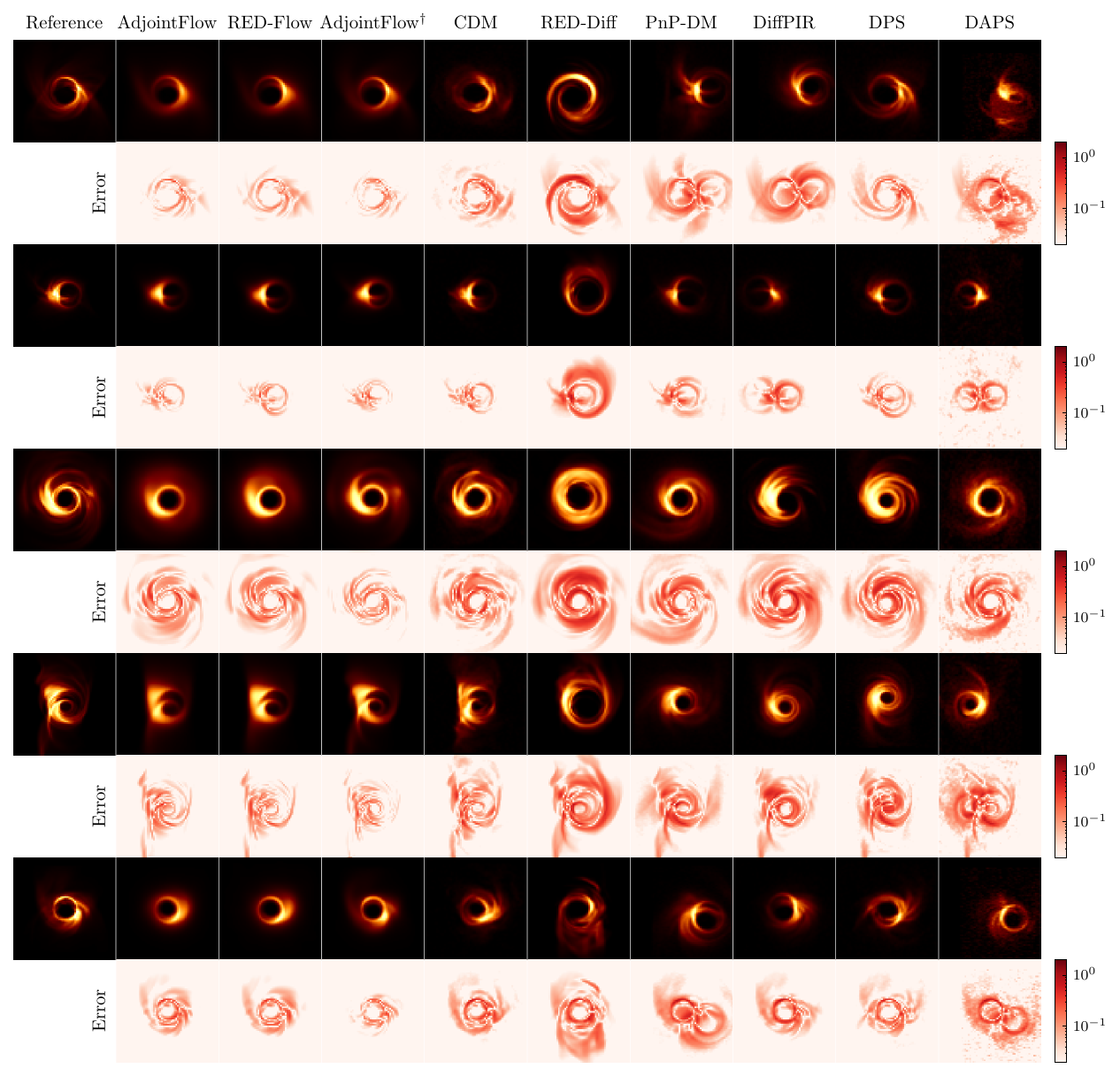}
    \caption{Further qualitative comparisons of \ourname against diffusion baselines for black hole imaging with 3\% observation ratio. }
    \label{fig:blackhole_comparison_full}
\end{figure*}

We provide additional qualitative comparisons for nonlinear inverse scattering on \textit{Brain Scans} in Figure~\ref{fig:brain_extended}, linear inverse scattering in Figure~\ref{fig:inv_scatter_extended}, seismic imaging in Figure~\ref{fig:fwi_extended}, and black holes in Figure~\ref{fig:blackhole_comparison_full}. These examples complement the representative reconstructions shown in the main text and illustrate performance across a broader set of test instances.

\end{document}